\documentclass[sigconf, 9pt]{acmart}

\usepackage[ruled,linesnumbered]{algorithm2e} % For algorithms
\usepackage{booktabs} % For formal tables
\usepackage{balance} % For balanced columns on the last page
\usepackage{graphicx}
\usepackage{url}
\usepackage{caption}
\usepackage{epstopdf}
\usepackage{bm}
\usepackage{subcaption}
\usepackage{pdfpages}
\usepackage{capt-of}
\usepackage{tabularx}
\usepackage{epsfig}
\usepackage{xcolor}
\usepackage{multirow}
\usepackage{nomencl}
\usepackage{amsmath}
\usepackage{mathtools}
\usepackage{siunitx} 
\usepackage{enumitem}

\usepackage[mathscr]{euscript}
\newcommand{\sysname}{\texttt{milliEgo}\xspace}

\newcommand{\revise}[1]{{\color{black}{#1}}}

\newcommand{\norm}[1]{\left\lVert#1\right\rVert}
\newcommand{\sect}{\textsection}
\title[milliEgo]{\sysname: Single-chip mmWave Radar Aided Egomotion Estimation via Deep Sensor Fusion}
\author{
Chris Xiaoxuan Lu$^{1}$, Muhamad Risqi U. Saputra$^{2}$, Peijun Zhao$^{2}$, Yasin Almalioglu$^{2}$,}
\author{Pedro P. B. de Gusmao$^{2}$, Changhao Chen$^{2}$, Ke Sun$^{3}$, Niki Trigoni$^{2}$, Andrew Markham$^{2}$}

\affiliation{%
  \institution{$^{1}$ University of Edinburgh, Edinburgh, Scotland, United Kingdom}
  \institution{$^{2}$ University at Oxford, Oxford, England, United Kingdom}
  \institution{$^{3}$ University of California San Diego, San Diego, California, USA}
}



\renewcommand{\shortauthors}{C. Lu et al.}
\begin{abstract}

Robust and accurate trajectory estimation of mobile agents such as people and robots is a key requirement for providing spatial awareness for emerging capabilities such as augmented reality or autonomous interaction. Although currently dominated by optical techniques e.g., visual-inertial odometry, these suffer from challenges with scene illumination or featureless surfaces. As an alternative, we propose \sysname, a novel deep-learning approach to robust egomotion estimation which exploits the capabilities of low-cost mmWave radar. Although mmWave radar has a fundamental advantage over monocular cameras of being metric i.e., providing absolute scale or depth, current single chip solutions have limited and sparse imaging resolution, making existing point-cloud registration techniques brittle. We propose a new architecture that is optimized for solving this challenging pose transformation problem. Secondly, to robustly fuse mmWave pose estimates with additional sensors, e.g. inertial or visual sensors we introduce a mixed attention approach to deep fusion. Through extensive experiments, we demonstrate our proposed system is able to achieve 1.3\% 3D error drift and generalizes well to unseen environments. We also show that the neural architecture can be made highly efficient and suitable for real-time embedded applications.

\end{abstract}

\copyrightyear{2020} 
\acmYear{2020} 
\setcopyright{acmcopyright}\acmConference[SenSys '20]{The 18th ACM Conference on Embedded Networked Sensor Systems}{November 16--19, 2020}{Virtual Event, Japan}
\acmBooktitle{The 18th ACM Conference on Embedded Networked Sensor Systems (SenSys '20), November 16--19, 2020, Virtual Event, Japan}
\acmPrice{15.00}
\acmDOI{10.1145/3384419.3430776}
\acmISBN{978-1-4503-7590-0/20/11}

\begin{document}

\maketitle

%%%%%%%%%%%%%%%%%%%%%%%%%%%%%%%%%%%%%%%%%%%%%%%%%%%%%%%%%%%%%%%%%%%%%%%%%%%%%%
%!TEX root = ../main.tex
\section{Introduction}

\begin{figure}[!t]
	\centering
	\includegraphics[width=0.93\columnwidth]{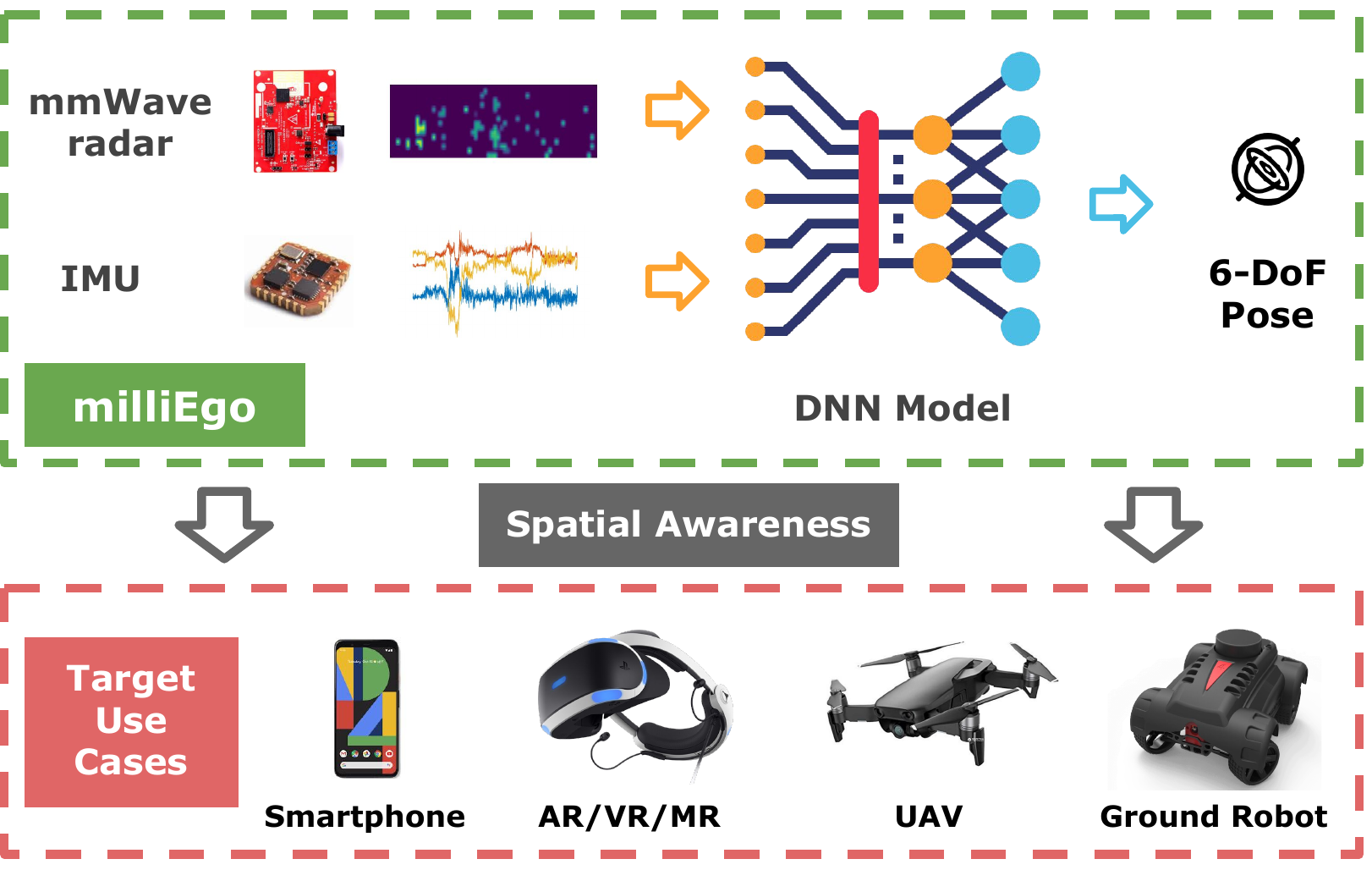} 
\caption{Our proposed \sysname uses a low-cost COTS mmWave radar and IMU coupled with a deep neural network model to accurately and robustly estimate 6-DOF egomotion. We envision that \sysname could be widely-integrated on mobile platforms to provide a high degree of spatial awareness to a wide range of applications.}
\label{fig:open_fig}
\vspace{-0.4cm}
\end{figure}

% What is egomotion estimation and why its important to many applications.
From navigating on distant planets to tracking the pose of an augmented reality headset, egomotion awareness plays a vital role in perception and interaction for mobile agents. Unlike map-based localization, egomotion estimation\footnote{Egomotion estimation is also known as odometry estimation in robotics and computer vision community. In this paper, we use \emph{egomotion} to exclusively represent the process of relative pose estimation while \emph{odometry} exclusively refers to the composed trajectory from series of relative poses.} does not require any prior knowledge (e.g., floor plans) about the environment nor any infrastructure setup (e.g., wireless access points), yet it can ubiquitously determine the position and orientation of a mobile agent over time by analyzing sensory data from its movement. Accurate odometry is also a necessary and essential precursor for building mapping, e.g., with SLAM techniques \cite{lu2018simultaneous,cadena2016past}.

%\chris{new motivation -  why current IMU+X is not enough}

Owing to their low cost and ubiquity, MEMs inertial sensors (IMUs) have been widely used as the \emph{de-facto} solution to egomotion estimation for a variety of mobile platforms \cite{lymberopoulos2015realistic,shen2018closing}. Unfortunately, the accuracy of MEMs IMUs is limited by noise and bias and consequently inertial odometry suffers from large drift (e.g., SINS \cite{chen2018ionet}) or is impacted by motion dynamics and sensor attachment changes (e.g. PDR \cite{radu2013himloc}). To address the limitations of IMU, multi-modal odometry systems that integrate inertial information with other sensory data (e.g., visual or ranging information) have been proposed. Amongst these system, visual-inertial odometry (VIO) \cite{leutenegger2015keyframe} is one of the most dominant, with its advantage of being ubiquitously available in mobile phones etc. However, the performance of VIO degrades or even fails under adverse lighting conditions (e.g., RGB cameras cannot operate in the dark and depth cameras suffer from glare and strong illumination)~\cite{saputra2020deeptio} or when operating in relatively featureless environments. Similar visibility issues also impact lidar-inertial odometry (LIO), especially in the presence of airborne obscurants (e.g., dust, fog and smoke)~\cite{richardson2011strengths}. Moreover, lidars are often heavy, bulky and expensive in comparison to cameras and are typically found in high-end robotics rather than in micro-robots or wearable applications. 

In order to explore inexpensive alternatives to the above optically based systems, we propose egomotion estimation with single-chip millimetre wave (mmWave) radar in combination with an IMU. Driven by recent advances in the nanoscale CMOS technology \cite{guermandi201779}, these COTS devices have emerged as an innovative low-cost, low-power sensor modality in the automotive industry \cite{timmwave}. 
A key advantage of mmWave radar over vision is in its robustness to environmental conditions, e.g., it is agnostic to scene illumination and to airborne obscurants.  Compared with lidar or mechanically scanning radar (e.g., CTS350-X \cite{weston2018probably}), single-chip mmWave radars use electronic beamforming and are therefore lightweight and able to fit the payloads of micro robots and form factors of mobile or wearable devices, e.g., the TI IWR6843 even integrates the antennas within the chip package. Smartphones such as the Google Pixel $4$ have recently adopted mmWave radar as their on-board sensor for motion sensing \cite{pixel4} while commercial drones have used mmWave radars for obstacle detection \cite{radar_drone}.  Furthermore, as it is Radio Frequency (RF) based, it does not require optical lenses and can be integrated into plastic housings \cite{ti_plastic}, making them highly resilient to water and dust ingress. We therefore envision that odometry based on a mmWave radar will allow robust egomotion estimation in complex situations (e.g., in subway tunnels, for firefighting and in collapsed buildings), as well as serve as a new enabler for ubiquitous mobility with mobile devices, e.g., for VR/AR.

% Conventional point registration method fall short to deal with the sparse and noisy point clouds. How to effectivelly integrate mmWave with other sensors remain unknown. 

Transforming this vision into a reliable indoor odometry system, however, requires addressing multiple challenges. \emph{Firstly}, due to specular (mirror-like) reflections, diffraction \cite{maccartney2016millimeter} and significant multi-path, the radar returns are corrupted by noise - our studies indicate as many as $75\%$ of points are outliers. \emph{Secondly},  due to hardware constraints on the number of antennas, the resultant point clouds are highly sparse due to limited angular resolvability, e.g., objects less than fifteen degrees apart are merged into a single point. Thus, a typical radar point cloud has 100x fewer points than a corresponding lidar scan. Such low-quality data makes conventional methods designed for lidar data (e.g., ICP \cite{civera20101}) fail when directly applied to mmWave data. 
\emph{Thirdly}, although multi-sensor odometry can lower the estimation drift \cite{chen2019selective}, it remains unknown to what extent an mmWave radar can complement inertial and other pervasive sensors e.g. RGB camera, and how best to fuse this. \emph{Lastly}, when applying recent advances in deep neural networks (DNNs) as used in visual or lidar odometry, computational load can be significant which hampers their adoption on mobile, wearable devices and other resource-constrained devices. We investigate whether mmWave radar can be made more computationally efficient due to its inherent sparsity.

% due to incremental pose estimation without the aid of a global reference, all odometry systems are subject to drift, where errors accumulate over time. A practical way to lower the drift is to use multi-modal odometry (e.g., Visual Inertial Odometry (VIO) \cite{leutenegger2015keyframe}) and exploit the complementary nature of different sensors. 

% Nevertheless, it remains unknown to what extent an mmWave radar can complement other pervasive sensors and how best to fuse this. 

Towards addressing these challenges, we propose \sysname, a novel mmWave(-aided) odometry framework that is able to robustly estimate the egomotion of a mobile platform. \sysname follows an end-to-end design and leverages data-driven learning to combat the intrinsic limitations of conventional point registration methods. \sysname demonstrates the feasibility of mmWave odometry and provides a fusion framework to develop a robust mmWave-Inertial Odometry. Despite the use of a deep neural network (DNN), \sysname achieves low latency estimates on embedded platforms. In summary, our contributions are as follows:

\begin{itemize}[leftmargin=*]
    \item A first-of-its-kind DNN based odometry approach that can estimate the egomotion from the sparse and noisy data returned by a single-chip mmWave radar. Unlike conventional methods relying on explicit point matching, \sysname directly learns the motion transformation, making odometry feasible and reliable.
    \item \revise{Systematically investigating the ideal combination of different neural attention mechanisms for multimodal sensor fusion.} Observing the limitation of single-stage self attention, \sysname introduces a two-stage cross-modal attention layer to promote complementary sensor behaviors, yielding robust egomotion estimation ($1.3\%$ 3D error drift) in the wild.   
    \item A real-time prototype implementation with extensive real-world evaluations, including testing for both mobile robots and hand-held devices. The dataset and code are released to the community.
\end{itemize}

%!TEX root = ../main.tex
\section{Primer} % (fold)
\label{sec:primer}

\begin{figure*}[!t]
	\centering
	\includegraphics[width=0.8\textwidth]{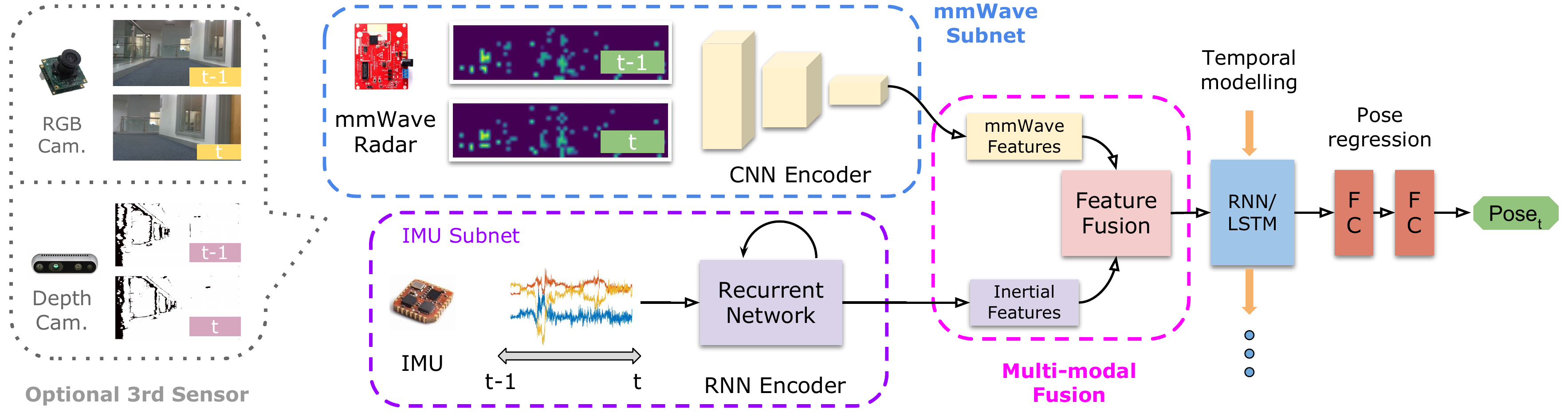} 
\caption{Overview of \sysname, in which \emph{mmWave subnet} and \emph{multi-modal fusion} are two key submodules in its design. The default sensors of \sysname is mmWave and IMU, it can also flexibly integrate the third sensor such as RGB or depth camera.}
\label{fig:milliEgo_overview}
\vspace{-0.3cm}
\end{figure*}

\subsection{Principles of mmWave Radar} % (fold)
\label{sub:principles_of_mmwave_radar}

\noindent \textbf{Range Measurement.}
The single chip mmWave radar uses a frequency modulated continuous wave (FMCW) approach \cite{uttam1985precision}, and has the ability to simultaneously measure both the range and relative radial speed of the target. In FMCW, a radar uses a linear `chirp' or swept frequency transmission. When receiving the signal reflected by an obstacle, the radar front-end performs a dechirp operation by mixing the received signal with the transmitted signals, which produces an Intermediate Frequency (IF) signal. The distance between object and radar can be calculated from the IF signal \cite{ti_training}. For the radar platform used, the range resolution is $\approx$4~cm.

% \begin{equation}
% d=\frac{f_{IF}c}{2S}
% \end{equation}
% where $c$ represents the light speed $3 \times 10^8m/s$, $f_{IF}$ is the frequency of the IF signal, and $S$ is the frequency slope of the chirp. In the presence of multiple obstacles at different ranges, a fast Fourier transform (FFT) is performed on the IF signal, where each peak after FFT represents one or more obstacles at a corresponding distance. 

\noindent \textbf{Angle Measurement.}
A mmWave radar estimates the obstacle angle by using a linear receiver antenna array.
Signal received at different receiver antennas might have different phase due to the distance between the receivers. Based on the differences in phase of the received signals and the distance between them, the angle of arrival for the reflected signal can be estimated \cite{rong2006angle}. Although massive MIMO phased antennas are an active area of research \cite{zhao2020m}, they are not commercially available in single chip radars. Instead, the radar used consists of a $3\times4$ MIMO array, yielding 12 virtual antennas. The resulting angular resolution is poor ($15^{\circ}$ in azimuth, $58^{\circ}$ in elevation) and targets which are closely spaced will be `smeared' together. Subsequent to range and angle estimation, strong peaks are detected which yield a compact set of 3-D points. 

%\noindent \textbf{Radar Signal Processing}
%Subsequent to range and angle FFTs, peaks corresponding to strong reflectors are extracted using an adaptive threshold algorithm such as CFAR~\cite{}. This yields a compact set of 3-D points, typically 10-100 for a typical scene. This point-cloud is used for the following phase of point set registration as discussed in the following.

% Formally, the AoA estimated from any two receiver antennas can be calculated as:
% \begin{equation}
% \psi=sin^{-1}(\frac{\lambda \omega}{2\pi d})
% \end{equation}
% where $\omega$ denotes the phase difference, $d$ represents the distance between consecutive antennas and $\lambda$ is the wave length.

% When more than two receiver antennas are available, sophisticated beamforming or super-resolution algorithms, such as MUSIC \cite{odendaal1994two} can be used to obtain the AoA. At this point, the position of a reflecting obstacle can be jointly determined by its angle and range.

\subsection{Point Set Registration} % (fold)
\label{sub:point_set_registration}

At the heart of egomotion estimation from consecutive point clouds is point set registration \cite{myronenko2010point}, also known as point matching or point alignment. It aims to find the relative positions and orientations i.e., the transformation of the separately acquired views through maximization of the set of overlapping points. Assume two finite size point sets $\mathcal{A} = \{\mathbf{a}_i\}_{i=1}^{N}$ and $\mathcal{B} = \{\mathbf{b}_j\}_{j=1}^{M}$ acquired by a sensor in a 3D space $\mathbb{R}^{3}$. In the case $N \leq M$ where $\mathcal{A}$ serves as the reference, a registration starts by searching for correspondence in $\mathcal{B}$. Detected corresponding points are re-ordered in accordance to their counterparts in $\mathcal{A}$, denoted as a new set $\mathcal{\bar{B}} = \{\mathbf{b}_i\}_{i=1}^{N}$. Once the correspondence is available, a transformation model $T$ on point set $\mathcal{A}$ can be estimated to yield the best alignment between the transformed set $T(\mathcal{A})$ and the corresponding point set $\mathcal{\bar{B}}$. Under the rigid transformation assumption, deriving the optimal $T$ is equivalent to solving a least square minimization problem for all pairs:

% \begin{equation}
%  dist(T({\mathcal{A}}), \mathcal{B}) = \sum_{\mathbf{a}_i\in T({\mathcal {A}})}\sum_{\mathbf{b}_j\in {\mathcal {B}}}(\mathbf{a}_i-\mathbf{b}_j)^{2}
%  \label{eq:dist_origin}
% \end{equation}

\begin{equation}
 \min_{\mathbf{R}\in SO(3), \mathbf{t} \in \mathbb{R}^{3}}
  \sum_{\mathbf{a}_i\in T({\mathcal {A}})}\sum_{\mathbf{b}_j\in {\mathcal{\bar{B}}}}(\mathbf{b}_j - \mathbf{R}\mathbf{a}_i - \mathbf{t})^{2}
 \label{eq:ego_motion}
\end{equation}

\noindent where $\mathbf{R}\in SO(3)$ and $\mathbf{t} \in \mathbb{R}^{3}$ are the unknown rotation and translation in model $T$, from which the egomotion is determined. Intuitively, the precision of point set registration heavily relies on the accuracy of point association. If association is noisy, the subsequent transformation will be inaccurate which will cause egomotion estimation to rapidly degrade over time.

%!TEX root = ../main.tex

\section{Overview} % (fold)
\label{sec:overview}

\subsection{Problem Statement} % (fold)
\label{sub:problem_statement}

In this work, we consider the general problem of multi-modal odometry, taking as inputs raw data from different types of sensors. An arbitrary sensor input is represented as $\mathcal{X_s}=\{\mathbf{x}_i\}_{i=1}^{K}$, where $K$ represents the size/length of the data points. 
In the case of \sysname, we consider a collection $\mathcal{X}=\{\mathcal{X}_M, \mathcal{X}_I\}$ where the data are collected by a mmWave radar ($M$) and an inertial sensor ($I$). However, note that our framework is able to directly generalize to a three sensor case, e.g. $\mathcal{X}=\{\mathcal{X}_M, \mathcal{X}_I, \mathcal{X}_V\}$ when a camera ($V$) is also available. The underlying relative translation $\mathbf{t}$ and rotation $\mathbf{r}$ between a pair of frames is observed by a moving platform\footnote{For the ease of model learning and fast convergence, we follow the practice of \cite{saputra2020deeptio} and represent the rotation $\mathbf{r}$ as an Euler angle rather than a rotation matrix $\mathbf{R}$ as discussed in \sect{\ref{sub:point_set_registration}}.}. The system goal is to estimate the platform's 6-DoF egomotion $\mathbf{y}=[\mathbf{t}, \mathbf{r}]$. This estimation process is equivalent to learning an effective mapping $\mathbf{x} \rightarrow f(\mathbf{x})$ so as to minimize the error between the prediction $f(\mathbf{x})$ and the ground truth $\mathbf{y}$. Notably, the output of the egomotion model is the 6-DoF relative pose, computed over a pair of consecutive samples.  For this reason, we follow \cite{wang2018end} and leverage a SE(3) composer to incrementally stitch a series of relative poses into global \emph{odometry} poses, i.e., the 6-DoF trajectory. As stated in \sect{\ref{sub:experimental_setting}}, the performance evaluation will be mainly based on the trajectory tracking.

\subsection{System Overview} % (fold)
\label{sub:methodology_overview}

From a system-level perspective, \sysname consists of multi-modal mmWave radar and inertial sensing which is then input to algorithms for egomotion estimation; this multi-modal framework is flexible to extend a third sensor in fusion as shown in Fig.~\ref{fig:milliEgo_overview}. 
The main contribution of \sysname is the end-to-end trainable deep learning model to estimate the platform egomotion. Two novel submodules are critical to the model efficacy that address different design challenges respectively: (i) the mmWave Subnet and (ii) multi-modal fusion module. 
In what follows, we will first describe the overall network architecture of the proposed  \sysname in \sect{\ref{sec:neural_network_architecture}}.

%\noindent \textbf{Multi-modal Sensing}. This module serves as the frontend, in which a mmWave radar and other sensors collect the data during the mobile platform movement. As this component is essentially a practical implementation, we will describe the detail later in \sect{\ref{sec:implementation}}.

% Given the collected mmWave point clouds, this proposed module estimates the relative transformation (i.e., odometry) between a pair of point clouds with mmWave radar only. Instead using the error-prone two-way registration approach, we propose a novel deep neural network that simultaneously models the point corresponding and transformation optimization by the end-to-end odometry learning.

% \noindent \textbf{Multi-modal Odometry Fusion}. The last module is proposed to push the estimation robustness and combines the mmWave radar with other on-board sensors. By stacking deep odometers of distinct sensors (including mmWave), this generic fusion module learns hybrid attention masks that can adaptively select deep features within and across sensors to deal with different conditions. Notably, this fusion framework can also be end-to-end learned while the mask weights provide model interpretation.

% As shown in Fig.~\ref{fig:milliEgo_overview}, our proposed systems consists of three modules, including Multi-modal Sensing (2) Deep mmWave Odometry and (3) Multi-modal Odometry Fusion. The contribution of this work mainly lies into the latter two.
% section overview (end)

%!TEX root = ../main.tex

\section{Neural Network Architecture} % (fold)
\label{sec:neural_network_architecture}

As shown in Fig.~\ref{fig:milliEgo_overview}, the proposed DNN egomotion model starts with multiple egomotion subnets coming from different sensor sets. These multi-modal egomotion features are then fused by a specific fusion module in the DNN. A recurrent neural network models the temporal dependency of the fused features and forward it to fully-connected layers. These layers learn to regress the final features to a pose supervised by the loss function of 6-DoF estimation. 

For readability, we focus on the basic DNN architecture in this section, before introducing the \emph{mmWave Subnet} in \sect{\ref{sec:deep_odometry_model}} and the \emph{multi-modal fusion module} in \sect{\ref{sec:multi_modal_odometry}} .

\noindent \textbf{Multi-modal Egomotion Subnets}.
Besides our proposed mmWave subnets, the DNN model also has egomotion subnets from IMU and other sensors depending on their availability (see Fig.~\ref{fig:milliEgo_overview}). In particular, we use the IONET structure \cite{chen2018ionet} to develop the egomotion subnet for IMU and DeepVO \cite{wang2017deepvo} structure to develop subnets for RGB and depth cameras. These subnets output their egomotion features to the mixed-attention sensor fusion block. 

\noindent \textbf{RNN Temporal Dependency Modeling}. 
On receiving the fused multi-modal features, an RNN is used to model the long-term dynamics.
A prominent advantage of using an RNN for egomotion estimation lies in its superior ability in maintaining the memory of hidden states over long term even if the given inputs are noisy. The issue associated with RNNs, however, is that they cannot directly model high-dimensional raw sensory data such as images. But thanks to the CNN extractors, the data fed to RNN are already compressed in a low-dimensionality space. In particular, the RNN used is a two-layer Long Short-Term Memory (LSTM), each of which contains $512$ hidden units. This RNN module is on top of the CNN feature extractor layer.

\noindent \textbf{FC Pose Regressor}. 
Given the outputs of RNNs, three fully connected (FC) layers are introduced to regress these sequential features and predict relative transformation. These FC layers have $128$, $64$ and $6$ units respectively, with the last layer regressing features into a $6$-DoF relative pose estimate: $\mathbf{\hat{y}}=[\mathbf{\hat{t}}, \mathbf{\hat{r}}]$, where $\mathbf{\hat{y}} \in \mathbb{R}^6$. To promote model generalization ability, we also use a dropout \cite{dahl2013improving} rate of $0.25$ between FC layers to help regularization.

\noindent \textbf{Loss Function}.
Putting it together, deriving the parameters of our network model requires minimizing the following Mean Squared Error (MSE) loss:
\begin{equation}
\label{eq:loss_func}
	\mathcal{L} = \frac{1}{K} \sum_{k=1}^{K} ||(\mathbf{\hat{t}}_k-\mathbf{t}_k)||^2_2 + \gamma  ||(\mathbf{\hat{r}}_k-\mathbf{r}_k)||^2_2
\end{equation}
\noindent where $\norm{\cdot}$ denotes $L2$-norm and $\gamma$ is a hyper-parameter to balance the weights of translation and rotation errors. Training details of the model are provided in \sect{\ref{sec:implementation}}.

%!TEX root = ../main.tex
\section{Deep mmWave Egomotion Subnet} % (fold)
\label{sec:deep_odometry_model}

Although deep models for inertial and visual egomotion \cite{chen2018ionet,wang2017deepvo,clark2017vinet} are well established and can be reused, mmWave based egomotion remains an open area and to the best of our knowledge has not been considered from a data-driven perspective. As such, we will first elaborate on the technical challenges created by the single-chip mmWave radar and then introduce our end-to-end solution.

% In this section, we explore the feasibility of using a low-cost mmWave radar for egomotion estimation, answering the question: \emph{is it possible to build up an usable odometry from the very sparse and noisy mmWave point clouds?}. Not only can this understanding allow the design of a standalone mmWave odometry, but provide insights for more effective sensor fusion in \sect{\ref{sec:multi_modal_odometry}}. In what follows, we start by introducing the technical challenges brought by low-quality point clouds and then describe our proposed deep learning based odometry model to address them. 

\begin{figure}[!t]
	\centering
	% \begin{subfigure}[b]{0.26\textwidth}\centering
	% 	\includegraphics[width=\columnwidth]{figures/pcl/lidar_before.pdf} 
	% 	\caption{Lidar: Before Reg.}
	% 	\label{fig:lidar_before}
	% \end{subfigure}%
	\begin{subfigure}[b]{0.26\textwidth}\centering
		\includegraphics[width=\columnwidth]{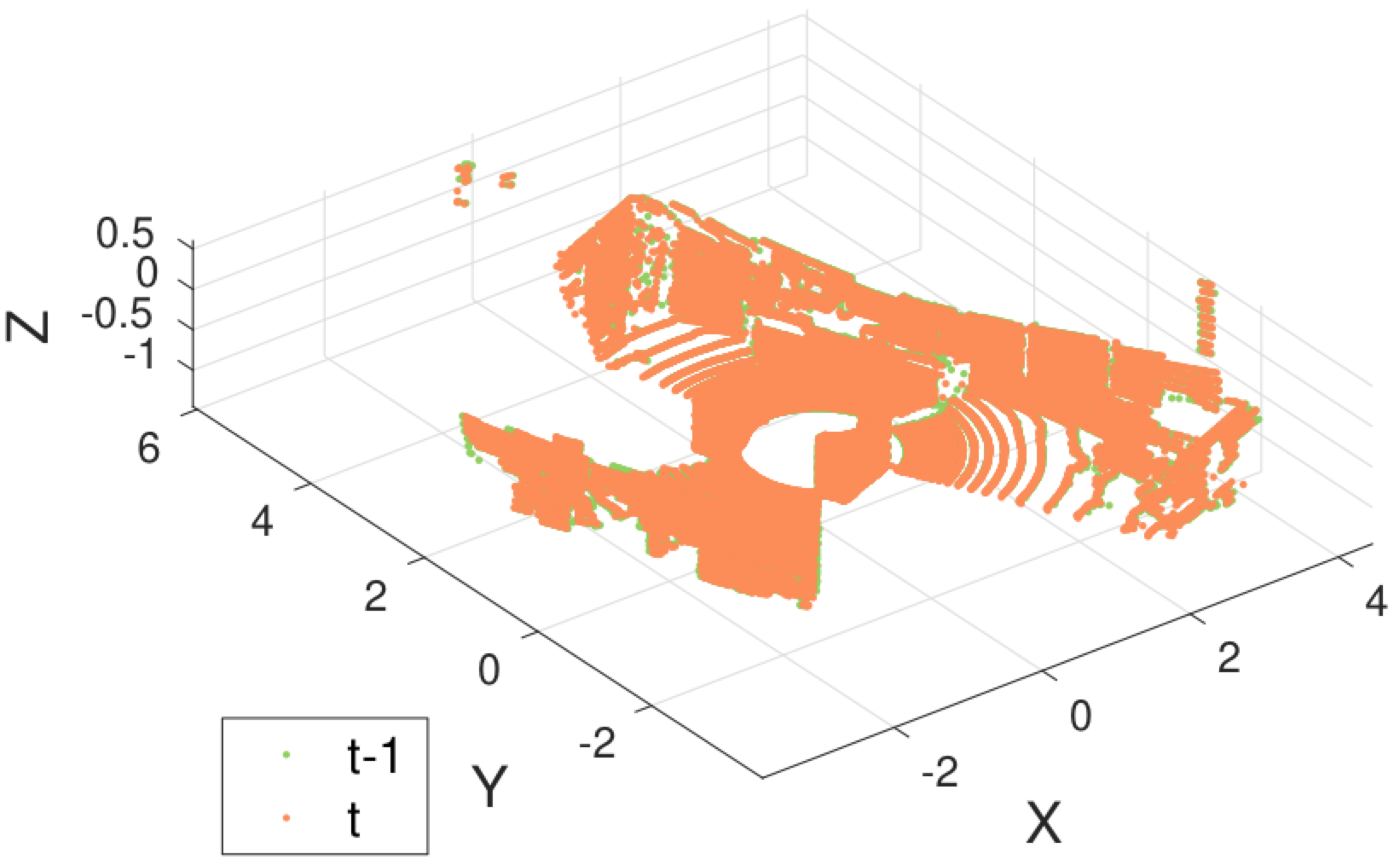} 
		\caption{Lidar}
		\label{fig:lidar_after}
	\end{subfigure}%
	% \begin{subfigure}[b]{0.22\textwidth}\centering
	% 	\includegraphics[width=\columnwidth]{figures/pcl/mmwave_before.pdf} 
	% 	\caption{mmWave: Before Reg.}
	% 	\label{fig:_before}
	% \end{subfigure}%
	\begin{subfigure}[b]{0.23\textwidth}\centering
		\includegraphics[width=\columnwidth]{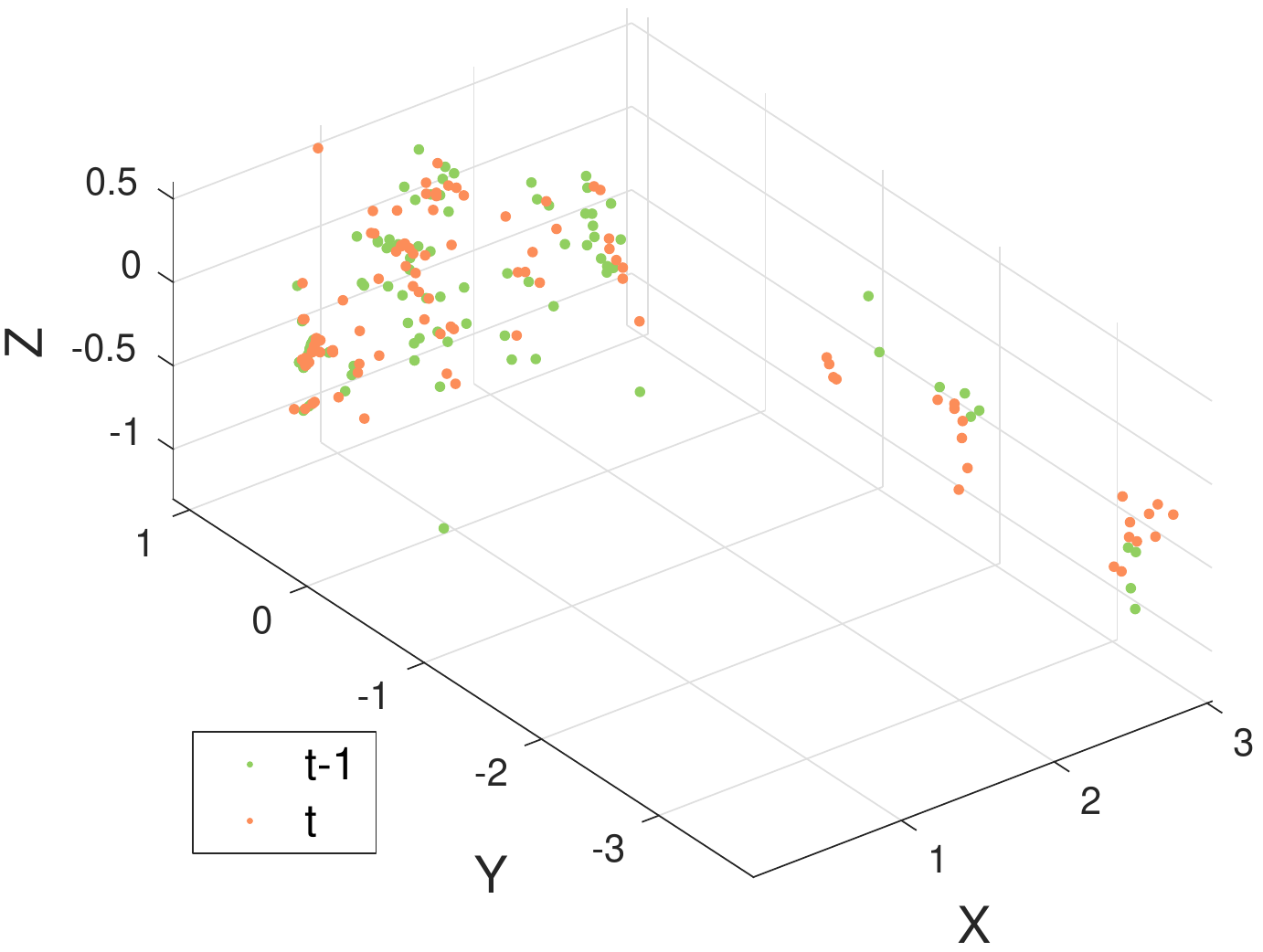} 
		\caption{mmWave}
		\label{fig:mmwave_after}
	\end{subfigure}%
\caption{Comparison between registering lidar and mmWave point clouds (captured at the same instants) with ICP, a commonly used technique. The effectiveness of registration is evidenced by the level of overlap between two consecutive point clouds, shown in orange and green. (a) shows how the dense lidar data can be robustly aligned. In contrast (b) shows  that the sparse and noisy mmWave radar point clouds cannot be robustly aligned due to insufficient correspondence. This figure is best viewed by zooming in.}
\label{fig:icp_fail}
\vspace{-0.4cm}
\end{figure}

\subsection{Challenge: Noisy Correspondence} % (fold)
\label{sub:chaotic_point_correspondences}

The fundamental modeling challenge associated with the use of single-chip radar is with their noisy and sparse point clouds e.g. containing tens of points rather than the thousands as typically acquired by lidar or depth sensors. As a result it is very challenging finding reliable transformations between consecutive frames when using traditional point matching methods~\cite{gold1998new}.  

\subsubsection{Why does the challenge arise?} % (fold)
\label{ssub:why_is_the_challenge_}

As described in Sec.~\ref{sub:point_set_registration}, the quality of point set registration between consecutive point clouds underpins accurate egomotion estimation. Critical to most registration techniques is the determination of correspondence between spatially localized 3D points within each cloud. Such a registration problem is hard to solve when facing issues of noise, missing or spurious data \cite{gold1998new}, all of which are characteristic of the point cloud generated by the single-chip mmWave radar. In particular, the two main limitations are (1) multi-path effects and (2) sparsity, both of which are discussed as follows.

\noindent \textbf{Multi-path Noise:}
Multi-path noise is a longstanding issue for almost all RF technologies. Signal propagation of mmWave radar in indoor environments is subject to multi-path \cite{yan2016multipath} due to beam spreading, diffraction and reflection from surrounding objects. As a result, reflected signals arriving at a receiver antenna are often from two or more paths, incurring smearing and jitter. the primary contributor to the non-negligible proportion of noise artefacts and spurious data in the mmWave point cloud. 

\noindent \textbf{Sparsity:}
On top of the strong noise, mmWave point clouds also have severe sparsity issue, owing to (1) the fundamental \emph{specularity} of mmWave signals and (2) the low-cost single-chip design. Wireless mmWave signals are highly specular i.e., the signals exhibit mirror-like reflections from objects \cite{lu2013measurement}. As a result, not all reflections from the object propagate back to the mmWave receiver and major parts of the reflecting objects do not appear in the point cloud. Moreover, as opposed to massive MIMO radar technologies, the mmWave radar in this case only has $3\times4$ antennas which is effective in both cost and size but greatly limits the ability to resolve spatially close targets. Moreover, in order to lower bandwidth and improve signal-to-noise ratio, algorithms such as CFAR (Constant False Alarm Rate) \cite{ward1969handbook} are used for data processing and \emph{only} provide an aggregated point cloud, further reducing density. For these reasons, the resulting point clouds only contain approximately $\sim100$ reflective points per frame which is over $100\times$ sparser than a lidar. 

\subsubsection{The failure of traditional methods} % (fold)
\label{ssub:what_influence_on_traditional_methods_}

The multi-path noise and sparsity issues discussed above collectively impact the performance of conventional geometry based methods. Voting-based registration methods are rendered ineffective \cite{huttenlocher1990recognizing} through extreme sparsity. The other established class is the Iterated Closest Point algorithm (ICP) and its variants, achieving greater robustness by iteratively interleaving correspondences matching and transformation estimation. Nonetheless, ICP methods require a good initial estimate which is typically achieved with  Random sample consensus (RANSAC) \cite{hu2012robust} and other bootstrapping methods. Again, due to the sparsity and lack of resolution for feature extraction, these algorithms perform poorly, exacerbating convergence to local rather than global minima  \cite{bouaziz2013sparse}. 
Fig.~\ref{fig:icp_fail} compares the performance of using RANSAC-ICP algorithm to register lidar and mmWave point clouds, both captured at the same time from the same view. Firstly, it is immediately apparent how much denser the lidar point cloud is, accurately representing 3-D scene structure. Secondly, although the lidar point clouds are accurately registered, the mmWave radar point clouds cannot be accurately registered as seen by the lack of alignment. Experimentally, we observe that pose estimation error is $\sim 9$-fold higher for mmWave radar compared with lidar. The direct consequence of such poor and ineffective registration is degraded egomotion estimation, rapidly causing the global trajectory to drift.

%In fact, the direct consequence of such ineffective registration is a poor egomotion estimation. As illustrated in Fig.~\ref{fig:icp_error_logs}, the ICP estimation error per frame with mmWave data is $\sim 9$-fold than the lidar counterparts. \sect{\ref{sub:mobile_robot_performance}} will provide more evaluation comparison with traditional methods.

% \begin{figure}[!t]
% 	\centering
% 	\includegraphics[width=0.75\columnwidth]{figures/pcl/err_compare.pdf} 
% 	\caption{The consequent large egomotion estimation error of ICP per frame due to the poor point set registration (refer to Fig.~\ref{fig:icp_fail}).\acm{Not convinced this figure is necessary - it might be easier just to mention in the text and be done.}}
% \label{fig:icp_error_logs}
% \end{figure}

% \begin{figure*}[!t]
% 	\centering
% 	\includegraphics[width=0.8\textwidth]{figures/mmwave_subnet.pdf} 
% \caption{The architecture of mmWave Subnet. Notation - \emph{Channels@Height$*$Width}. By stacking two consecutive panoramic mmWave images (each with a dimension of $3$@$64*256$) as the input, the subnet consists of convolution layers with different kernel sizes, 
% followed by LeakyReLU non-linearity activation everytime. The latent features are finally flattened and serve as the mmWave feature map for subsequent modules.}
% \label{fig:mmwave_subnet}
% \end{figure*}

\subsection{mmWave Sub-network} % (fold)
\label{sub:mmwave_odometry_sub_network}

\noindent \textbf{Motivation.}
Addressing the challenge of noisy correspondences requires more sophisticated estimation algorithms beyond simple heuristics. We thus propose to use data-driven deep learning based approaches which have proven to be effective in extracting useful motion information from complex sensor observations.
Critical to this design choice is that deep learning approaches allow for end-to-end modeling and treat the challenging mmWave egomotion problem as a subnetwork or a branch in the full neural network model. Our key insight is that \revise{instead of a traditional two-step registration process, it is possible to use a deep neural network that jointly optimizes the transformation estimation and implicit correspondence association between consecutive frames. This end-to-end learning approach helps the system effectively counter the above challenges.} 
Therefore, the knock-on effect between the interleaving correspondence association and transformation estimation is mitigated. Furthermore, the end-to-end design also allows us to readily integrate odometry branches from other sensors, since they are treated as other subnets in the full network.

\noindent \textbf{mmWave Subnet Inputs.} 
Determining an effective input format to the mmwave subnet is non-trivial. Point cloud data are unordered and irregularly sampled in 3D space. Extracting the structured motion patterns from such unstructured data is already difficult, not to mention the issues with ambiguous ordering due to severe sparsity and noise. This is likely the reason why existing networks designed for motion flow extraction often require structured inputs (e.g., \cite{ye20183d,dosovitskiy2015flownet}). To combat the impact of lack of data structure and irregularity, we encode mmWave point clouds into 2D panoramic-view images, following the convention of \cite{li2016vehicle}: 
\begin{equation} \label{eq1}
	\begin{split}
	\alpha & = arctan2(p_y,p_x),  \quad r = \lfloor \alpha / \Delta\alpha \rfloor \\
	 \beta & = arcsin(p_z/\sqrt{p_x^2+p_y^2+p_z^2}), \quad c = \lfloor \beta / \Delta\beta \rfloor 
	\end{split}
\end{equation}
\noindent where $(p_x, p_y, p_z)$ denotes a 3D point and $(r, c)$ is the 2D map position of its projection. $\alpha$ and $\beta$ denote the azimuth and elevation angle when observing the point. $\Delta\alpha$ and $\Delta\beta$ are the average horizontal and vertical angle resolution between consecutive beam emitters respectively. After panoramic projection, we further normalize values to the range $[0, 255]$. Points closer to the sensor are assigned higher values. Through this transformation, unstructured 3D point clouds are encoded to structured 2D \emph{`depth'} images, in a similar spirit to RGB images which are tractable for many established DNN feature extractors. One key observation we found in this work is that, with right DNN, meaningful egomotion features can be learned from such \emph{`depth'-like} mmWave input representations. 

\noindent \textbf{CNN Feature Extractor}. Given the subnet inputs, the next step is to design a feature extractor to obtain useful mmWave features for motion estimation. As the input mmWave data have been converted into the format of images at this point, we adopt a CNN structure to process it and extract egomotion features. Notably this design rationale is in a similar spirit of \cite{dosovitskiy2015flownet}, in which implicit correspondences (or the optical flow) between a pair of RGB images was estimated by a CNN. 
% Fig.~\ref{fig:mmwave_subnet} illustrates the architecture information of mmWave Subnet. 
Taking as inputs two consecutive panoramic images ($\mathbf{x}_{k-1}^M$ and $\mathbf{x}_{k}^M$), our extractor comprises $9$ convolutional layers and each layer is followed by a Leaky version of a Rectified Linear Unit (LeakyReLU) non-linearity activation. The sizes of the receptive fields in the network gradually reduce from $7 \times 7$ to $5 \times 5$ and then $3 \times 3$ to incrementally capture smaller features. The extracted latent mmWave feature representation not only compresses the original high-dimensionality panoramic data into a compact description, but facilitates the temporal dependency modeling. When there is only one mmWave radar in \sysname, the extracted features from the mmWave subnet are used as a branch to be fused with other modalities. However, we will demonstrate the effectiveness of this architecture in enabling mmWave-only odometry in \sect\ref{ssub:potential_impact_of_multiple_mmwave_radars}, where multiple mmWave radars are co-located on a mobile platform. 

\section{Multi-modal Egomotion Fusion} % (fold)
\label{sec:multi_modal_odometry}

\begin{figure}[!t]	
	\centering
	\includegraphics[width=\columnwidth]{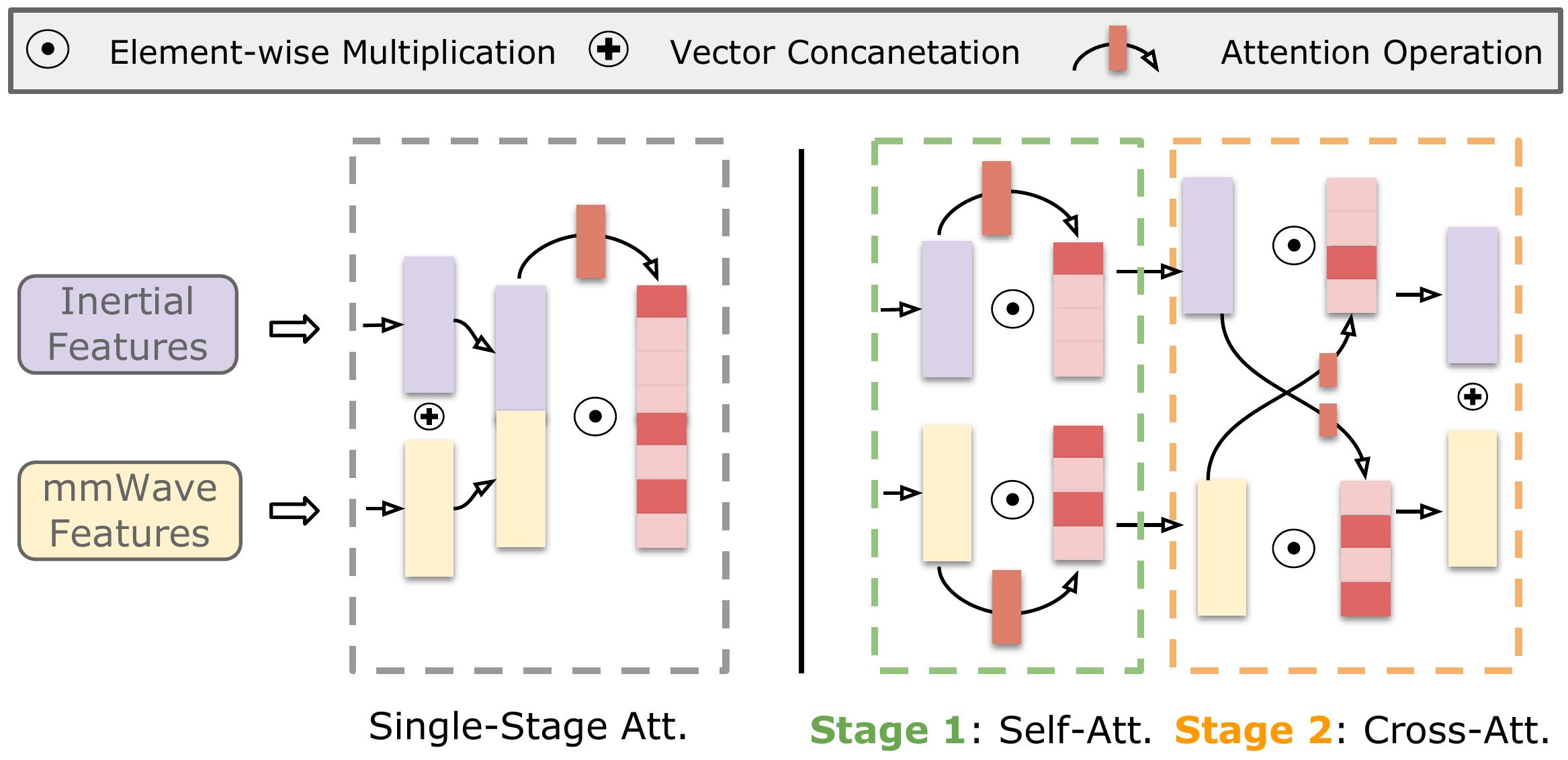} 
\caption{Architectural comparison between single-stage attention mechanism (e.g. \cite{chen2019selective,saputra2020deeptio}) and our proposed mixed (two-stage) attention mechanism.}
\label{fig:attention_mech}
\end{figure}

So far we have introduced the egomotion estimation subnets of mmWave radar and other sensors. The question remains is how to fuse the multi-modal egomotion features such that the model can adaptively cope with different situations by using the `right sensors' at the right time. %This requires the \emph{complementary behaviors} between mmWave radars and other sensors that are non-trivial to achieve. Again, we start with the broad challenge and then describe our solution. 

\subsection{Challenge of Adaptive Fusion} % (fold)
\label{sub:challenge_of_adaptive_fusion}

A robust multi-modal egomotion estimation is based on the premise of complementary interactions amongst different sensors, whose roles are adaptively altered or re-weighted in response to observation uncertainties or self/environmental dynamics. For example, the mmWave sensor should dominate egomotion estimation when the IMU is degraded in performance. Although classical fusion methods can realize such an adaptation by incorporating physical models into the algorithm design (e.g., Bayesian filtering or fixed-lag smoothers) they require expert design. However,  adaptive fusion becomes much harder in the deep feature space. This challenge is rooted in the long-standing criticism of DNN: it is difficult for one to manually incorporate an adaptation mechanism into a black box model due to the complex multilayer nonlinear structure in DNN \cite{agrawal2016learning}.

% It has been found that \cite{yao2019sadeepsense,chen2019selective}, owning to the dynamic observation quality, treating different odometry sensors evenly in a deep network often results in sub-optimal estimation and lacks adaptation against environment uncertainty. 

\subsection{Mixed Attention for Fusion} % (fold)
\label{sub:mixed_attention_for_fusion}
\revise{
To endow a multimodal deep egomotion network with the ability to adapt to environmental uncertainties, \revise{we systematically investigate the combination of different attention mechanisms in order to best exploit complementary sensor behaviors.} The intuition behind attention in general is that features are not always of equal importance and their relative contribution varies on the context. Critical to this module is to study the effectiveness of a mixed or two-stage attention mechanism. As shown in Fig.~\ref{fig:attention_mech}, mixed attention consists of two attention layers corresponding to the intra- and inter- sensor behavior regulator in our fusion framework. 
% Unlike the single-stage attention introduced in \cite{saputra2020deeptio,chen2019selective}, our model is inspired by Spence's \emph{`separate-but-linked'} cross-modal attention model for human spatial perception. The key idea behind the model is that there are separate auditory, visual, and tactile attentional systems in our cognitive processes, with subsequent links amongst them during perception \cite{spence2004crossmodal,spence1996audiovisual}. 
% Our \emph{individual} senses regularly meet and \emph{crosstalk} in the brain so as to provide accurate impressions of the world \cite{bleicher2012edges,seubert2012know}.
Our goal is to investigate whether the mixed attention framework can better address the fusion challenge than single-stage attention method and effectively learn how best to exploit the complementary nature of the individual sensors. 
In what follows, we introduce the details of the \emph{self-attention} module for individual sensors, and the \emph{cross-modal attention} module which deals with interactions across sensor modalities.}

\subsection{Intra-sensor Self-Attention} % (fold)
\label{sub:self_fusion_for_intra_feature_correlation}

% \noindent \textbf{Motivation}.
% A promising direction to enable model adaptation is letting individual odometry branches adapt themselves first, similar to a self-filtering process that autonomously sieves informative features.
% This idea conceptually falls into the technique of `self-attention', which was initially designed for machine translation and image transformer to address the long dependency issue \cite{vaswani2017attention,dou2019dynamic,parmar2018image}. Inspired by these early successes, we consider using the self-attention mechanism to realize odometry self-regulation. Particular, without resorting to the complex query-key-value pipeline in \cite{vaswani2017attention}, we adopt a space/spacetime non-local framework akin to \cite{wang2018non} which is proven to adaptively re-weighs feature importance. Note that unlike \cite{yao2019sadeepsense,chen2019selective}, we only trigger self-attention for individual odometry branches.

% \noindent \textbf{Self-attention}.
A promising way to enable model adaptation is by letting individual odometry branches adapt themselves first, similar to a self-filtering process that autonomously sieves informative features. This concept of `self-attention' was initially designed for machine translation and image transformation to address the long dependency issue \cite{vaswani2017attention,dou2019dynamic,parmar2018image}.  Inspired by these early successes, we consider using the self-attention mechanism to realize egomotion self-regulation. Without resorting to the complex query-key-value pipeline in \cite{vaswani2017attention}, we adopt a space/spacetime non-local framework akin to \cite{wang2018non} which is proven to adaptively reweight feature importance. 

For ease of illustration, we start by attending the features provided by the mmWave subnet. Given an extracted feature vector $\mathbf{z}_M$ by the CNN extractor introduced in \sect{\ref{sub:mmwave_odometry_sub_network}}), a self-attention module first computes the similarity between two embedding spaces and uses the similarity to generate an attention map:
\begin{equation}
     \mathbf{a}_M = \sigma[(\mathbf{W}_M^\varrho   \mathbf{z}_M)^{T} \mathbf{W}_M^\varphi   \mathbf{z}_M]
\label{eq:self-att}
\end{equation}     
\noindent where $\mathbf{W}_M^\varrho$ and $\mathbf{W}_M^\varphi$ are learnable weight matrices that project the original features to embedding spaces $\varrho(\mathbf{z}_M)$ and $\varphi(\mathbf{z}_M)$ respectively. $\sigma$ represents a non-linear activation function, e.g., softmax or sigmoid. After applying the generated focus mask, the attended features $\mathbf{\tilde{z}}_M$ are given by:
\begin{equation}
     \mathbf{\tilde{z}}_M = \mathbf{a}_M \odot \mathbf{z}_M
\end{equation}  
\noindent where $\odot$ denotes element-wise multiplication. From the perspective of filtering, self-attention forces the model to focus on stable and geometrically meaningful features, whilst ignoring distracting or noisy components, so as to regulate the estimation model in response to complex environment uncertainty. In this way, we can similarly generate self-regulated deep odometry features for inertial sensors ($\mathbf{\tilde{z}}_I$), RGB cameras ($\mathbf{\tilde{z}}_V$) or depth cameras ($\mathbf{\tilde{z}}_D$) depending on their availability. As introduced in \cite{wang2018non,vaswani2017attention}, through the similarity comparison between two embeddings, self-attention is able to capture the long-range dependencies and global correlations of the features. This property is of paramount importance for mmWave and optical odometry estimation, where widely separated spatial regions should jointly considered. 

\begin{figure}[!t]
		\includegraphics[width=\columnwidth]{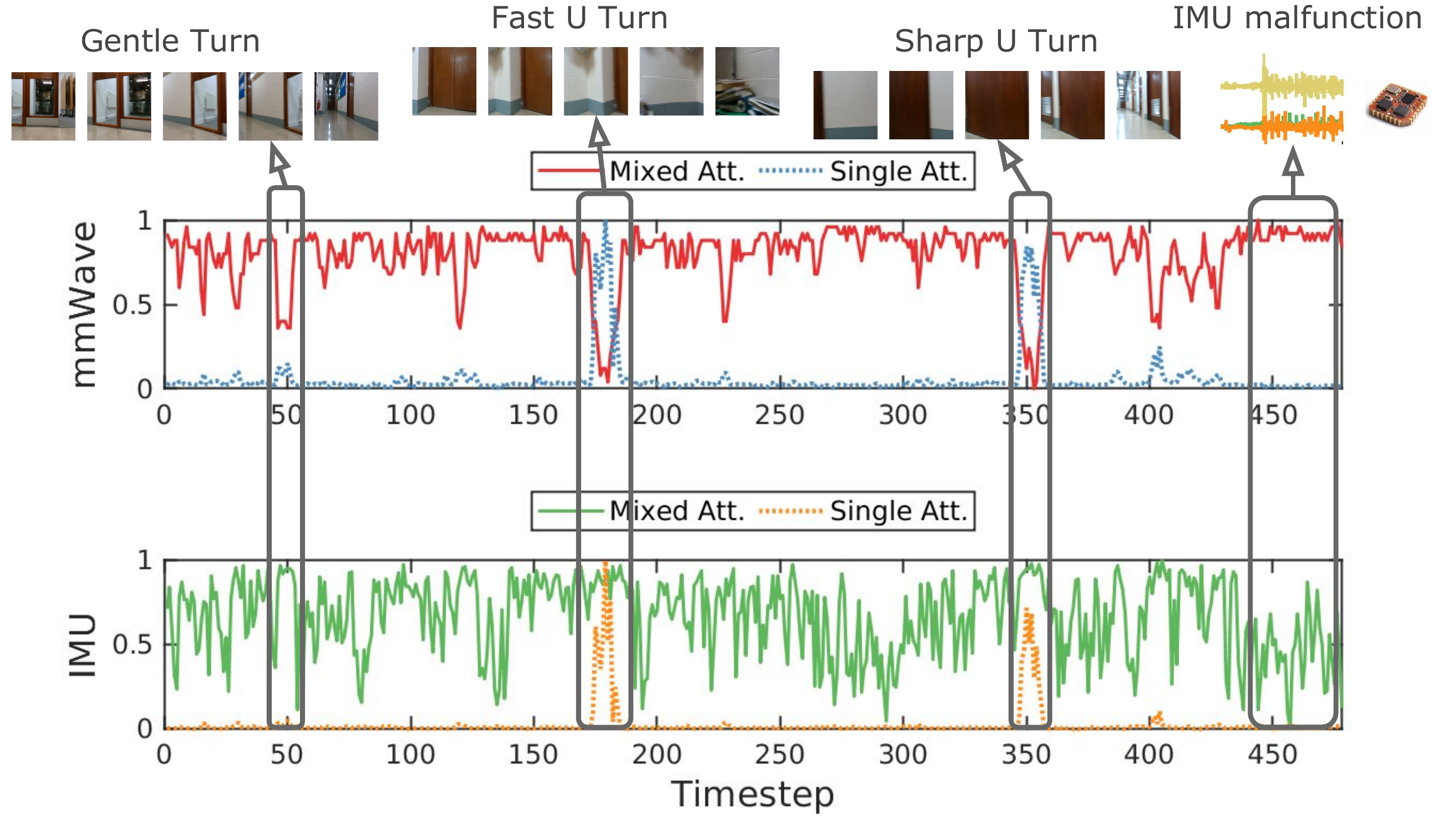} 
		\caption{Attention masks comparison between Mixed (Two-stage) and Single-stage strategy. We aggregate a mask vector by counting the ratio of its sigmoid outputs over $0.5$, and visualize the aggregated values over the sequence. Both attention weights are normalized to $[0,1]$. It can be seen that with our proposed mixed strategy, mmWave and IMU \emph{interchangeably tones down} in response to different challenging situations to them. While with the baseline single-stage method, mmWave and IMU importance change \emph{synchronously} implying an ineffective complementary fusion.}
\label{fig:mixed_attention}
% \vspace{-0.4cm}
\end{figure}

\subsection{Inter-sensor Cross-Attention} % (fold)
\label{sub:cross_modal_fusion}

The self-attention layer reweighs the importance of per-sensor descriptors in the deep feature space. But such importance masks ignore the correlation among different sensors, motivating us to move forward to inter-sensor correlations. The cross-attention serves as the second-stage of reweighting, which is inspired by the mechanism of human perception.  Analogously, we expect a robust multi-modal egomotion system to adaptively shift relative sensor influence depending on the particular scenario in question.

To give these egomotion sensors the ability to ``crosstalk'', we consider generating attention masks across sensor descriptors, where the conditioning process is akin to self-attention. Consider the case combining a mmWave radar with an inertial sensor. Given their features $\mathbf{\tilde{z}}_M$ and $\mathbf{\tilde{z}}_I$ after self-attention, two respective cross-attention masks are derived as follows:
\begin{equation}
\begin{split}
	\mathbf{a}_{I \rightarrow M} &= \sigma[(\mathbf{W}_{I \rightarrow M}^\varrho   \mathbf{\tilde{z}}_I)^{T}\mathbf{W}_{I \rightarrow M}^\varphi   \mathbf{\tilde{z}}_I] \\
	\mathbf{a}_{M \rightarrow I} &= \sigma[(\mathbf{W}_{M \rightarrow I}^\varrho   \mathbf{\tilde{z}}_M)^{T}\mathbf{W}_{M \rightarrow I}^\varphi   \mathbf{\tilde{z}}_M]
\end{split}
\end{equation}  	
where the family of $\mathbf{W}$ are learnable weight matrices. The final multi-modal egomotion descriptors can be modeled by concatenating the second-stage attended features:
\begin{equation}
	\mathbf{\bar{z}}_{MI} = [\mathbf{a}_{M \rightarrow I} \odot \mathbf{\tilde{z}}_I; \mathbf{a}_{I \rightarrow M} \odot \mathbf{\tilde{z}}_M]
\end{equation}
When more than two egomotion sensors are available, e.g., mmWave radars ($\mathbf{\tilde{z}}_M$), inertial sensors ($\mathbf{\tilde{z}}_I$) and RGB cameras ($\mathbf{\tilde{z}}_V$), we randomly leave one out and concatenate the remainder and then generate an attention mask conditioned on it. For example, generating a mmWave attention mask from RGB and inertial sensors, can be described as:
\begin{equation}
  	\mathbf{a}_{VI \rightarrow M} = \sigma[(\mathbf{W}_{VI \rightarrow M}^\varrho   \mathbf{\tilde{z}}_{VI})^T\mathbf{W}_{VI \rightarrow M}^\varphi   \mathbf{\tilde{z}}_{VI}]
\label{eq:cross-att}
\end{equation}  
where $\mathbf{\tilde{z}}_{VI} = [\mathbf{\tilde{z}}_{V}; \mathbf{\tilde{z}}_{I}]$, are the concatenated features from two sensors after self-attention. The final attended features are:
\begin{equation}
	\mathbf{\bar{z}}_{MIV} = [\mathbf{a}_{MV \rightarrow I} \odot \mathbf{\tilde{z}}_I; \mathbf{a}_{VI \rightarrow M} \odot \mathbf{\tilde{z}}_M; \mathbf{a}_{MI \rightarrow V} \odot \mathbf{\tilde{z}}_V]
\end{equation}
This completes the mixed attention. The final attended features are fed into the RNN module, as discussed in \sect{\ref{sec:neural_network_architecture}}.

% \begin{figure}[!t]
% 	\centering
% 		\includegraphics[width=0.9\columnwidth]{figures/sel_attention_mask.pdf} 
% 		\caption{Single-stage Attention Tendency.  The attention masks generated by single-stage attention change synchronously with the gyroscope change in the z-axis, implying an ineffective complementary fusion.}
% \label{fig:single_stage_attention}
% \end{figure}

\subsection{Discussion of Mixed Attention} % (fold)
\label{sub:discussion_of_mixed_attention}

In contrast to our proposed two-stage mixed attention, a natural thought, however, is why not directly concatenate the extracted features from different odometry modules into a ``big'' vector, and simply perform self-attention on it (single-stage in Fig. \ref{fig:attention_mech}). In this way, the role of the cross-attention is implicitly surrogated into a single-stage (self-)attention. We now justify why our two-stage attention is a better design choice from the perspectives of both efficacy and complexity.

\noindent \textbf{Efficacy}.
Recall the end goal of mixed-attention is to realize adaptation through complementary sensor interactions. Although \sect{\ref{ssub:impact_of_attention_strategies}} will further confirm the superiority of our method based on quantitative results, a more intuitive way to examine this is by visualizing the generated masks and validating their change in behaviour across different scenarios. As we can see in Fig.~\ref{fig:mixed_attention}, the masks of different sensors generated by single-stage attention do not indicate cross-modal complementariness and change \emph{synchronously} or simultaneously over time. By inspecting the data, we find that, the concurrent weight increases are activated following the change in gyroscope readings, rather than a complementary response for adaptation.
On the other hand, the focus maps generated by our two-stage attention strategy give rise to clear complementary behaviors. In particular, we can clearly see that how mmWave and IMU interchangeably dominate the estimation in response to different situations: mmWave tones down in response to small parallax due to turning, while IMU shifts roles to mmWave when encountering unexpected values caused by malfunction. 
We hypothesize that attention disentangling is the key to the different attention behaviors here. Again, as discussed in \sect{\ref{sub:cross_modal_fusion}}, the basis of complementary behavior is laid down by sensor individuality, and we achieve this by a two-stage attention strategy. Note that the self-attention stage also plays an important role in our design, which we will discuss more in \sect{\ref{ssub:impact_of_attention_strategies}}. In a related context, disentangling self- and cross-attention is also found to be an effective manner to detect saliency from paired inputs \cite{detone2018superpoint}. 

% Modeling the c cross-modal attention separately leads to sensor individuality, 

% The individuality of sensors can only be preserved as each of them obtains a separate focus map for its features and more accurately hints the other one. 

% Such individuality arguably lays the basis for allowing the complementary behaviors, in a similar spirit to the human perception example motivated in \sect{\ref{sub:cross_modal_fusion}}. Note that self-attention also plays an important role in our two-stage strategy. Indeed, self-attention facilitates the follow-up cross-attention because intuitively, the filtered feature information is more reliable to be conditioned on. 

\noindent \textbf{Complexity}.
We further demonstrate that our two-stage attention does not increase computation cost and actually makes it simpler to train. Take the most complicated mmWave-Inertial-Visual odometry as example, where the lengths of their extracted feature vectors as $N_M$, $N_I$ and $N_V$. Inspecting Eq.~(\ref{eq:self-att}, \ref{eq:cross-att}), it is easy to find that the total weight matrix size is $2\times(N_M+N_I+N_V)^2$ for single-stage strategy, which is same as the space complexity of our mixed attention module:
\begin{equation*}
 	\underbrace{2\times(N_M^2 + N_I^2 + N_V^2)}_\text{1. self-attention} + \underbrace{4 \times (N_MN_I + N_VN_I + N_MN_V)}_\text{2. cross-attention}
 \end{equation*} 
In fact, it has been found that by breaking learning a fat weight matrix into several smaller matrices lined up, more efficient model inference and less network over-fitting can be also obtained \cite{srivastava2015training}. The runtime experimental results in \sect{\ref{sub:system_efficiency}} also supports this claim.  

%!TEX root = ../main.tex

\section{Implementation} % (fold)
\label{sec:implementation}

For the purpose of reproducing our approach, we release our dataset, containing more than 8km of trajectories, and the source code of our system\footnote{\url{https://github.com/ChristopherLu/milliEgo}}.

\begin{figure}[!t]
	\centering
	\begin{subfigure}[b]{0.23\textwidth}\centering
		\includegraphics[width=\columnwidth]{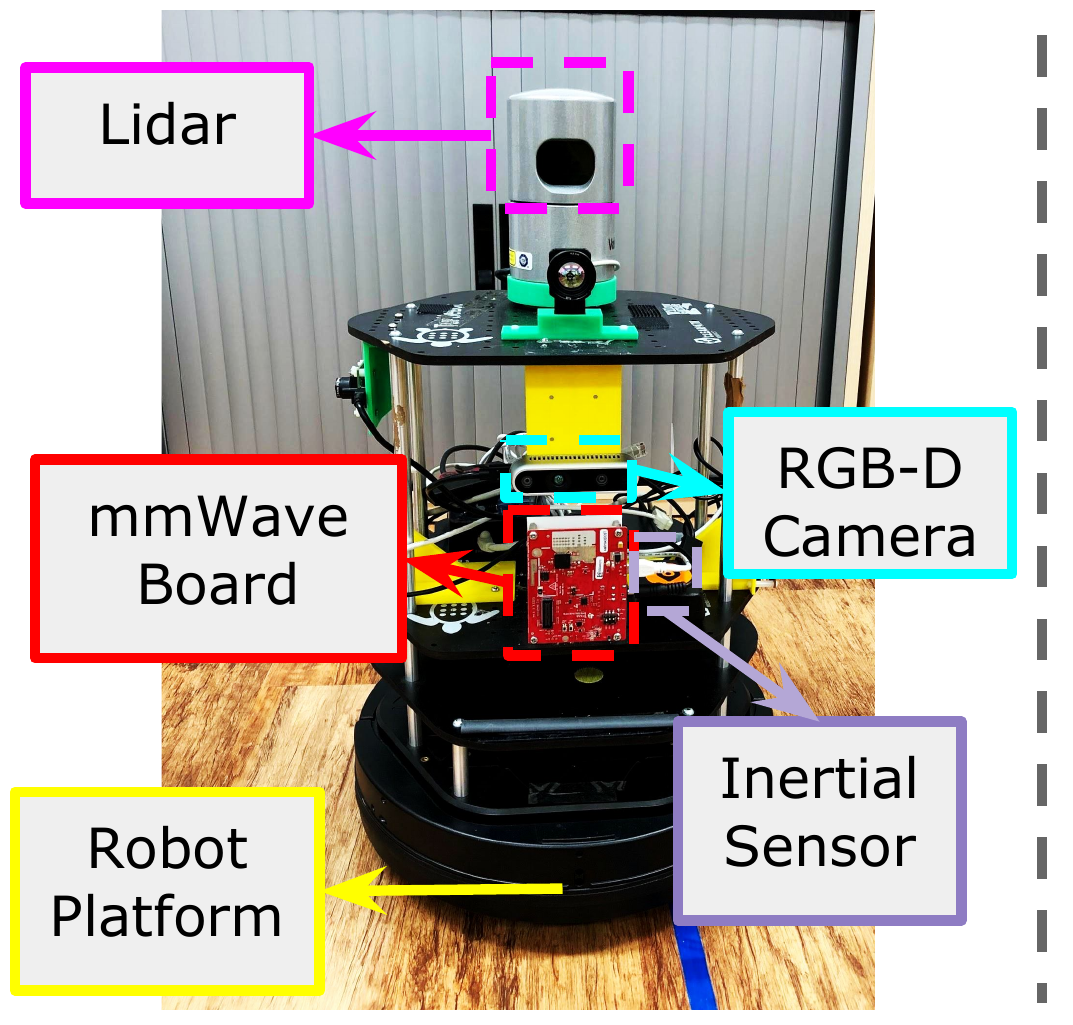} 
		\caption{Mobile Robot}
		\label{fig:robot_prototype}
	\end{subfigure}%
	\begin{subfigure}[b]{0.26\textwidth}\centering
		\includegraphics[width=\columnwidth]{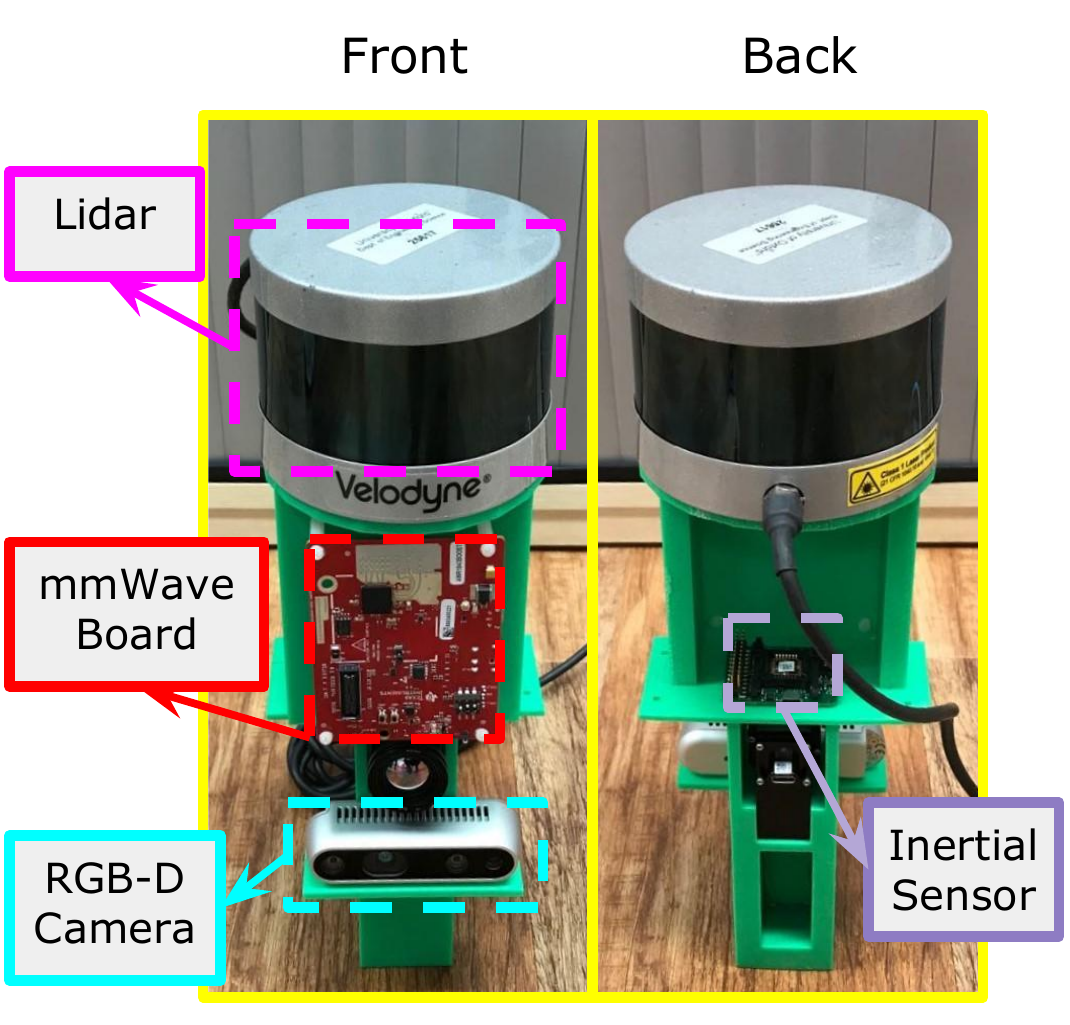} 
		\caption{Handheld}
		\label{fig:hand_prototype}
	\end{subfigure}%
\caption{Two of the prototypes. Note the bulky lidar for providing accurate ground truth.}
\label{fig:prototypes}
\vspace{-0.4cm}
\end{figure}

\noindent \textbf{Multi-modal Sensing Platform.}
We implement our multi-modal sensing system with both a mobile robot prototype and a handheld prototype. Fig.~\ref{fig:robot_prototype} shows the robot prototype, on which we equipped a Turtlebot 2 with multiple sensors including (1) a TI AWR1843 board to collect mmWave data, (2) an Xsens MTi-1 IMU for inertial measurements, (3) an Intel D435i Depth camera for both RGB and depth image capture, and (4) a Velodyne HDL-32E lidar for ground truth labeling. All the sensors are coaxially located on the robot along the vertical axis and synchronized through ROS on Turtlebot 2 \cite{quigley2009ros}. The handheld prototype is illustrated in Fig.~\ref{fig:hand_prototype}. We designed a 3D printed model that uses the same set of sensors as with the robot prototype. The only difference is the replacement of the Velodyne HDL-32E lidar with a more lightweight Velodyne Ultra Puck.

\noindent \textbf{Testbeds.}
Our testbeds have been explicitly chosen for their wide diversity, as a common concern of deep learning techniques is their potential for overfitting. 
\revise{The dataset consists of $17$ distinct floors from $6$ different multistorey buildings, including almost all accessible areas in a commercial building, including hallways, canteen, common room, building junction, atrium, office and cluttered store rooms. The smallest floor has an area of $\sim205m^2$ while the largest one has an area of more than $1500m^2$. These buildings have different internal designs and constructions and thus provide different challenges for odometry estimation, ranging from material to scale. More site descriptions can be found in our released dataset.} 
Notably, all the experimental buildings are installed with motion-triggered energy-saving lights, which causes nontrivial dynamic illumination challenges for RGB cameras. Meanwhile, glass balustrades can be found in almost all of the buildings, which is a known issue for depth cameras \cite{alt2013reconstruction}. We made such a testbed choice with the goal of investigating robust odometry and exploring if mmWave radar can be a cheap alternative for scenarios which are particularly challenging for optical sensors.

\noindent \textbf{Data Collection Procedure.} We ensure that we capture a wide diversity of trajectories, ranging from simple straight-line routes to multiple traverses of complex routes. The sampling rate of our mmWave radar is set to $20$Hz based on our empirically optimal SNR configuration on the TI AWR1843 TI board. Our final dataset contains data from the mmWave radar, IMU, RGB cameras, depth cameras and lidar. Sec.~\ref{sub:experimental_setting} introduces how collected data are split for training and testing. 

\noindent \textbf{Network Training Details.} For model training, we use RMSProp optimizer with a $1e-05$ initial learning rate, dropping by $25\%$ every $25$ epochs for a total of $200$ epochs. We normalize the input data by subtracting the mean over the dataset. The training sequence is randomly cut into small batches of consecutive pairs ($n=16$) to obtain better generalization. We also sub-sample the input frames to provide sufficient parallax between consecutive frames. We discuss the importance of this sampling strategy in \sect{\ref{ssub:impact_of_mmwave_sampling_rate}}. The regularization hyper-parameter $\gamma$ in Eq.~(\ref{eq:loss_func}) is set to $0.001$. \revise{The entire training time of \sysname is around $6$ hours on a NVIDIA Tesla V100.}

%!TEX root = ../main.tex
\section{Evaluation} % (fold)
\label{sec:evaluation}

\begin{figure*}[!t]
	\centering
	\includegraphics[width=0.93\textwidth]{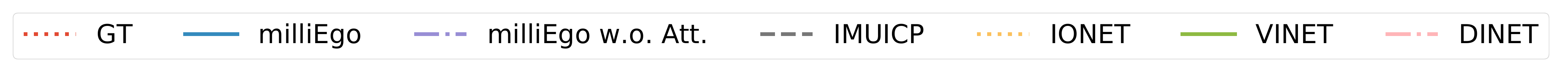}

	\begin{subfigure}[b]{0.24\textwidth}\centering
		\includegraphics[width=\columnwidth]{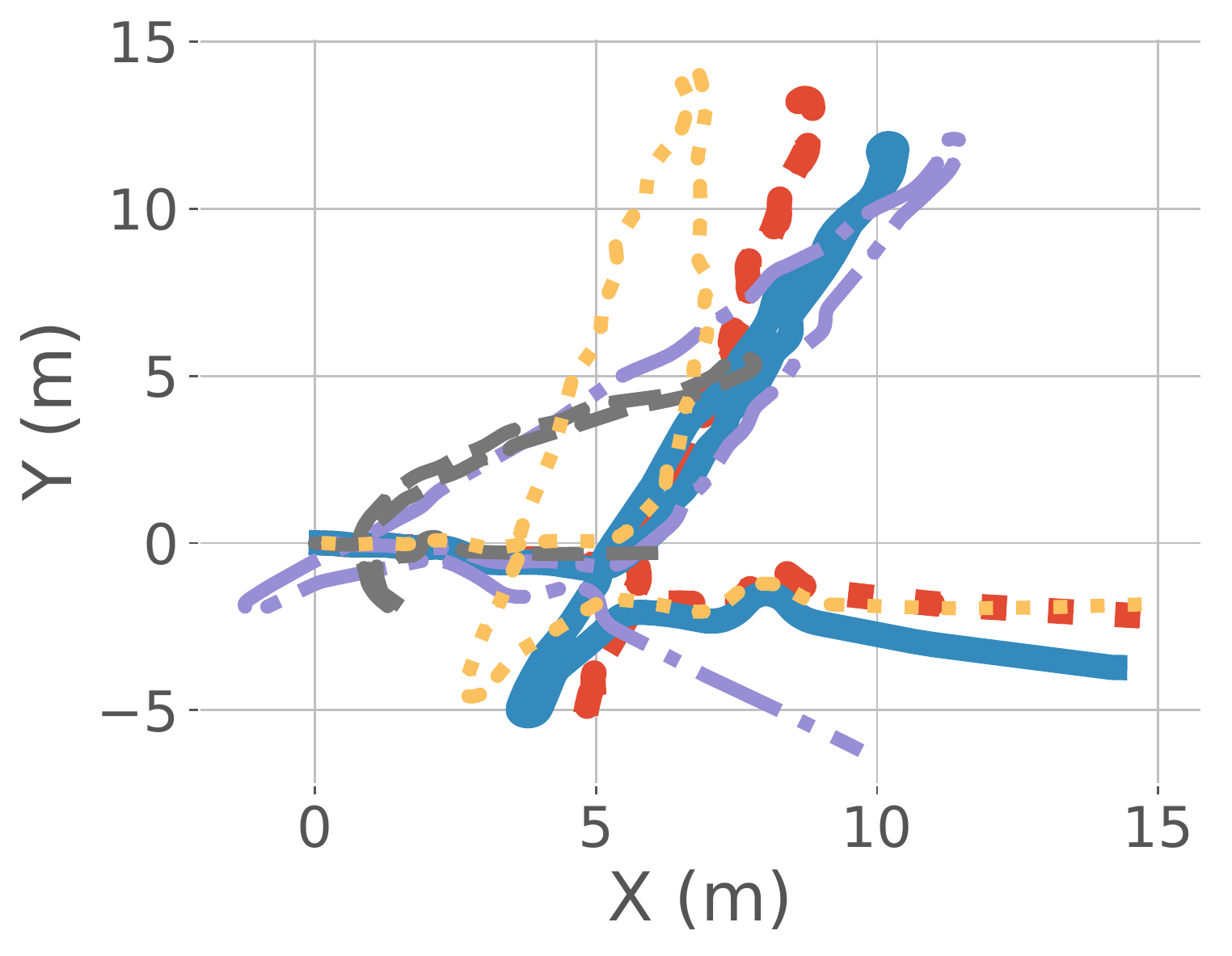} 
		\caption{Glass Corridor}
	\end{subfigure}%
	\begin{subfigure}[b]{0.24\textwidth}\centering
		\includegraphics[width=\columnwidth]{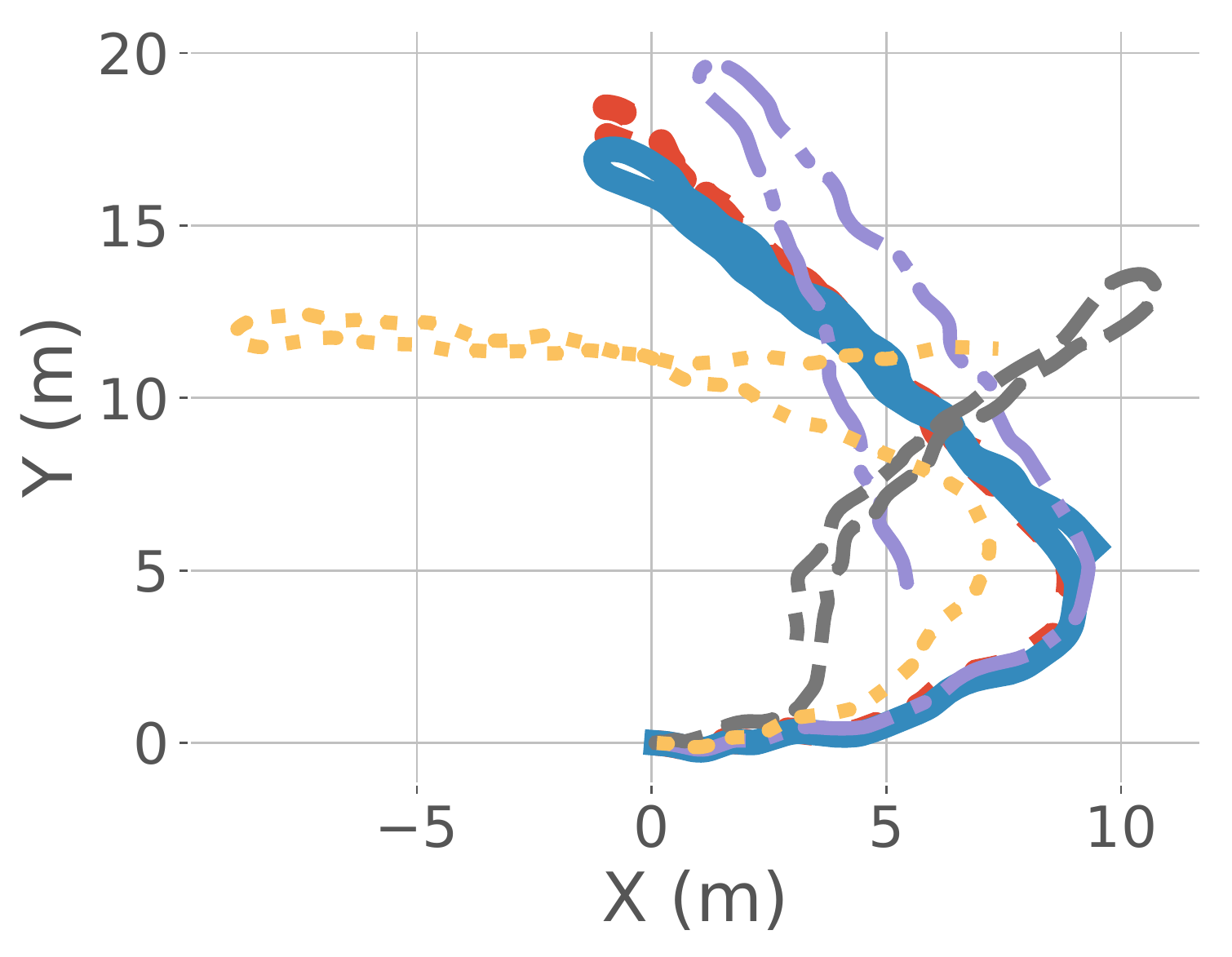} 
		\caption{Bumpy Floor}
	\end{subfigure}%
	\begin{subfigure}[b]{0.24\textwidth}\centering
		\includegraphics[width=\columnwidth]{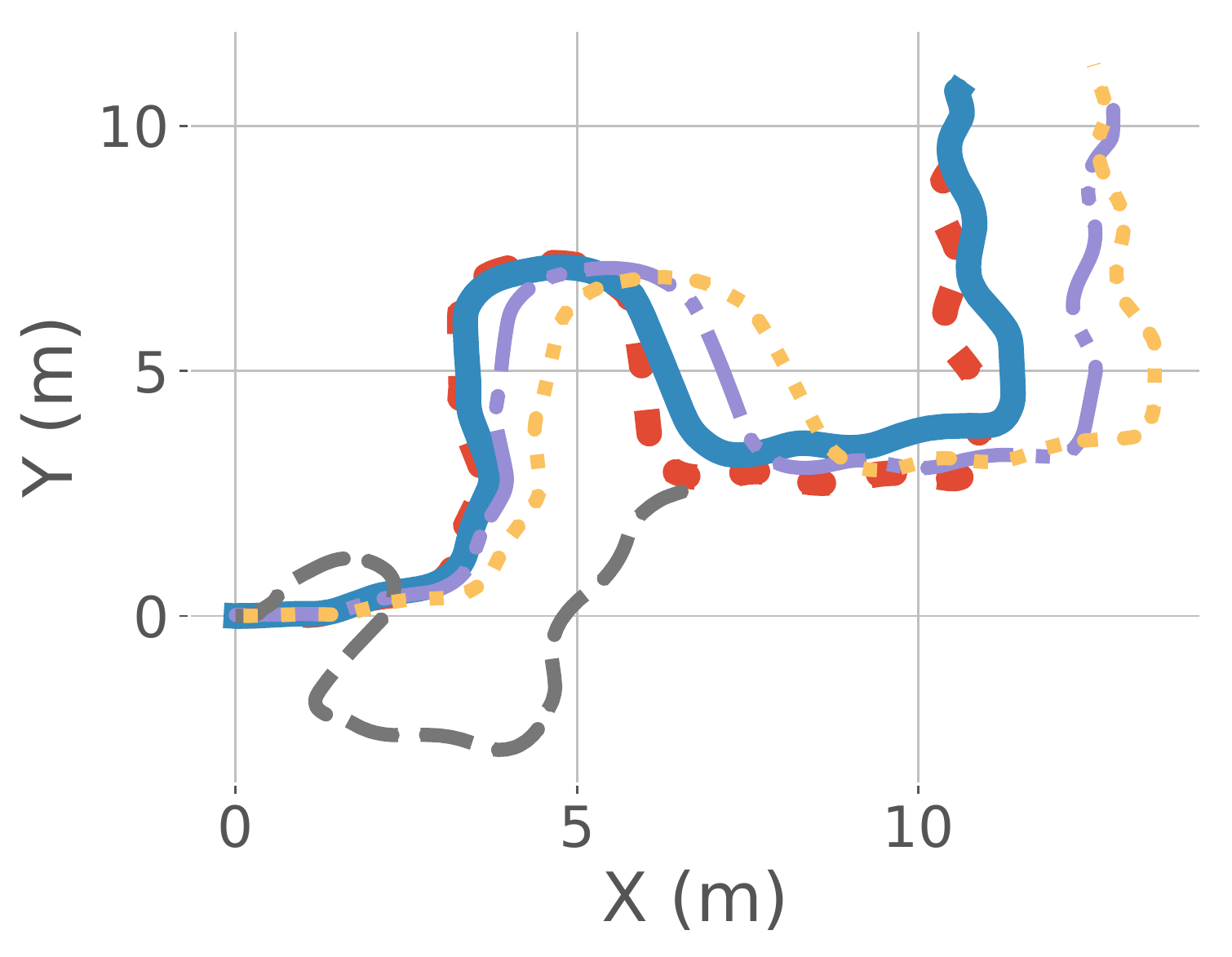} 
		\caption{Glare, Pathway}
	\end{subfigure}%
	\begin{subfigure}[b]{0.24\textwidth}\centering
		\includegraphics[width=\columnwidth]{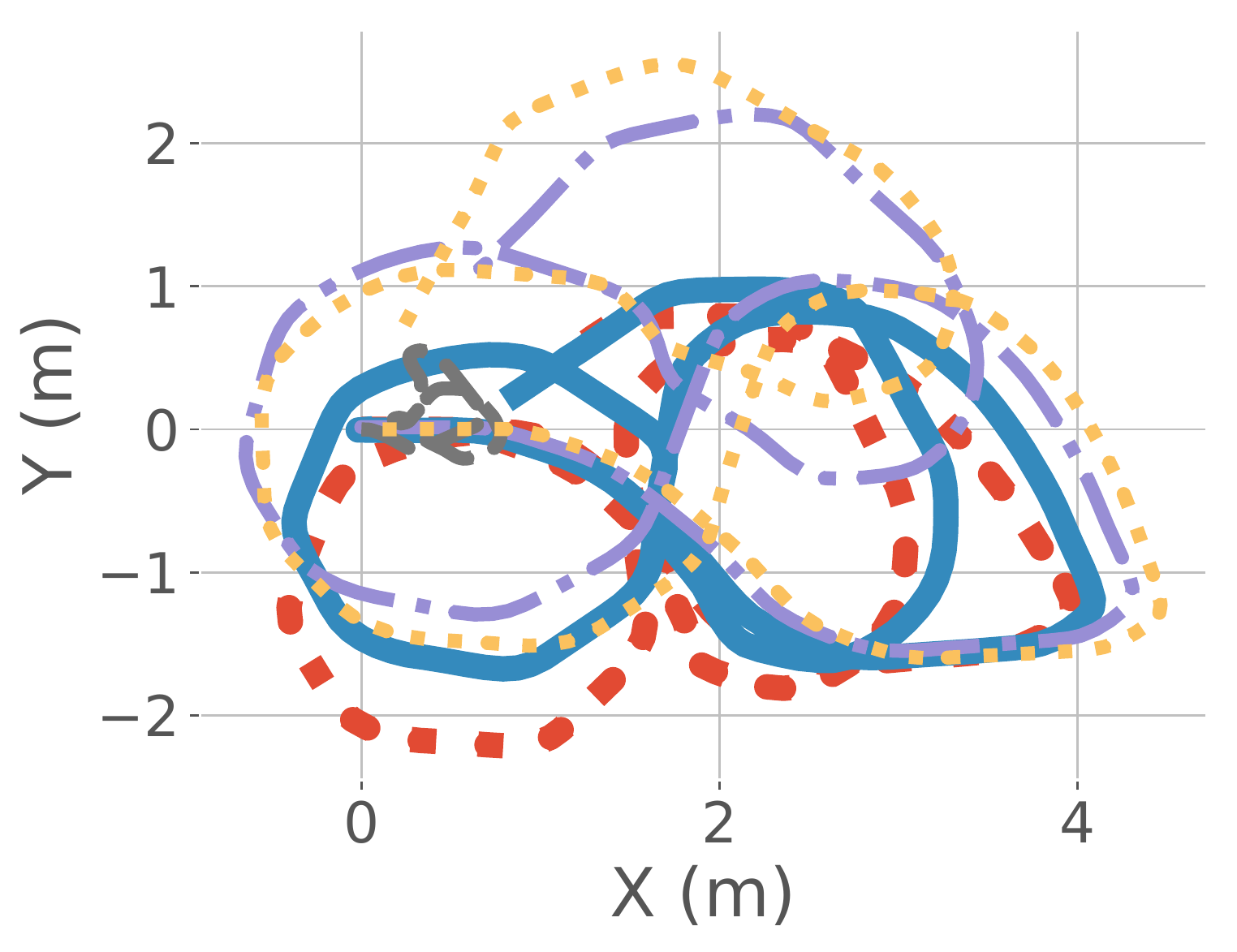} 
		\caption{Dark Room}
	\end{subfigure}%

	\begin{subfigure}[b]{0.24\textwidth}\centering
		\includegraphics[width=\columnwidth]{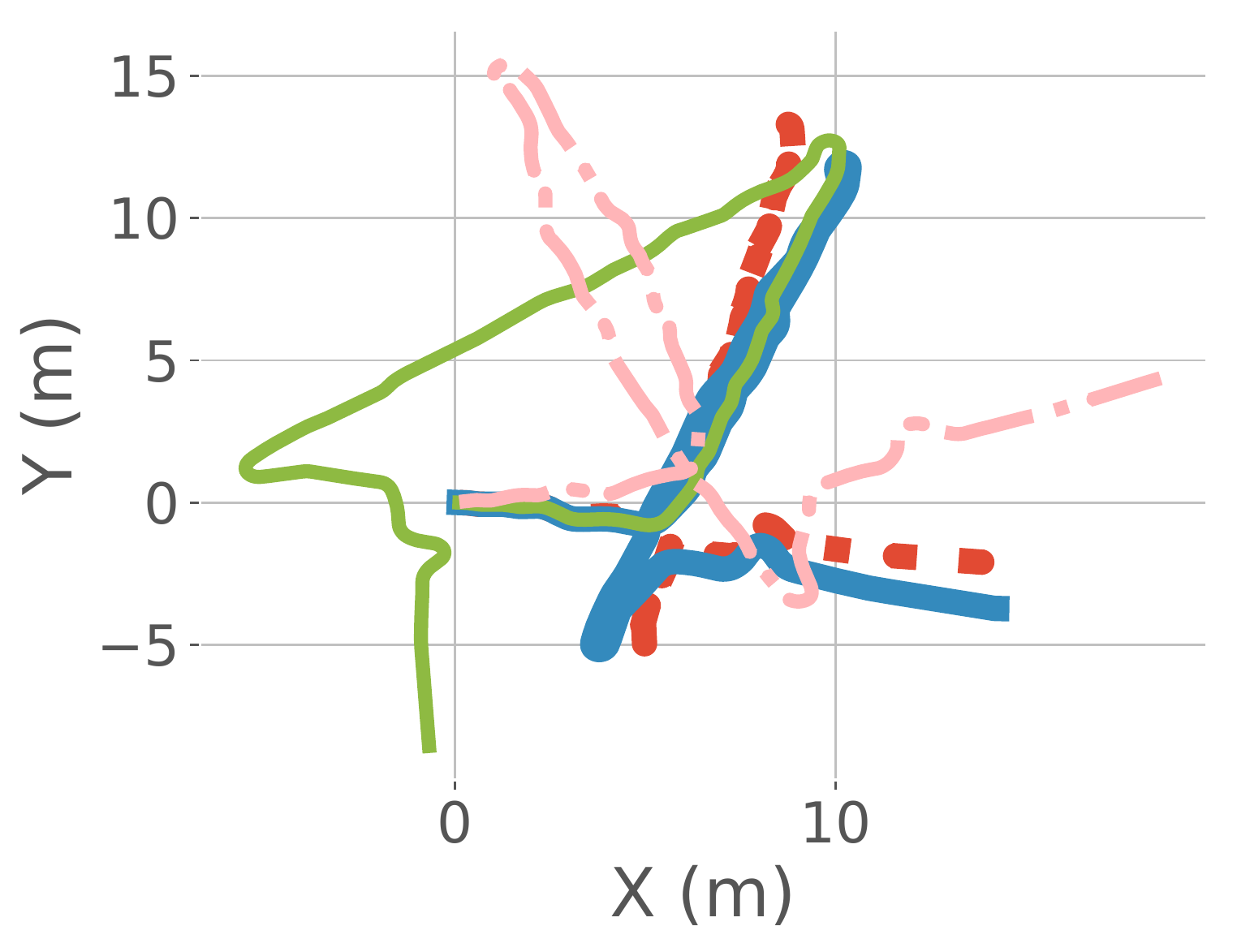} 
		\caption{Glass Corridor}
	\end{subfigure}%
	\begin{subfigure}[b]{0.24\textwidth}\centering
		\includegraphics[width=\columnwidth]{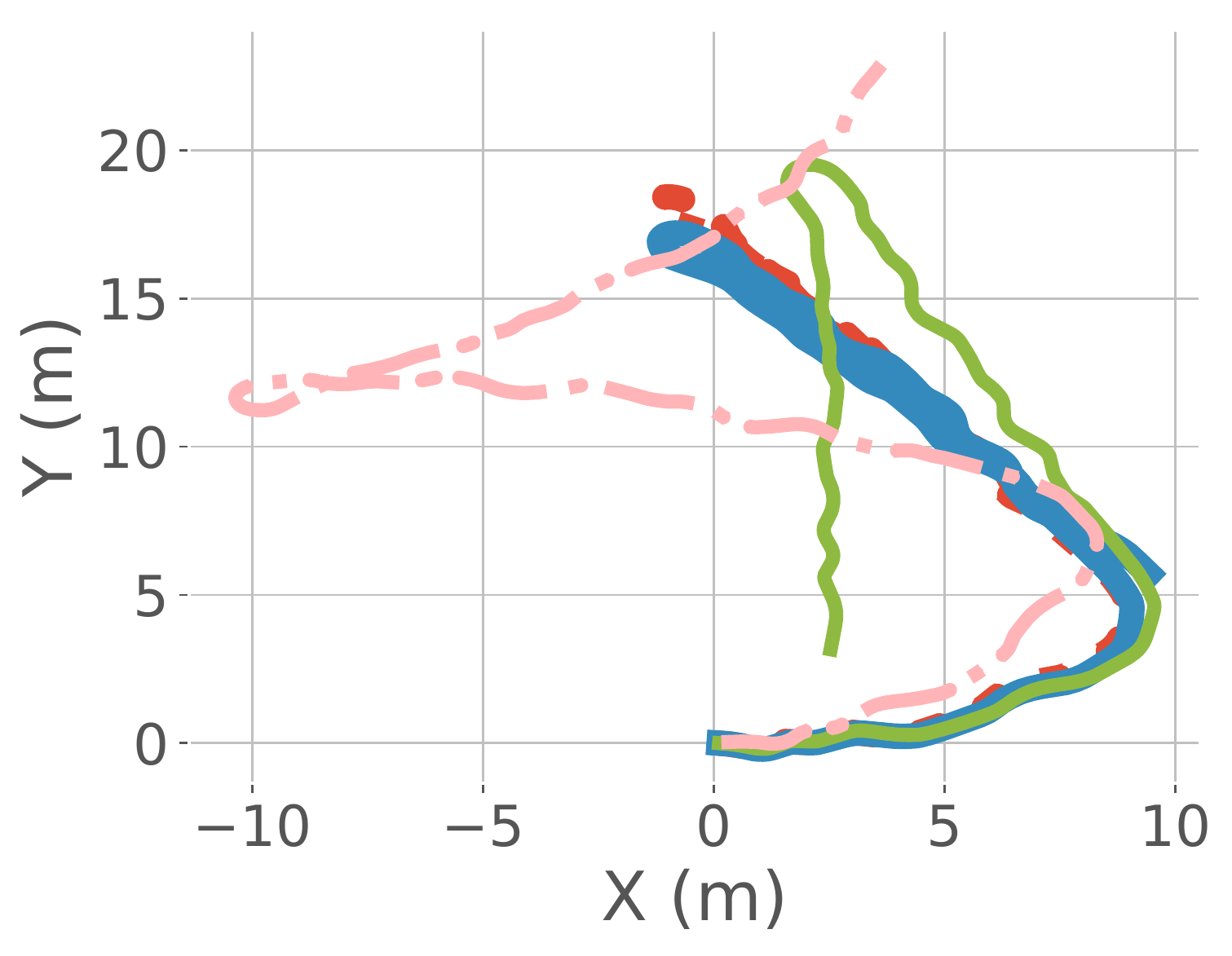} 
		\caption{Bumpy Floor}
	\end{subfigure}%
	\begin{subfigure}[b]{0.24\textwidth}\centering
		\includegraphics[width=\columnwidth]{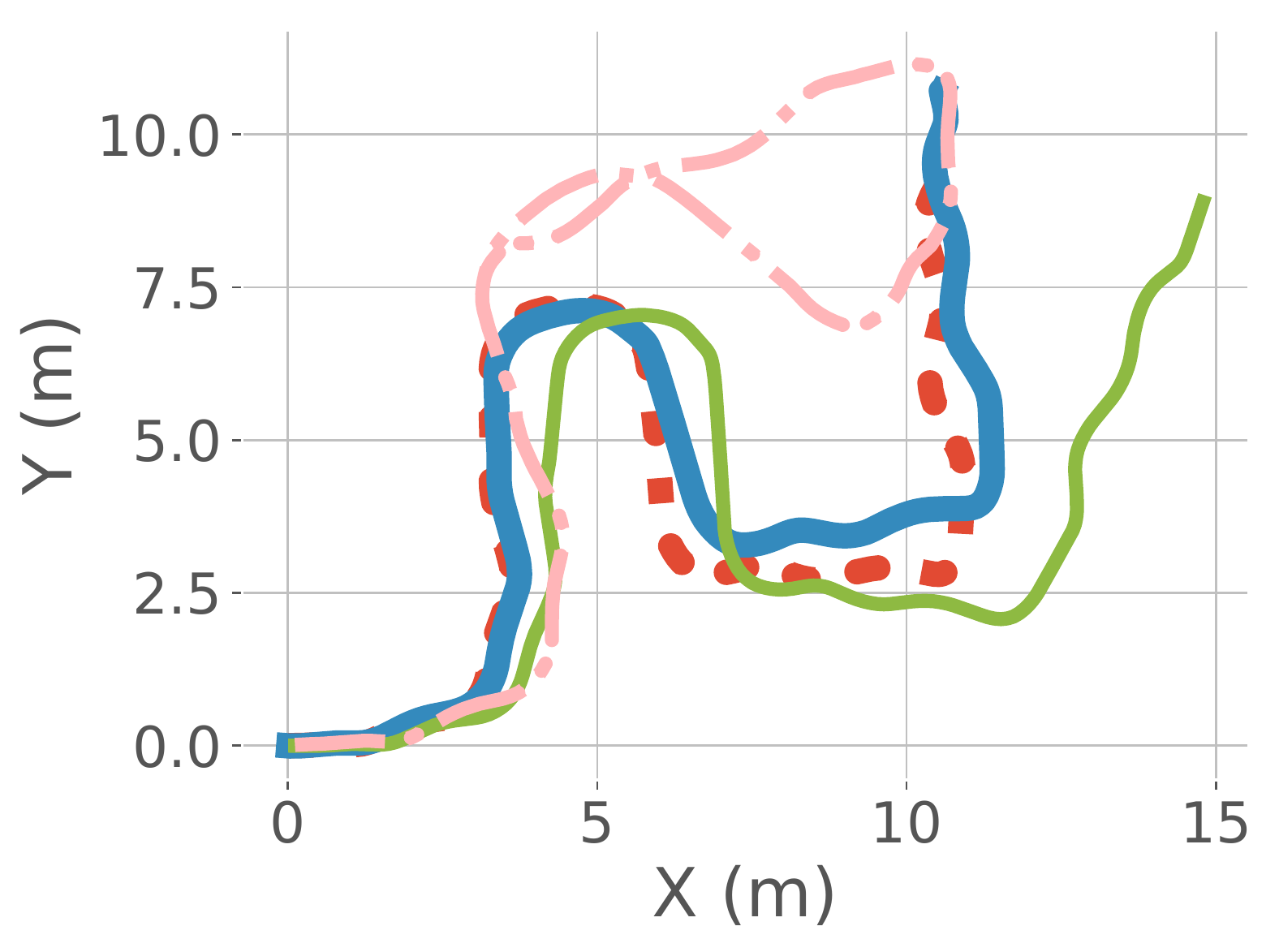} 
		\caption{Glare, Pathway}
	\end{subfigure}%
	\begin{subfigure}[b]{0.24\textwidth}\centering
		\includegraphics[width=\columnwidth]{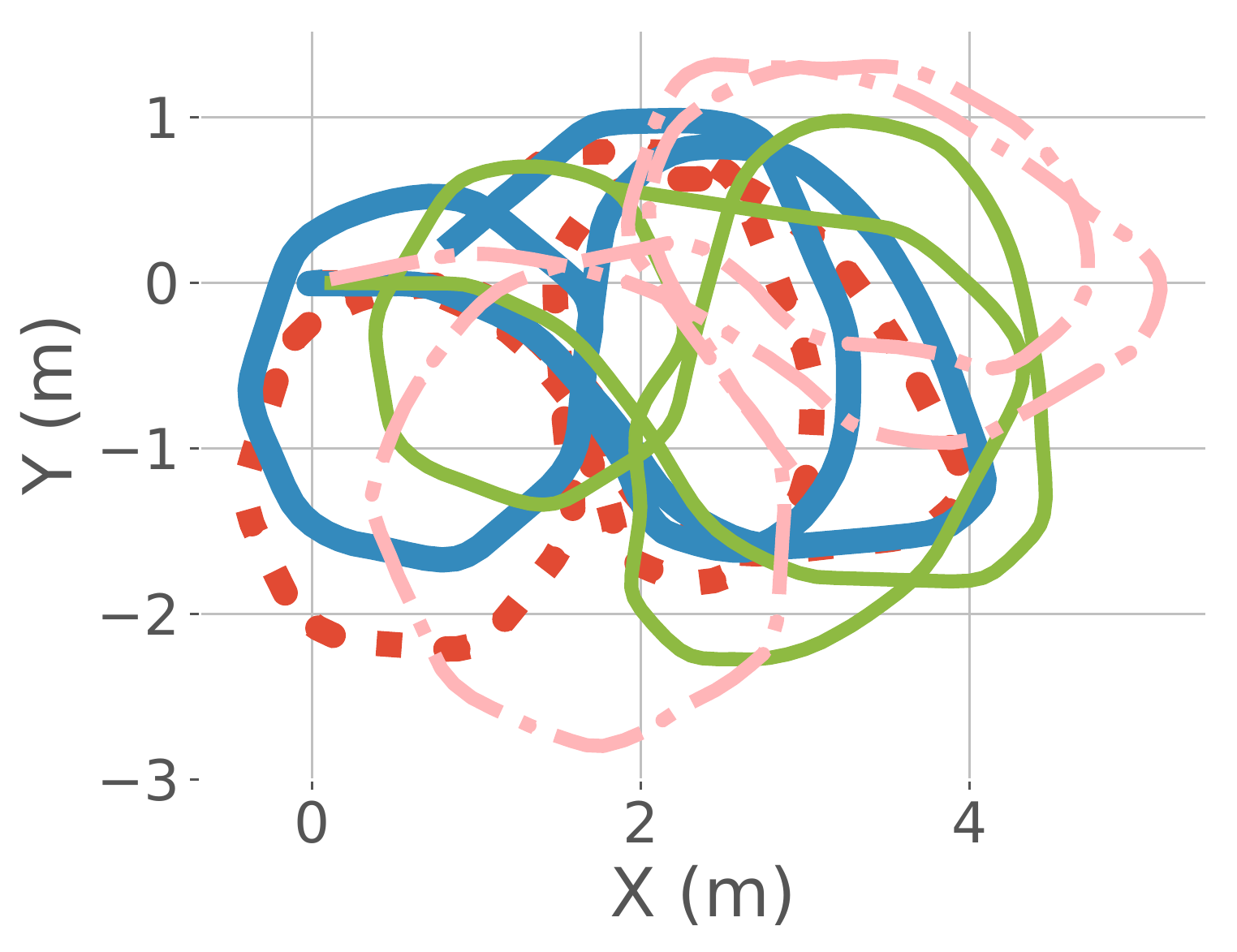} 
		\caption{Dark Room}
	\end{subfigure}%

	\begin{subfigure}[b]{0.24\textwidth}\centering
		\includegraphics[width=\columnwidth]{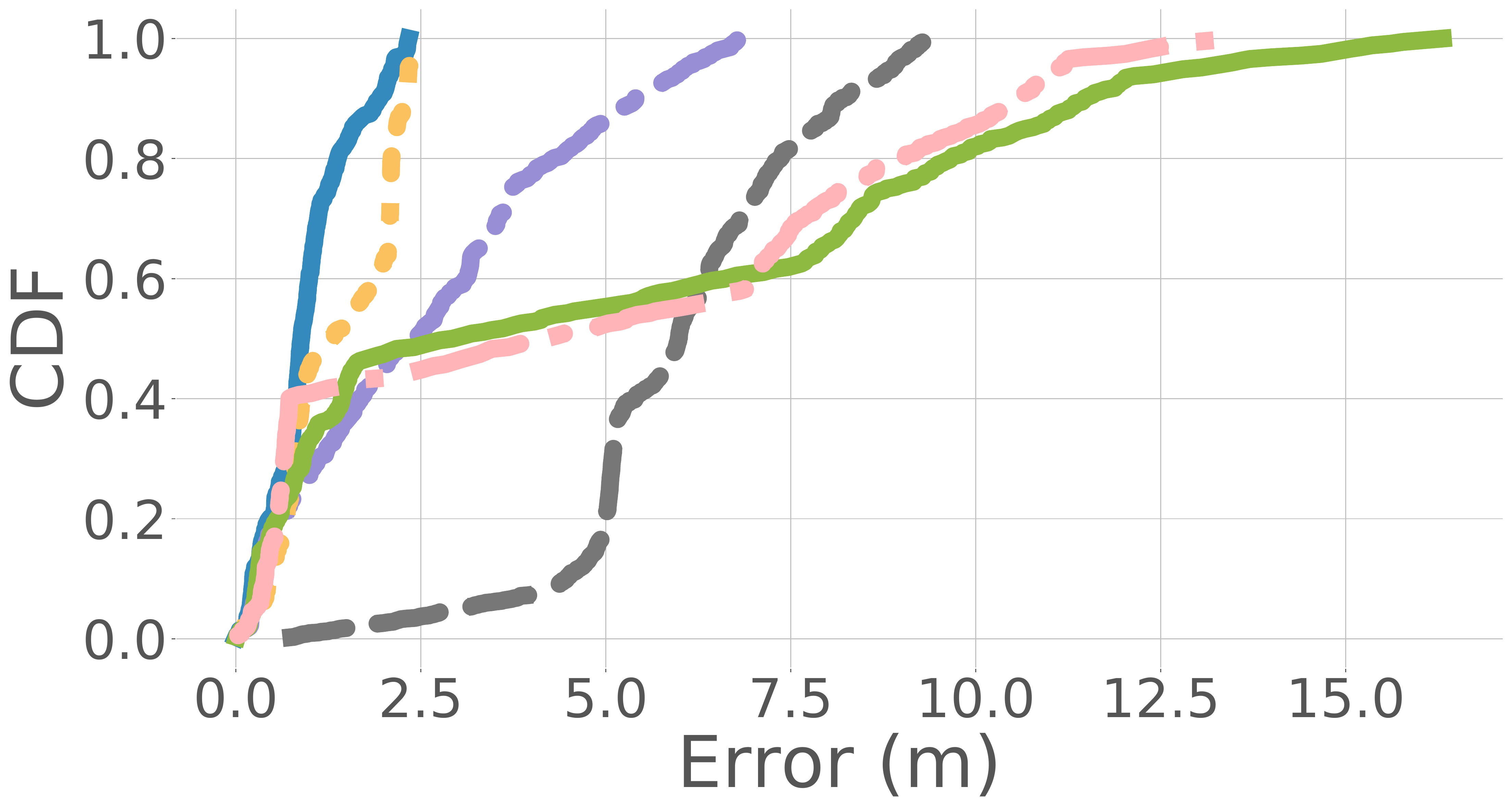} 
		\caption{Glass Corridor - $1.2\%$ ATE}
	\end{subfigure}%
	\begin{subfigure}[b]{0.24\textwidth}\centering
		\includegraphics[width=\columnwidth]{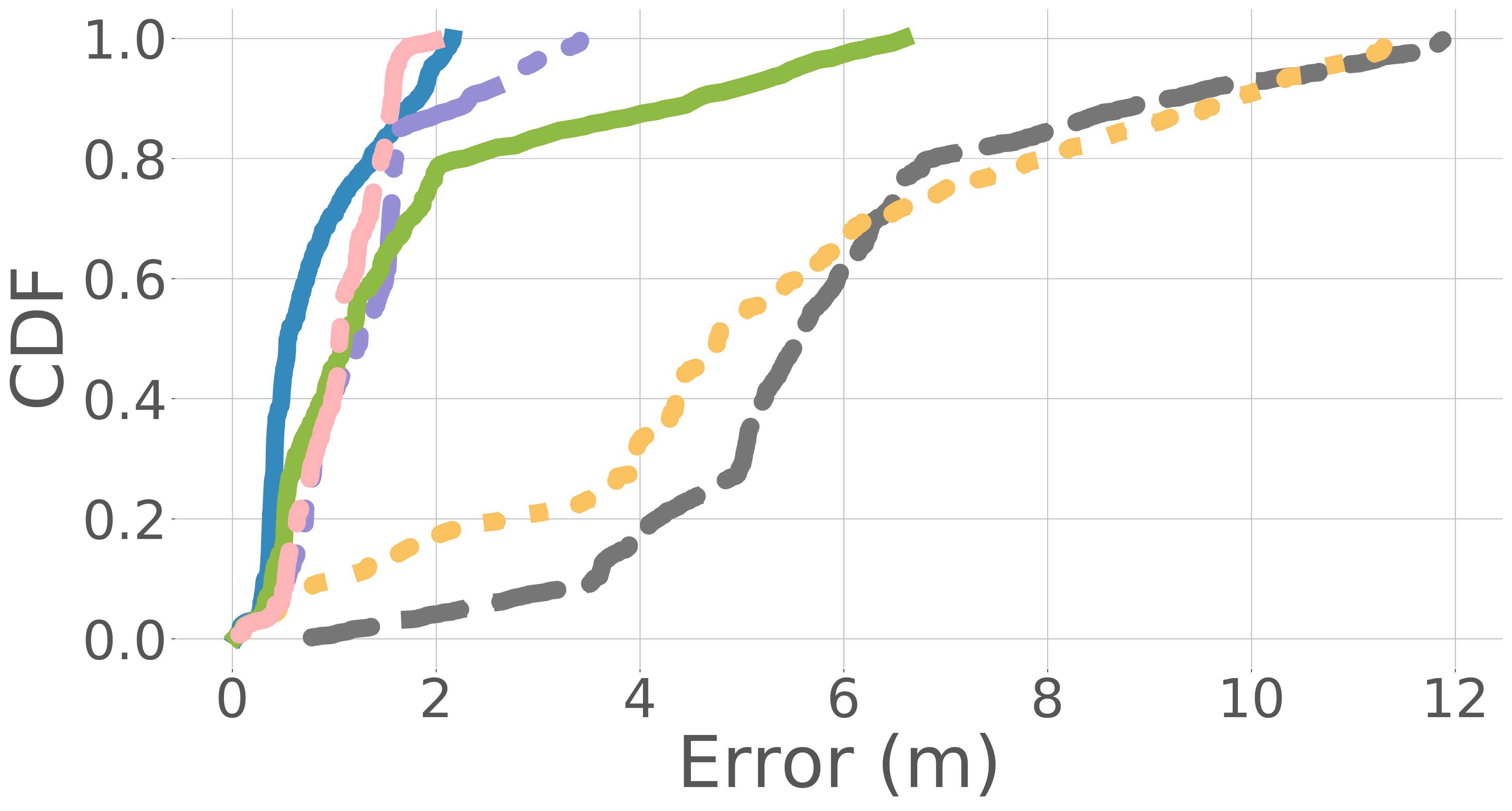} 
		\caption{Bumpy Floor - $1.3\%$ ATE}
	\end{subfigure}%
	\begin{subfigure}[b]{0.24\textwidth}\centering
		\includegraphics[width=\columnwidth]{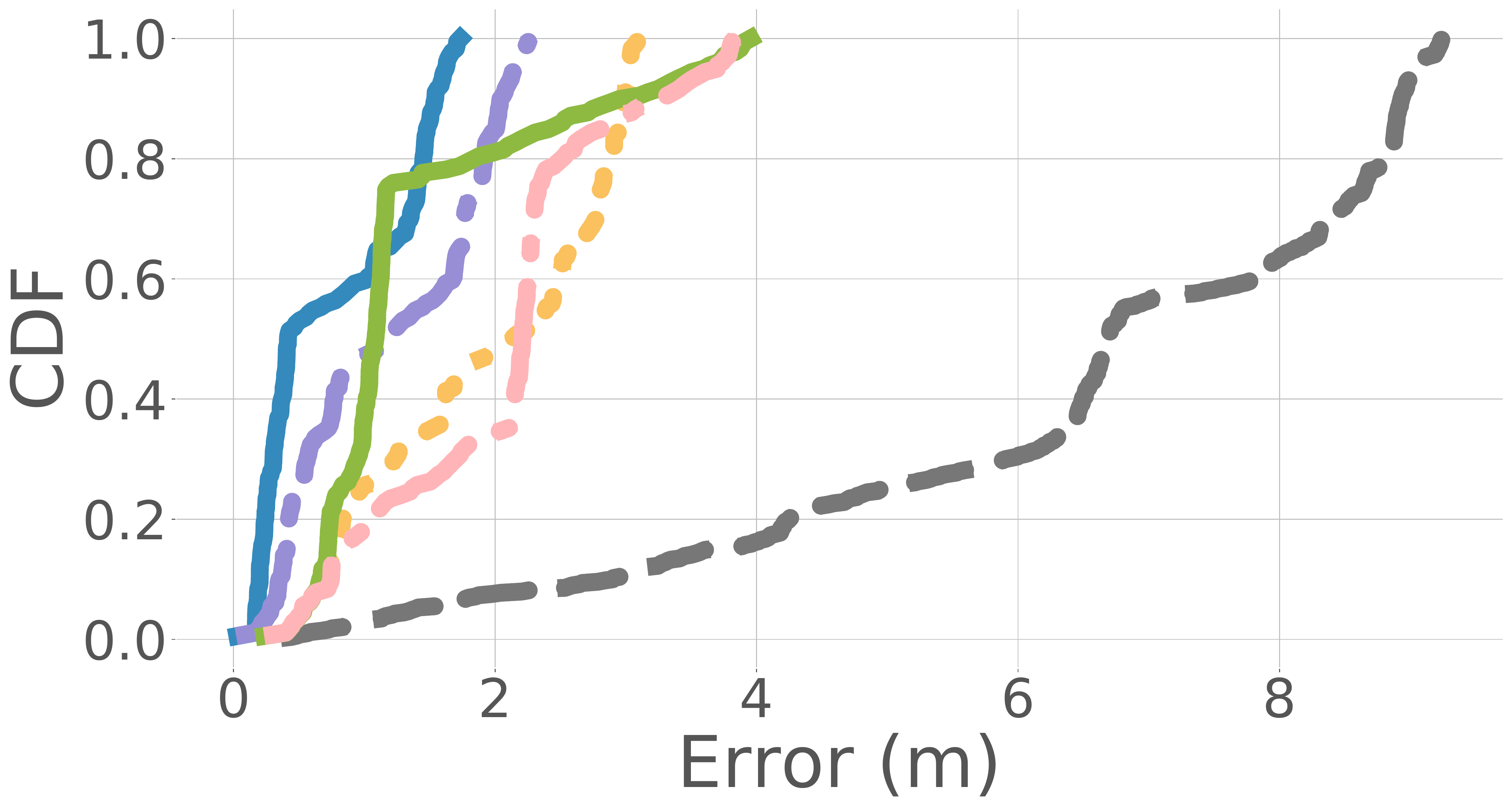} 
		\caption{Glare, Pathway - $1.8\%$ ATE}
	\end{subfigure}%
	\begin{subfigure}[b]{0.24\textwidth}\centering
		\includegraphics[width=\columnwidth]{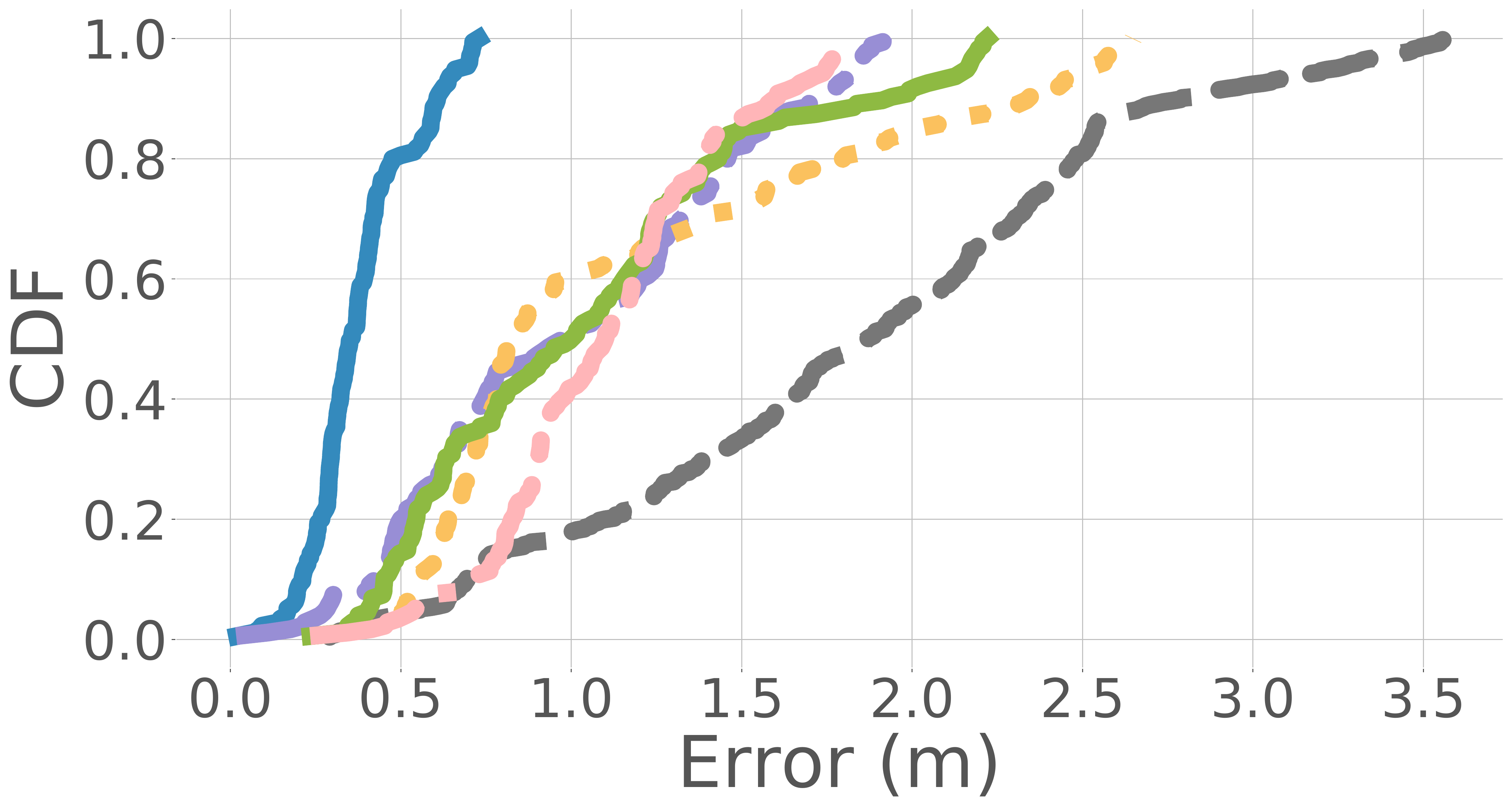} 
		\caption{Dark Room - $1.1\%$ ATE}
	\end{subfigure}%
\caption{Prediction comparison on four exemplar testing trajectories with the robot platforms. Top row: method-level comparison; Middle row: sensor-level comparison; Bottom row: cumulative distribution function (CDF) plots.}
\label{fig:robot_traj}
\end{figure*}

\subsection{Experimental Setting} % (fold)
\label{sub:experimental_setting}

% The robot platform is chosen as the main platform by which most experiments were carried out. Auxiliary evaluation with a handheld device was also conducted to provide more comprehensive results. 

\noindent \textbf{Metrics and Ground Truth}.
To evaluate the proposed model, we follow a widely used odometry benchmark \cite{handa2014benchmark} and adopt the Absolute Trajectory Error (\emph{ATE}) to quantify the tracking accuracy of an entire trajectory. The reason we use the entire trajectory is because the egomotion between consecutive frames is often too small to be numerically significant for the drifting effect. In contrast, ATE is considered a holistic metric that measures the composed long-term odometry performance. In particular, we report multiple statistics of ATE, including mean Root Mean Square Error (Mean), standard deviation (Std.) and max error (Max). 
To obtain the ground truth, we use a co-located lidar on the mobile platform and run laser-based SLAM (i.e., gmapping \cite{gmapping} in our case) to get its 6-DoF pose as our ground truth, following the practice of \cite{saputra2020deeptio}.

% To evaluate the proposed model, we adopt the mean square of Relative Pose Error (\emph{RPE}) and Absolute Trajectory Error (\emph{ATE}), as they are widely used for measuring the accuracy of odometry trajectories \cite{sturm2012benchmark,wang2017deepvo,chen2019selective}. RPE captures both $\mathbf{r}$ and $\mathbf{t}$ as discussed in \sect{\ref{sec:overview}. While RPE is an important metric to measure short-term pose (egomotion) accuracy, ATE is considered a more holistic and important metric than RPE as it measures the composed long-term odometry performance. To obtain the ground truth, we use a co-located lidar on the mobile platform and run laser-based SLAM (i.e., gmapping \cite{gmapping} in our case) to get its 6-DoF pose as our ground truth, following the practice of \cite{saputra2020deeptio}.

\noindent \textbf{Competing Approaches}. 
The baselines include both conventional and deep learning based methods. For the baseline combination of mmWave$+$IMU, we compare with the \emph{IMU-ICP} \cite{aghili2016robust}, which is a \textbf{traditional} fusion method that uses inertial sensors to bootstrap the ICP registration of the mmWave data via the adaptive Kalman filter. We also compare with other state-of-the-art multi-modal deep odometry with different sensors, including \emph{VINET} for RGB camera plus IMU and its variant \emph{DINET} for depth camera plus IMU. On top of these multi-modal baselines, we also compare with RANSAC-IMU \cite{civera20101} and TEASER \cite{yang2020teaser} in \sect{\ref{ssub:effectiveness_of_monet}} for examining standalone mmWave radar odometry i.e. without inertial aiding.

% As this is the first work using low-quality mmWave point clouds for egomotion estimation, we can only compare with 
% conventional methods that are designed for lidar point clouds. We therefore implement \emph{three} representative methods as our baselines including (1) ICP, (2) TEASER and (3) IMU-ICP.  

\noindent \textbf{Evaluation Protocol}.
Both datasets of mobile robot and handheld device are divided into training and test sets. 5-fold cross-validation is used to select the best model for testing. A key principle we follow in dividing them is to ensure that there is sufficient diversity between test and training sets to demonstrate generalization i.e. with distinct motion traces or tests on unseen environments. In this way, the model generalization ability can be fairly examined. Concretely, for the robot-platform evaluation, we train the model with $49$ training sequences and test the results on $6$ held-out sequences. We train another model with $27$ sequences for the handheld-device evaluation and test it on $3$ held-out sequences. The dataset is substantial in size: for the robot, the training set is 2878~m and the test set is 438~m. For the handheld case, the training set is 4380~m and the test set is 315~m.

\begin{table}[t]
\small
\caption{Overall results on the mobile robot. (Unit - Meter)}
\label{tab:robot_overall}
\centering
\begin{tabular}{|c|c|c|c|c|c|c|}
\hline
\multicolumn{2}{|c|}{\textbf{Sensors}} & \textbf{\begin{tabular}[c]{@{}c@{}}IMU\\ Only\end{tabular}} & \multicolumn{2}{c|}{\textbf{mmWave + IMU}} & \textbf{\begin{tabular}[c]{@{}c@{}}RGB\\ + IMU\end{tabular}} & \textbf{\begin{tabular}[c]{@{}c@{}}Depth\\ + IMU\end{tabular}} \\ \hline
\multicolumn{2}{|c|}{\textbf{Method}} & IONET & \begin{tabular}[c]{@{}c@{}}IMU-\\ ICP\end{tabular} & \sysname & VINET & DINET \\ \hline
\multirow{3}{*}{\textbf{\begin{tabular}[c]{@{}c@{}}3\\ D\end{tabular}}} & Mean & 2.644 & 5.054 & \textbf{0.814} & 1.955 & 2.255 \\ \cline{2-7} 
 & Std. & 1.843 & 1.742 & \textbf{0.444} & 1.479 & 1.526 \\ \cline{2-7} 
 & Max & 6.402 & 7.866 & \textbf{1.689} & 5.633 & 5.53 \\ \hline
\multirow{3}{*}{\textbf{\begin{tabular}[c]{@{}c@{}}2\\ D\end{tabular}}} & Mean & 2.45 & 5.008 & \textbf{0.764} & 1.936 & 2.239 \\ \cline{2-7} 
 & Std. & 1.748 & 1.745 & \textbf{0.438} & 1.478 & 1.525 \\ \cline{2-7} 
 & Max & 6.011 & 7.807 & \textbf{1.629} & 5.608 & 5.514 \\ \hline
\multicolumn{2}{|c|}{\textbf{Params (M)}} & 1.5 & - & 33.9 & 190.9 & 190.9 \\ \hline
\end{tabular}
\end{table}

\subsection{Mobile Robot Performance} % (fold)
\label{sub:mobile_robot_performance}

We start by evaluating the odometry performance on mobile robot as our primary platform. 

\subsubsection{Overall Performance} % (fold)
\label{ssub:overall_performance}

Tab.~\ref{tab:robot_overall} summarizes the overall performance in 2D and 3D space with fusion of two sensor modalities. One of the sensors is fixed as an IMU due its pervasiveness on modern mobile platforms. We compare against RGB+IMU and Depth+IMU, using benchmark deep learning approaches. For comparison, the odometry performance using only an IMU alone is also shown. 

As can be seen, \sysname achieves the lowest error amongst all methods, yielding a 3D ATE of $0.814$m, approximately a $1.3\%$ error in drift when taking the trajectory distances into account. This substantially outperforms the IMU-ICP methods by almost an order of magnitude, which shows the worst performance, even worse than IMU alone. This is due to its inability to deal with the low-quality mmWave point clouds. 
\sysname also largely outperforms the other dual-sensor systems like RGB+IMU and depth+IMU. As discussed in the \sect{\ref{sec:implementation}}, our datatset contains challenging illumination conditions that occur in the real world (e.g., glare, dimness, darkness and glass windows) that  impede their performance. Conversely, as mmWave is insensitive to ambient illumination, it is able to provide more accurate odometry estimation. \sysname's ATE is further reduced to $0.764$m in the 2D plane, which allows for sub-metre indoor tracking. 

\begin{table}[t]
\small
\caption{Extending \sysname to triple-sensor fusion. Sensor Key: \emph{M} - mmWave, \emph{I} - inertial, \emph{V} - RGB, \emph{D} - Depth.}
\label{tab:robot_3sensors}
\centering
\begin{tabular}{|c|c|c|c|c|c|c|}
\hline
\multirow{2}{*}{\textbf{Method}} & \multicolumn{3}{c|}{\textbf{3D}} & \multicolumn{3}{c|}{\textbf{2D}} \\ \cline{2-7} 
 & \textbf{Mean} & \textbf{Std.} & \textbf{Max} & \textbf{Mean} & \textbf{Std.} & \textbf{Max} \\ \hline
\begin{tabular}[c]{@{}c@{}}M + I + V\\ (w.o. Att.)\end{tabular} & 0.862 & 0.469 & 1.871 & 0.838 & 0.458 & 1.838 \\ \hline
\begin{tabular}[c]{@{}c@{}}M + I + D\\ (w.o. Att.)\end{tabular} & 1.194 & 0.791 & 2.791 & 1.008 & 0.746 & 2.563 \\ \hline
M + I + V & 0.702 & 0.399 & 1.608 & 0.673 & 0.381 & 1.552 \\ \hline
M + I + D & 0.769 & 0.498 & 1.821 & 0.761 & 0.496 & 1.813 \\ \hline
\end{tabular}
\end{table}

\subsubsection{Extending to Triple Sensor Egomotion} % (fold)
\label{ssub:extending_to_triple_sensor_odometry}

\sysname can also be extended to triple-sensor egomotion systems. On the basis of \sect{\ref{ssub:overall_performance}}, we use mmWave and IMU as the two baseline sensors and choose either the RGB camera or depth camera as the third,  as shown in Tab.~\ref{tab:robot_3sensors}. When comparing to Tab.~\ref{tab:robot_overall}, an interesting tradeoff between system complexity and performance can be noticed. Although extending \sysname to triple-sensor egomotion improves the accuracy, the delta is not substantial. Compared to the two-sensor version, only $\sim 13\%$ and $\sim 6\%$ ATE reduction are observed after incorporating an extra RGB and depth camera respectively. This, however, is at the cost of hardware/sensor overhead and computational latency. We thus suggest that end users carefully consider this tradeoff before extending to three sensors.

%however the 3D ATE gap can be as large as $30\%$ when the proposed attention module is not used. This demonstrates that the proposed deep fusion network can effectively select and reweigh sensor inputs according to their contextual utility.
%Similar to the analysis results of two-sensor odometry, is attributed to less complementary behaviors activated for adaptive modeling.
% More concretely, an effective network should automatically tone down the contribution of RGB cameras if adverse lighting is encountered and put more attention on other complementary sensors. This is however beyond the reach of a direct fusion (i.e., DeepMIVO) but requires our attention module. Fig.~\ref{} \todo{MAYBE add a figure? What do you think, Andrew?}.

\subsubsection{Effectiveness of mmWave Subnet} % (fold)
\label{ssub:effectiveness_of_monet}

We next examine the effectiveness of the mmWave subnet alone, i.e. without the use of the IMU or attention, as discussed in Sec.~\ref{sub:mmwave_odometry_sub_network}. We compare the performance of standalone subnet with two \textbf{conventional} point registration methods: (1) \emph{ICP} and (2) \emph{TEASER}. ICP \cite{civera20101} is the widely adopted point registration method to estimate egomotion. We follow the conventional practice that uses RANSAC algorithm \cite{kim2013image} for outlier rejection. TEASER \cite{yang2020teaser} is the latest state-of-the-art method using a sophisticated optimization to achieve robust point registration against strong outliers. Due to the very sparse and noisy mmWave point clouds, these two baselines unfortunately cannot succeed for most testing sequences (refer to \sect{\ref{sub:chaotic_point_correspondences}}). We therefore consider only $3$ out of our $7$ testing sequences on which the baselines can give reasonable odometry estimation. As reference, we also consider inertial only odometry (IONET).

\begin{figure}[!t]
	\centering
	\begin{subfigure}[b]{0.235\textwidth}\centering
		\includegraphics[width=\columnwidth]{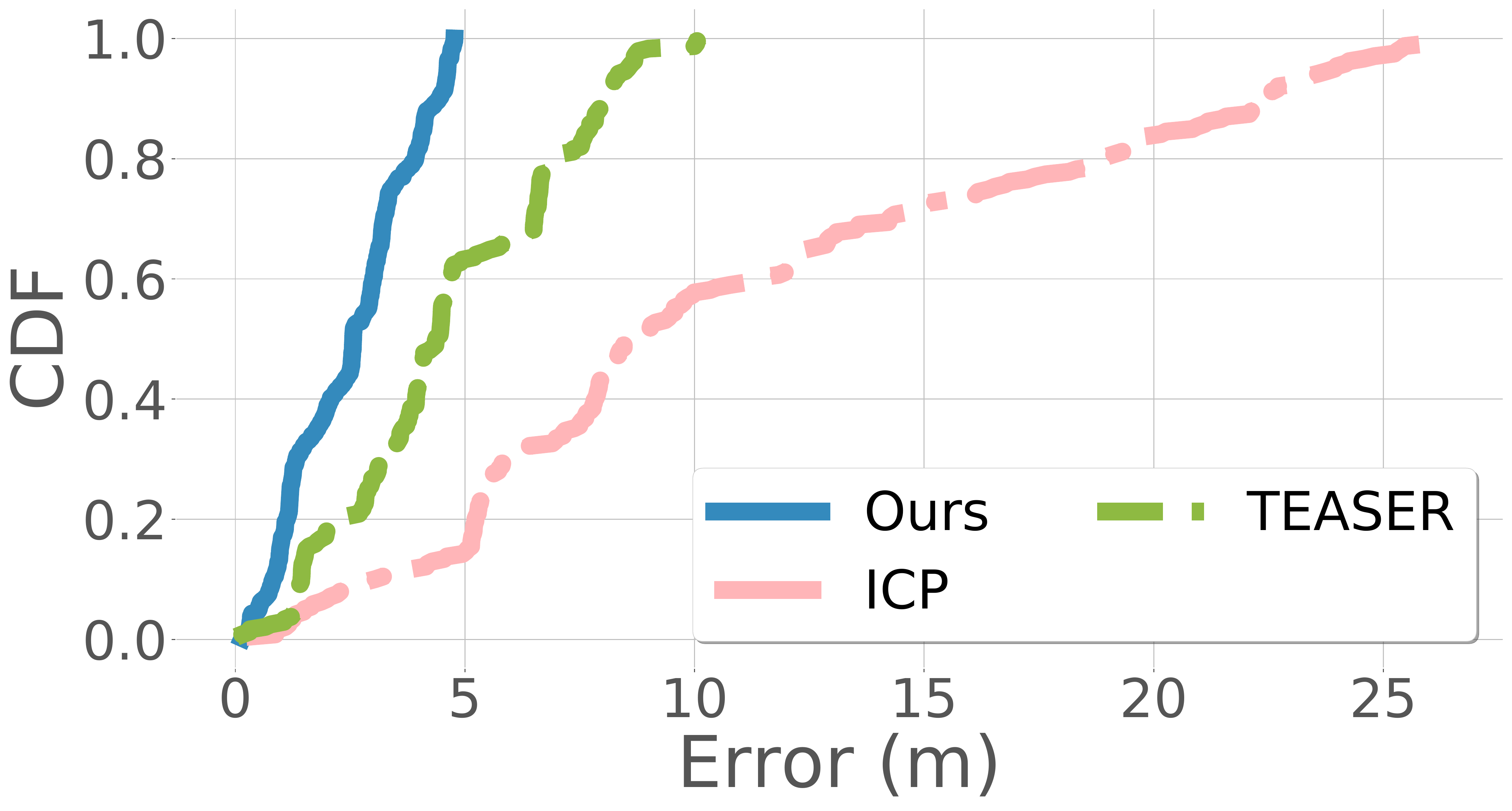} 
		\caption{Corridor CDF}
	\end{subfigure}
	\hfill
	\begin{subfigure}[b]{0.235\textwidth}\centering
		\includegraphics[width=\columnwidth]{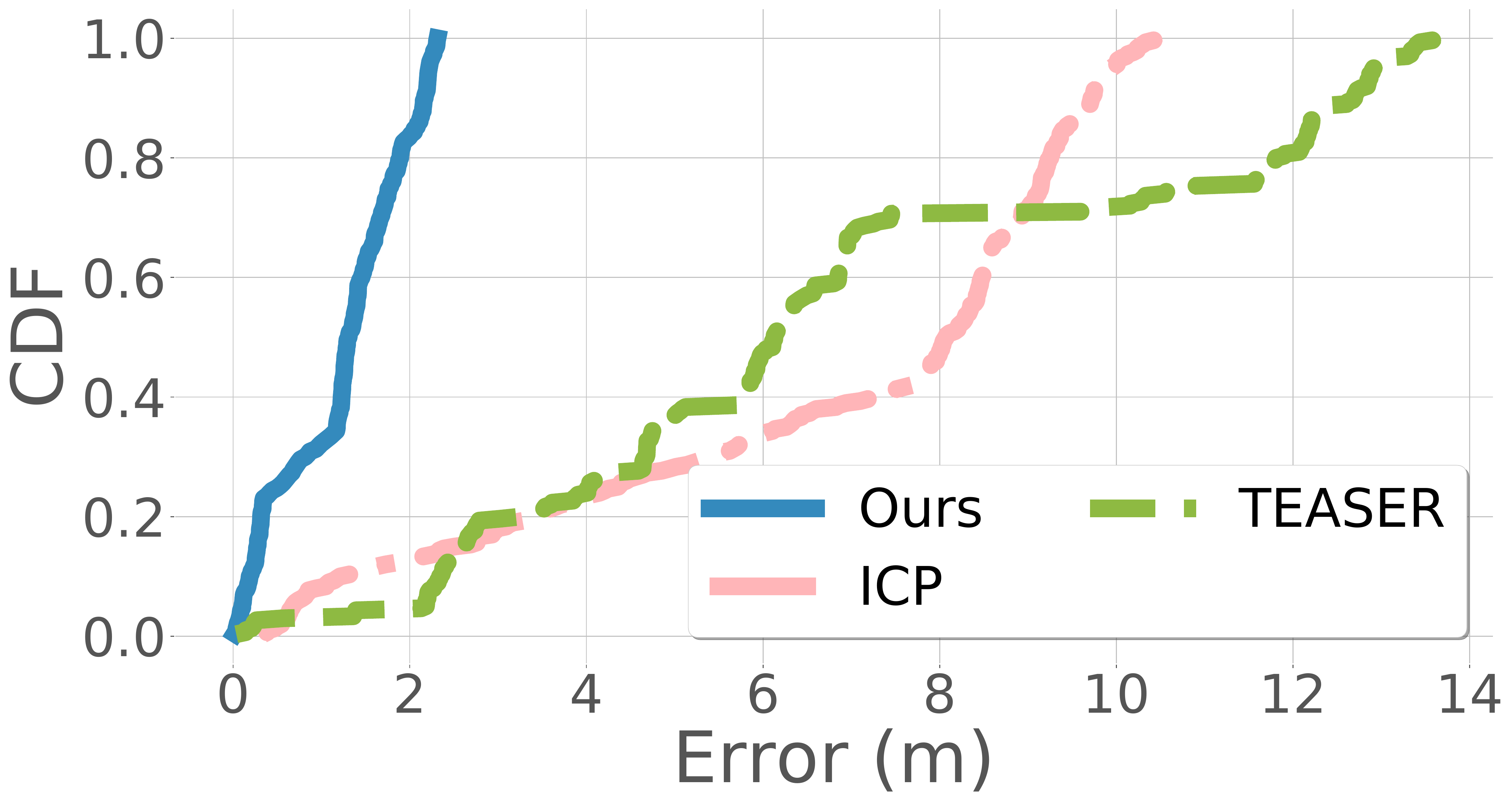} 
		\caption{Atrium CDF}
	\end{subfigure}%
\caption{CDF of two test sequences with mmWave Subnet.}
\label{fig:modom_cdf}
\end{figure}

\noindent \textbf{Results}.
As can be seen in Fig.~\ref{fig:modom_cdf}, the mmWave subnet surpasses both baselines by at least $3$-fold on all testing sequences. In particular, we found that the average 2D and 3D ATE are only $2.47$m and $2.50$m. In contrast, ICP and TEASER struggle in providing reliable egomotion estimation due to the sparse and noisy point clouds.

\begin{table}[!t]
\small
\caption{Investigation into different attention strategies}
\label{tab:robot_attention}
\centering
\begin{tabular}{|c|c|c|c|c|c|c|}
\hline
\multicolumn{2}{|c|}{\textbf{Method}} & \begin{tabular}[c]{@{}c@{}}No\\ Att.\end{tabular} & \begin{tabular}[c]{@{}c@{}}Single-Stage\\  Att. \cite{saputra2020deeptio,chen2019selective}\end{tabular} & \begin{tabular}[c]{@{}c@{}}w.o. \\ Cross Att.\end{tabular} & \begin{tabular}[c]{@{}c@{}}w.o. \\ Self Att.\end{tabular} & \sysname \\ \hline
\multirow{3}{*}{\textbf{\begin{tabular}[c]{@{}c@{}}3\\ D\end{tabular}}} & \textbf{Mean} & 1.373 & 1.494 & 0.923 & 0.949 & \textbf{0.814} \\ \cline{2-7} 
 & \textbf{Std.} & 0.784 & 0.838 & 0.437 & 0.532 & \textbf{0.444} \\ \cline{2-7} 
 & \textbf{Max} & 2.981 & 3.122 & 1.893 & 1.971 & \textbf{1.689} \\ \hline
\multirow{3}{*}{\textbf{\begin{tabular}[c]{@{}c@{}}2\\ D\end{tabular}}} & \textbf{Mean} & 1.369 & 1.441 & 0.91 & 0.935 & \textbf{0.764} \\ \cline{2-7} 
 & \textbf{Std.} & 0.781 & 0.793 & \textbf{0.437} & 0.528 & 0.438 \\ \cline{2-7} 
 & \textbf{Max} & 2.976 & 3.027 & 1.883 & 1.958 & \textbf{1.629} \\ \hline
\multicolumn{2}{|c|}{\textbf{Params (M)}} & 31.6 & 42.7 & 32.3 & 33.3 & 33.9 \\ \hline
\end{tabular}
% \vspace{-0.4cm}
\end{table}

\subsubsection{Potential impact of multiple mmWave radars} % (fold)
\label{ssub:potential_impact_of_multiple_mmwave_radars}

\begin{figure}[!t]
		\includegraphics[width=0.9\columnwidth]{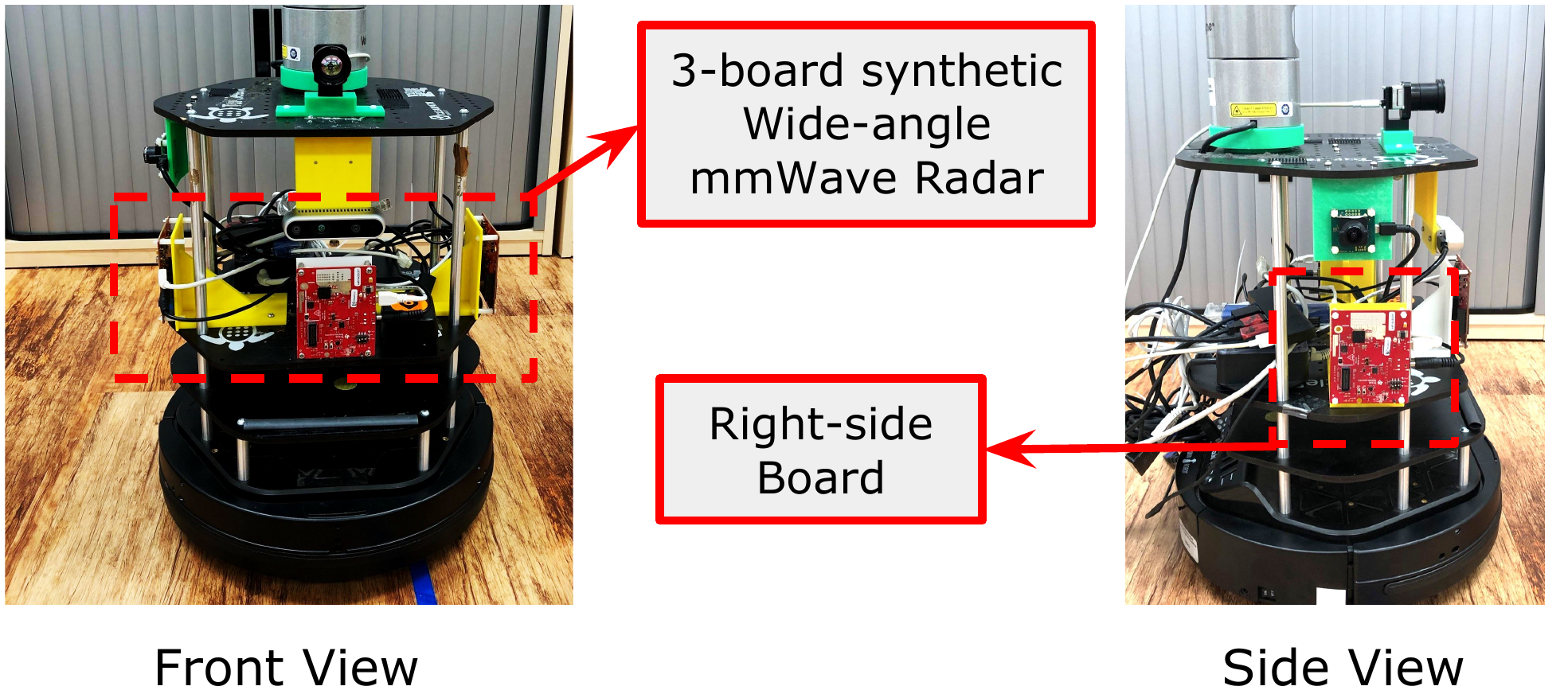} 
		\caption{The wide-angle mmWave radar synthesized by 3 co-located radars.}
\label{fig:wide_angle_robot}
\end{figure}

As an emerging technology, the Field of View (FoV) of current single-chip mmWave radars is still limited. However, given the promise of such technology and the current practice of installing multiple mmWave radars on an automobile \cite{radar_mutiple}, it is worth studying the potential impact of multiple co-located radars that collectively form a wide-angle sensing view. To fully understanding this impact, we evaluate the \emph{mmWave-only} version of \sysname on $5$ out of the $7$ test sequences, in which there were $3$ radars co-located on the robot (front, left and right sides as shown in Fig.~\ref{fig:wide_angle_robot}). This collocation changes the horizontal FoV from $120$ deg to $240$ deg. \revise{To align the radars, we determine the transformation to the center of our mobile robot, and then stitch the three transformed point clouds into a single, merged cloud.}
For baselines, we compare its egomotion estimation against other sensor modalities: RGB, depth and inertial.

\begin{figure}[!t]
	\centering
	\includegraphics[width=0.9\columnwidth]{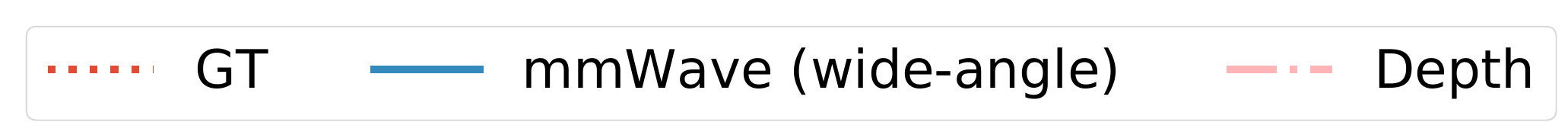}

	\begin{subfigure}[b]{0.24\textwidth}\centering
		\includegraphics[width=\columnwidth]{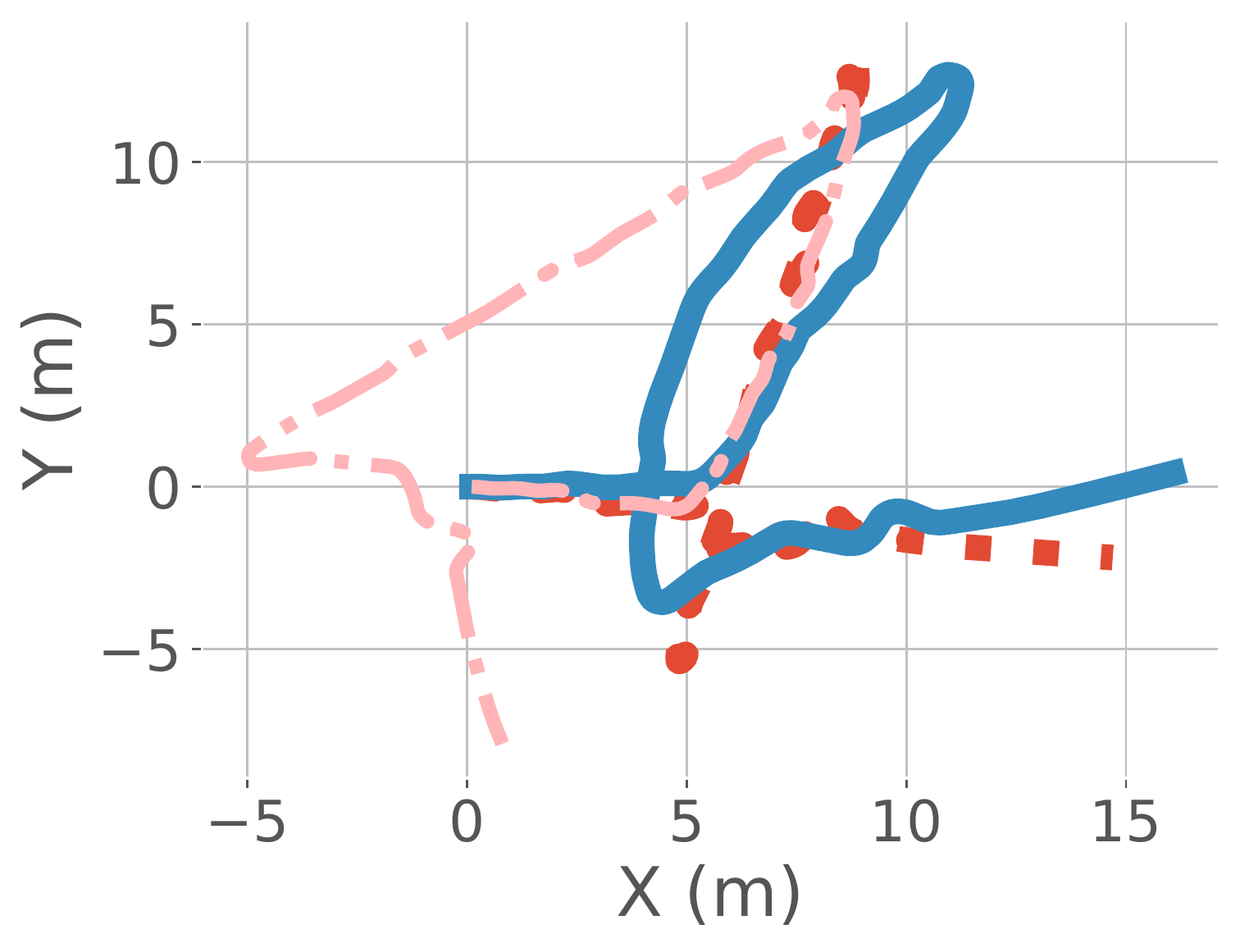} 
		\caption{Glass Corridor}
	\end{subfigure}%
	\begin{subfigure}[b]{0.24\textwidth}\centering
		\includegraphics[width=\columnwidth]{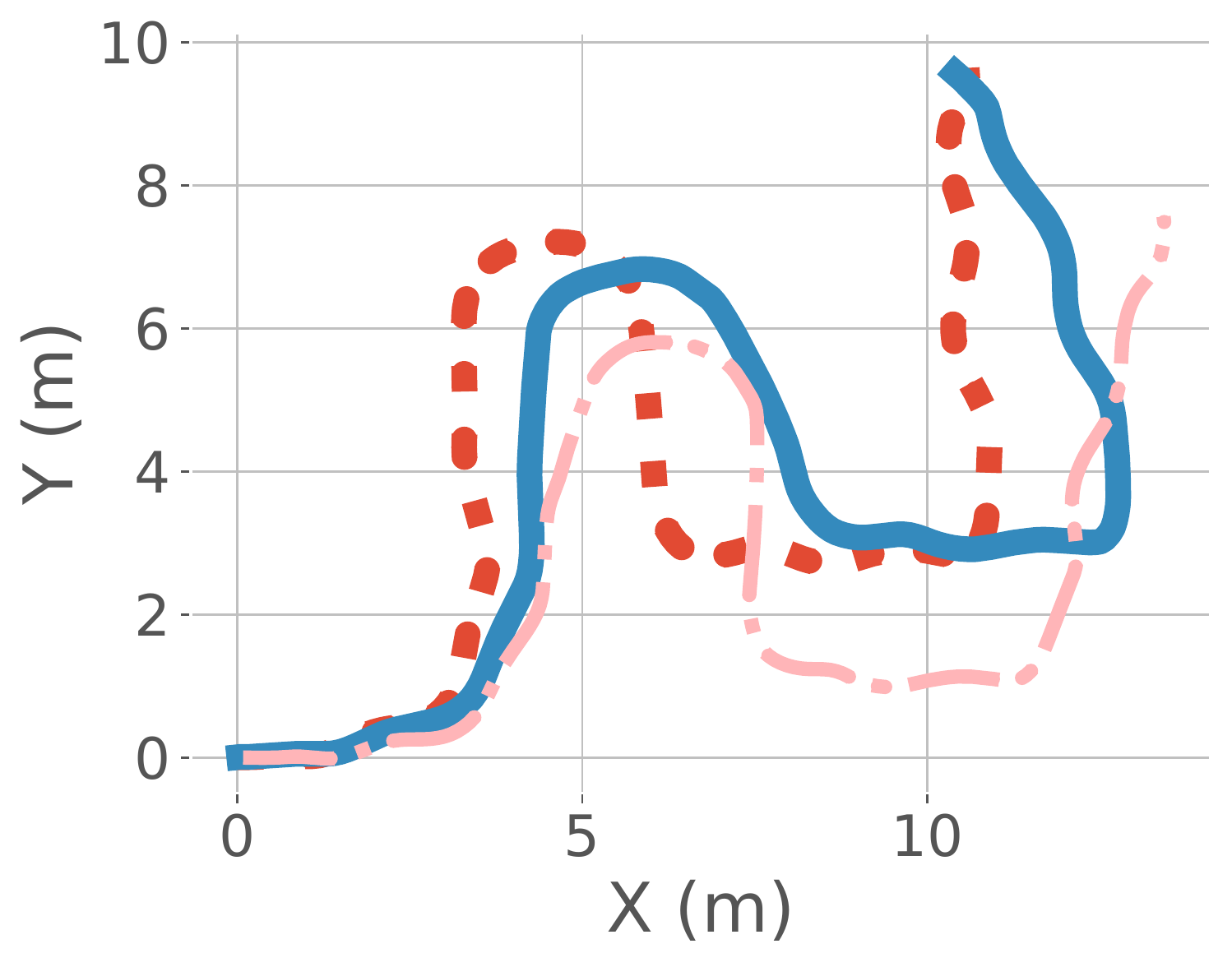} 
		\caption{Pathway}
	\end{subfigure}%
	
\caption{Two example testing sequences with wide-angle mmWave odometry alone. For clarity, we only show the trajectory for depth camera as the most competitive alternative in Tab.~\ref{tab:wide_angle_radar}.}
\label{fig:single_sensor_traj}
\end{figure}

\noindent \textbf{Results}.
As can be seen in Tab.~\ref{tab:wide_angle_radar}, the `wide-angle' view created by co-located radars significantly reduces the ATE of mmWave-only odometry over the single-radar case (i.e., normal view). Meanwhile, wide-angle mmWave-only odometry outperforms all other sensor modalities on every dimension of the ATE metric. Compared to the best baseline (i.e., Depth camera only), wide-angle \sysname reduces mean ATE by $35\%$. Fig.~\ref{fig:single_sensor_traj} shows two exemplar trajectory comparisons between wide-angle mmWave and depth camera.
\revise{This experiment demonstrates that a wide-angle view (e.g. through increased number of MIMO antenna elements) will lead to improved performance.} Another takeaway from Tab.~\ref{tab:wide_angle_radar} is that all single-modal odometry systems have their limitations in egomotion estimation, reinforcing the need for a multi-modal solution.

\begin{table}[!t]
\small
\caption{Performance of a `wide-angle' mmWave radar system.}
\label{tab:wide_angle_radar}
\centering
\begin{tabular}{|c|c|c|c|c|c|c|}
\hline
\multicolumn{2}{|c|}{\multirow{2}{*}{\textbf{Sensors}}} & \multicolumn{2}{c|}{\textbf{mmWave}} & \multirow{2}{*}{\textbf{Depth}} & \multirow{2}{*}{\textbf{IMU}} & \multirow{2}{*}{\textbf{RGB}} \\ \cline{3-4}
\multicolumn{2}{|c|}{} & Narrow & Wide Angle &  &  &  \\ \hline
\multirow{3}{*}{\textbf{\begin{tabular}[c]{@{}c@{}}3\\ D\end{tabular}}} & Mean & 2.611 & \textbf{1.239} & 1.882 & 2.285 & 1.983 \\ \cline{2-7} 
 & Std. & 1.396 & \textbf{0.783} & 1.220 & 1.134 & 1.312 \\ \cline{2-7} 
 & Max & 4.641 & \textbf{2.990} & 4.580 & 4.390 & 4.673 \\ \hline
\multirow{3}{*}{\textbf{\begin{tabular}[c]{@{}c@{}}2\\ D\end{tabular}}} & Mean & 2.507 & \textbf{1.208} & 1.879 & 2.275 & 1.973 \\ \cline{2-7} 
 & Std. & 1.218 & \textbf{0.792} & 1.220 & 1.137 & 1.309 \\ \cline{2-7} 
 & Max & 4.210 & \textbf{2.979} & 4.578 & 4.384 & 4.658 \\ \hline
\end{tabular}
\end{table}

\subsubsection{Effectiveness of Mixed Attention} % (fold)
\label{ssub:impact_of_attention_strategies}

Lastly we quantitatively validate our mixed attention introduced in \sect{\ref{sub:mixed_attention_for_fusion}} for multi-modal sensor fusion. Specifically, we investigate different fusion strategies in \sysname, including (1) No attention, (2) single-stage attention (see \sect{\ref{sub:discussion_of_mixed_attention}}), (3) \sysname without self-attention (4) \sysname without cross-attention.
Tab.~\ref{tab:robot_attention} suggests that our mixed attention strategy brings $\sim 40\%$ performance gain compared to \sysname without attention. Moreover, it consistently outperforms other attention strategies in both 2D and 3D planes,
providing more than a $50\%$ accuracy increase over single-stage attention. This is consistent with the analysis in \sect{\ref{sub:discussion_of_mixed_attention}}. Moreover, the single-stage attention model has more parameters than \sysname due to its `fat' mask generation on the concatenated vector.
On the other hand, it can be noticed from Tab.~\ref{tab:robot_attention} that every stage of attention contributes; removing either one causes sub-optimal performance. Interestingly, the ATE performances of self-attention only and cross-attention only are very close, suggesting their equal importance.

\subsection{Handheld Device Performance} % (fold)
\label{sub:handheld_device_performance}

Ground robots move in a horizontal plane and therefore are relatively constrained for egomotion estimation. We therefore investigate the performance of \sysname in a less constrained case i.e. carried in the hand.

\subsubsection{Overall Performance} % (fold)
\label{ssub:overall_performance_hand}

Tab.~\ref{tab:hand_overall} summarizes overall performance. As can be seen, \sysname clearly surpasses the other methods in both 3D and 2D planes. It yields an average 3D ATE of $1.895$m, equivalent to an $1.8\%$ trajectory drift. In the 2D space, its error is further reduced to $1.252$m.

% Even without using the mixed attention strategy, the combination of mmWave+IMU is still superior. 

\begin{figure}[!t]
	\centering
	\includegraphics[width=\columnwidth]{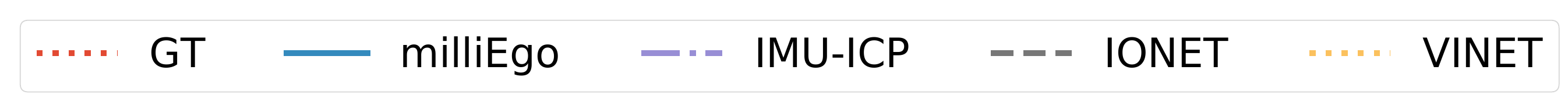}

	\begin{subfigure}[b]{0.24\textwidth}\centering
		\includegraphics[width=\columnwidth]{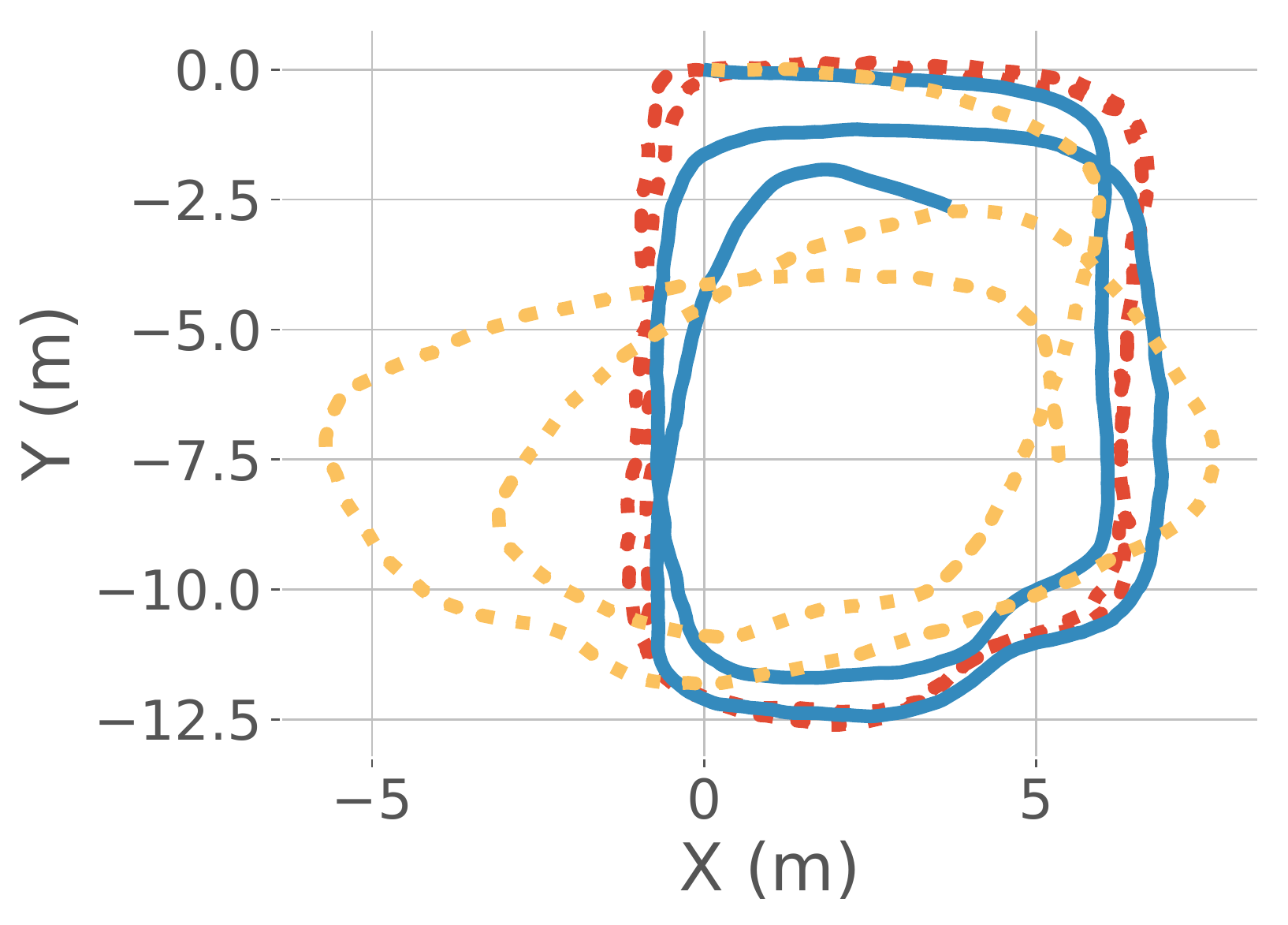} 
		\caption{Corridor Trajectory}
	\end{subfigure}%
	\begin{subfigure}[b]{0.24\textwidth}\centering
		\includegraphics[width=\columnwidth]{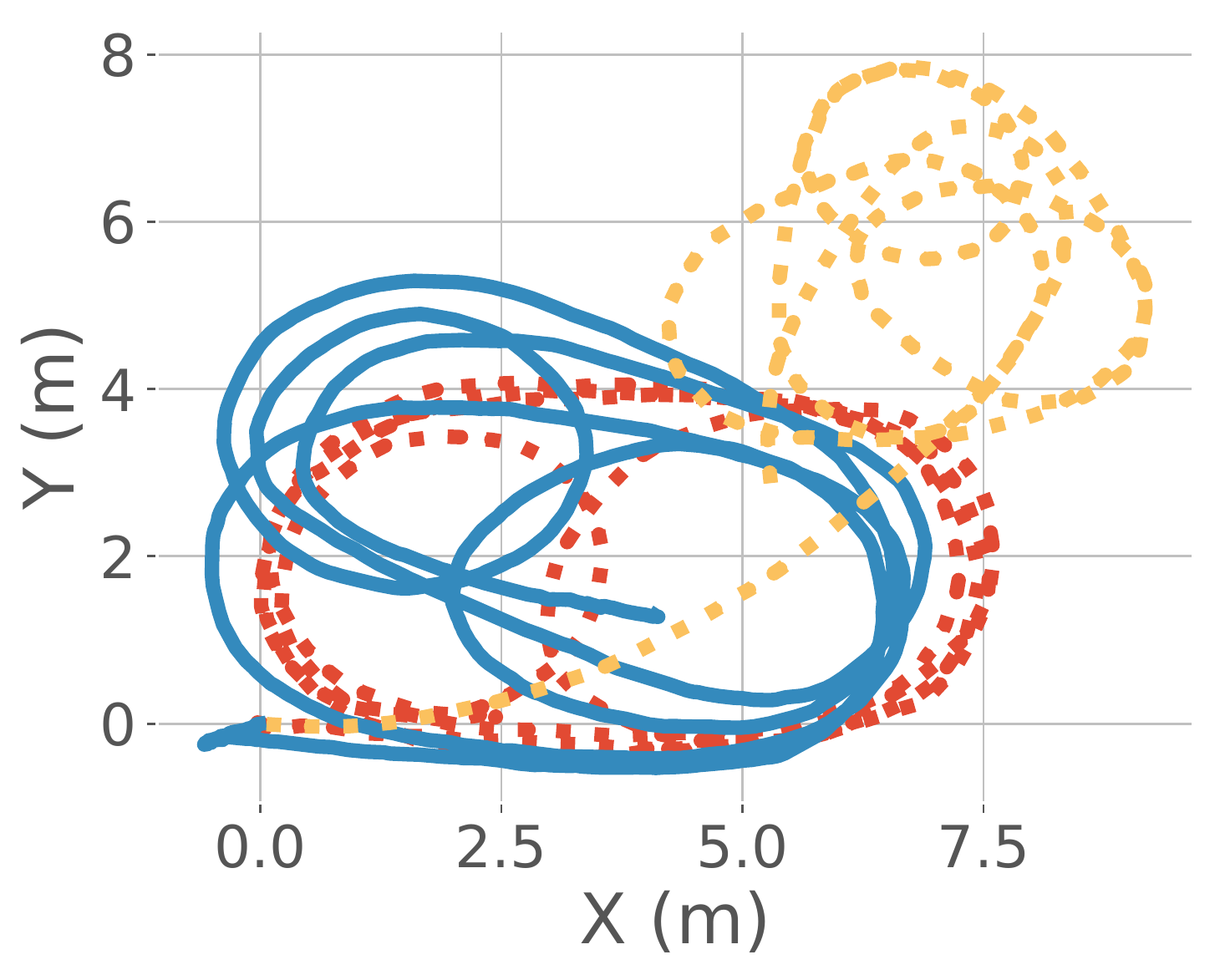} 
		\caption{Atrium Trajectory}
	\end{subfigure}%
	
	\begin{subfigure}[b]{0.24\textwidth}\centering
		\includegraphics[width=\columnwidth]{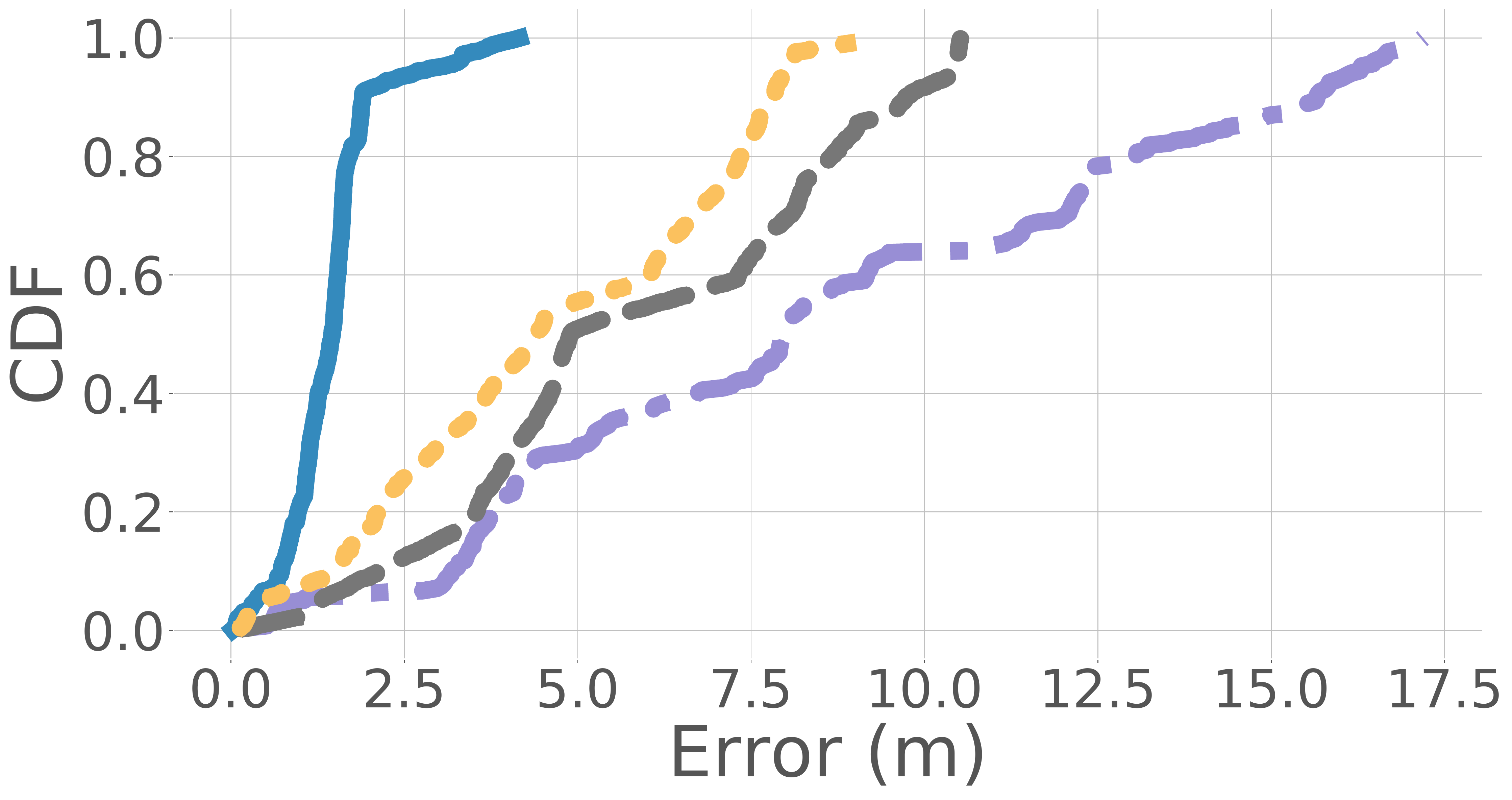} 
		\caption{Corridor CDF}
	\end{subfigure}%
	\begin{subfigure}[b]{0.24\textwidth}\centering
		\includegraphics[width=\columnwidth]{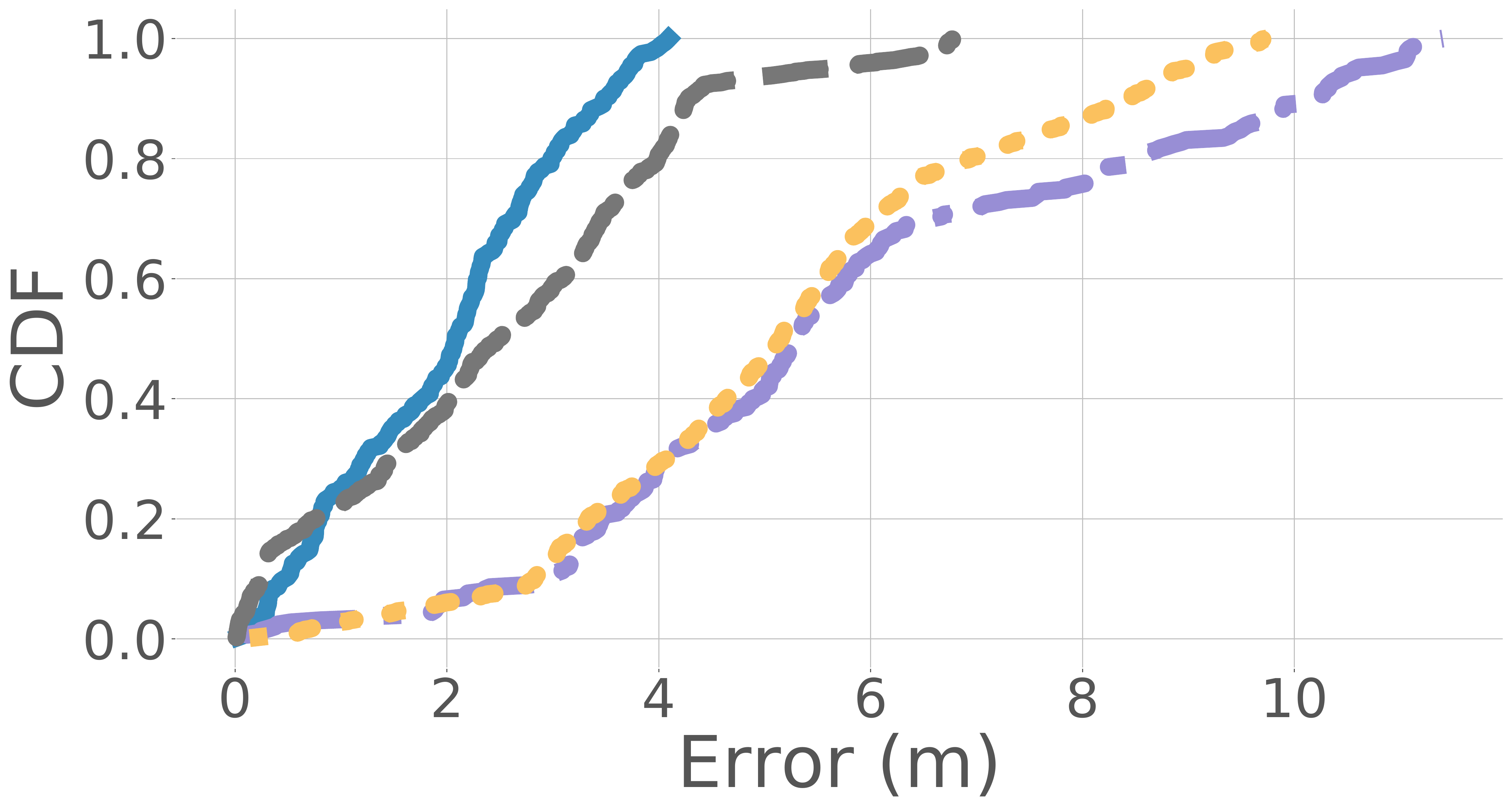} 
		\caption{Atrium CDF}
	\end{subfigure}%	
	
\caption{Two example testing sequences with handheld data. For clarity, we only show the trajectory for VINET in the top-row as the most competitive alternative.}
\label{fig:handheld_cdf}
\end{figure}

As shown in Fig.~\ref{fig:handheld_cdf}, our method copes well with different levels of trajectory complexity and constantly provides accurate odometry estimation. This confirms the versatility of our proposed framework to different mobility patterns and constraints.

\begin{table}[!t]
\small
\caption{Overall results with a handheld device.}
\label{tab:hand_overall}
\centering
\begin{tabular}{|c|c|r|c|c|c|}
\hline
\multicolumn{2}{|c|}{\textbf{Sensors}} & \multicolumn{1}{c|}{\textbf{\begin{tabular}[c]{@{}c@{}}IMU\\ Only\end{tabular}}} & \multicolumn{2}{c|}{\textbf{mmWave + IMU}} & \multicolumn{1}{c|}{\textbf{\begin{tabular}[c]{@{}c@{}}RGB\\ + IMU\end{tabular}}} \\ \hline
\multicolumn{2}{|c|}{\textbf{Method}} & \multicolumn{1}{c|}{IONET} & \multicolumn{1}{c|}{IMU-ICP} & \multicolumn{1}{c|}{\sysname} & \multicolumn{1}{c|}{VINET} \\ \hline
\multirow{3}{*}{\textbf{\begin{tabular}[c]{@{}c@{}}3\\ D\end{tabular}}} & Mean & 3.452 & 5.843 & \textbf{1.857} & 4.044 \\ \cline{2-6} 
 & Std. & 1.769 & 2.966 & \textbf{1.026} & 1.847 \\ \cline{2-6} 
 & Max & 7.071 & 11.712 & \textbf{4.193} & 7.734 \\ \hline
\multirow{3}{*}{\textbf{\begin{tabular}[c]{@{}c@{}}2\\ D\end{tabular}}} & Mean & 3.406 & 4.979 & \textbf{1.459} & 3.741 \\ \cline{2-6} 
 & Std. & 1.717 & 2.78 & \textbf{0.815} & 1.854 \\ \cline{2-6} 
 & Max & 7.055 & 10.928 & \textbf{3.990} & 7.424 \\ \hline
\end{tabular}
\end{table}

\subsubsection{Impact of mmWave Sampling Rate} % (fold)
\label{ssub:impact_of_mmwave_sampling_rate}

\begin{figure}[!t]
	\centering
	\begin{subfigure}[b]{0.23\textwidth}\centering
		\includegraphics[width=\columnwidth]{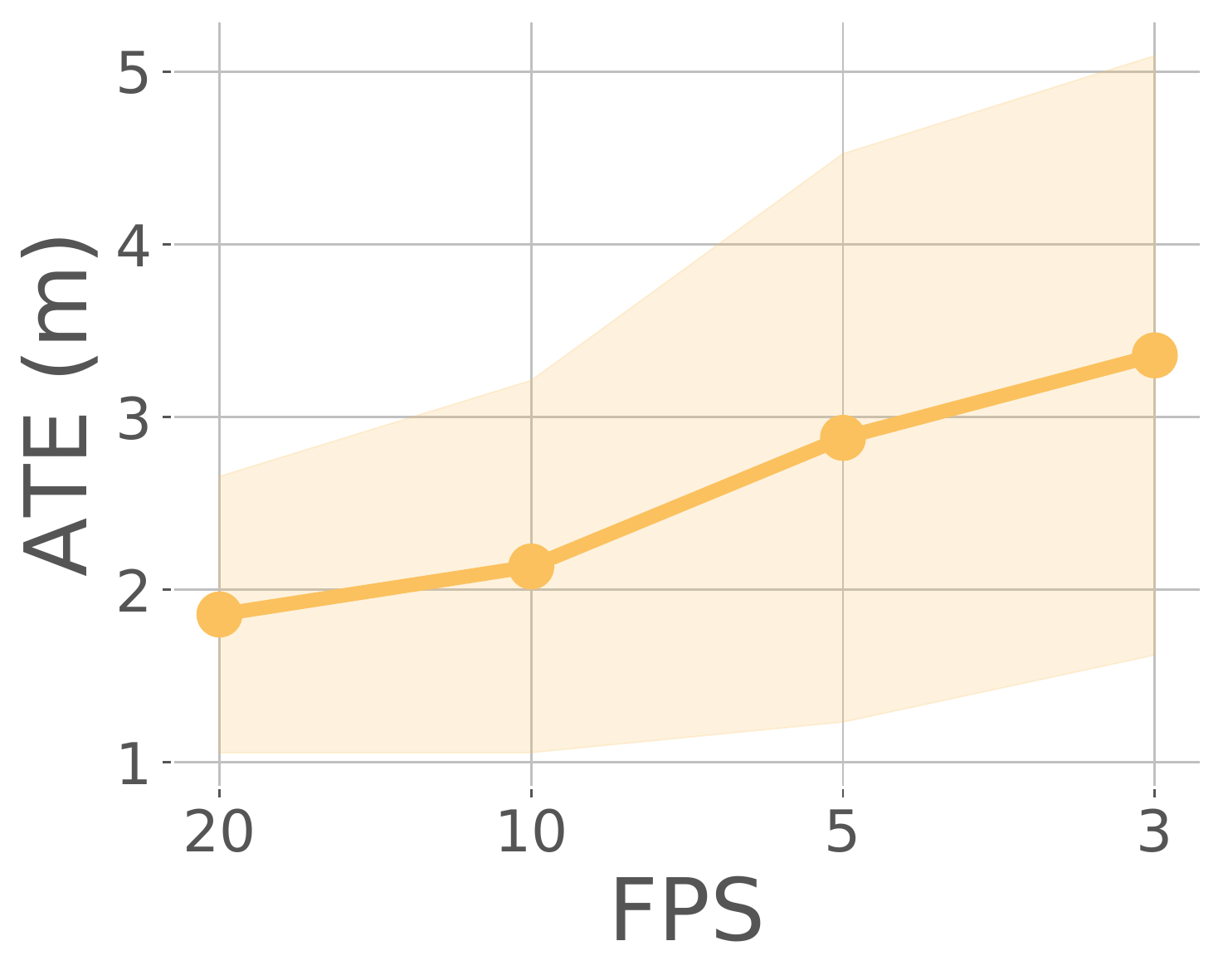} 
		\caption{3D}
	\end{subfigure}%
	\hfill
	\begin{subfigure}[b]{0.23\textwidth}\centering
		\includegraphics[width=\columnwidth]{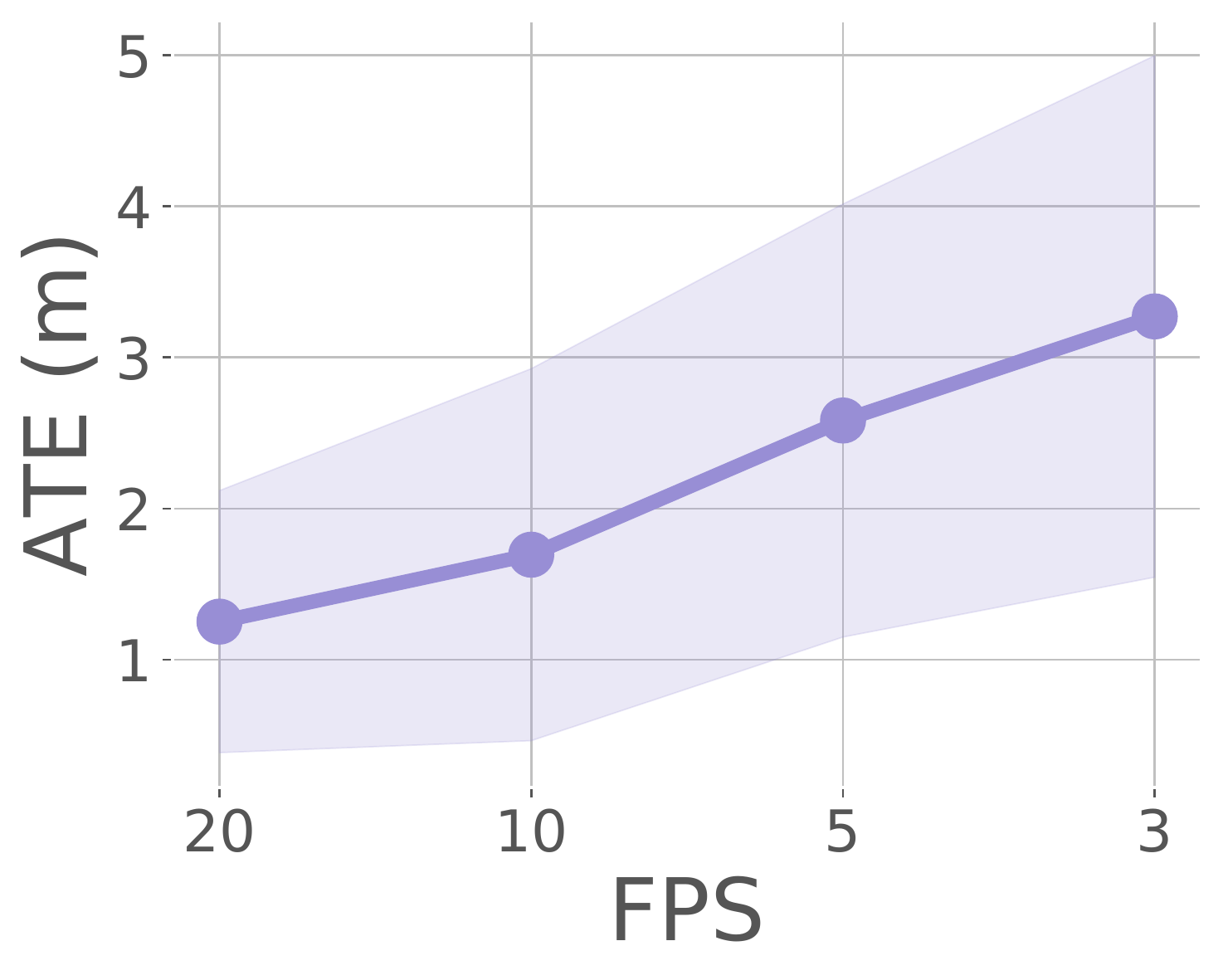} 
		\caption{2D}
	\end{subfigure}%
\caption{Impact of sampling rate with handheld data.}
\label{fig:sp_rate}
\end{figure}

The key to accurate egomotion estimation lies in the extent of \emph{parallax/baseline} which is the distance between successive frames when moving~\cite{saputra2018visual}. Intuitively, when a fast-moving platform is equipped with a low frame-per-second (FPS) sensor, the parallax is large, resulting in less information correspondence, and ultimately a large error. Due to the  low-quality mmWave point cloud, an adequate sampling rate is vital.  Fig.~\ref{fig:sp_rate} illustrates the ATE of \sysname trained with different mmWave sampling rates. As can be seen, our model gives the best performance with a $20$ FPS rate, with consistent performance decline with lower rates, indicating the importance of an adequate sampling rate. Notably, the $20$ FPS adopted here is 5-fold as large as the robot's $4$ FPS, which is reasonable due to difference in speed between the robot and a human walking.
In practice, we suggest end users set the the best FPS based on the expected platform dynamics.

\subsubsection{Generalization to Unseen Environments} % (fold)
\label{ssub:generalization_to_unseen_environment}

\begin{figure}[!t]
	\centering
	\begin{subfigure}[b]{0.23\textwidth}\centering
		\includegraphics[width=\columnwidth]{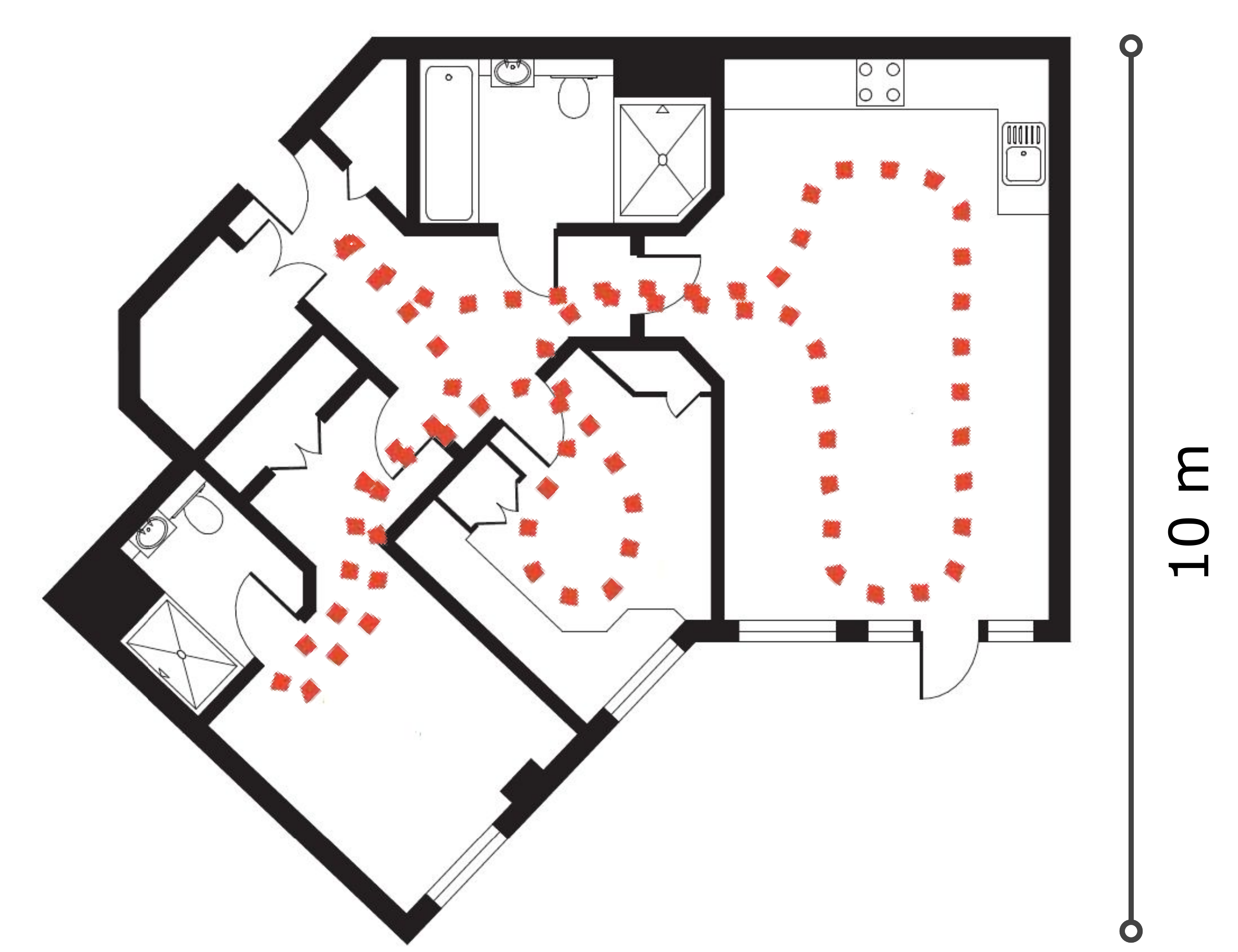} 
		\caption{True trajectory (red)}
	\end{subfigure}%
	\hfill
	\begin{subfigure}[b]{0.23\textwidth}\centering
		\includegraphics[width=\columnwidth]{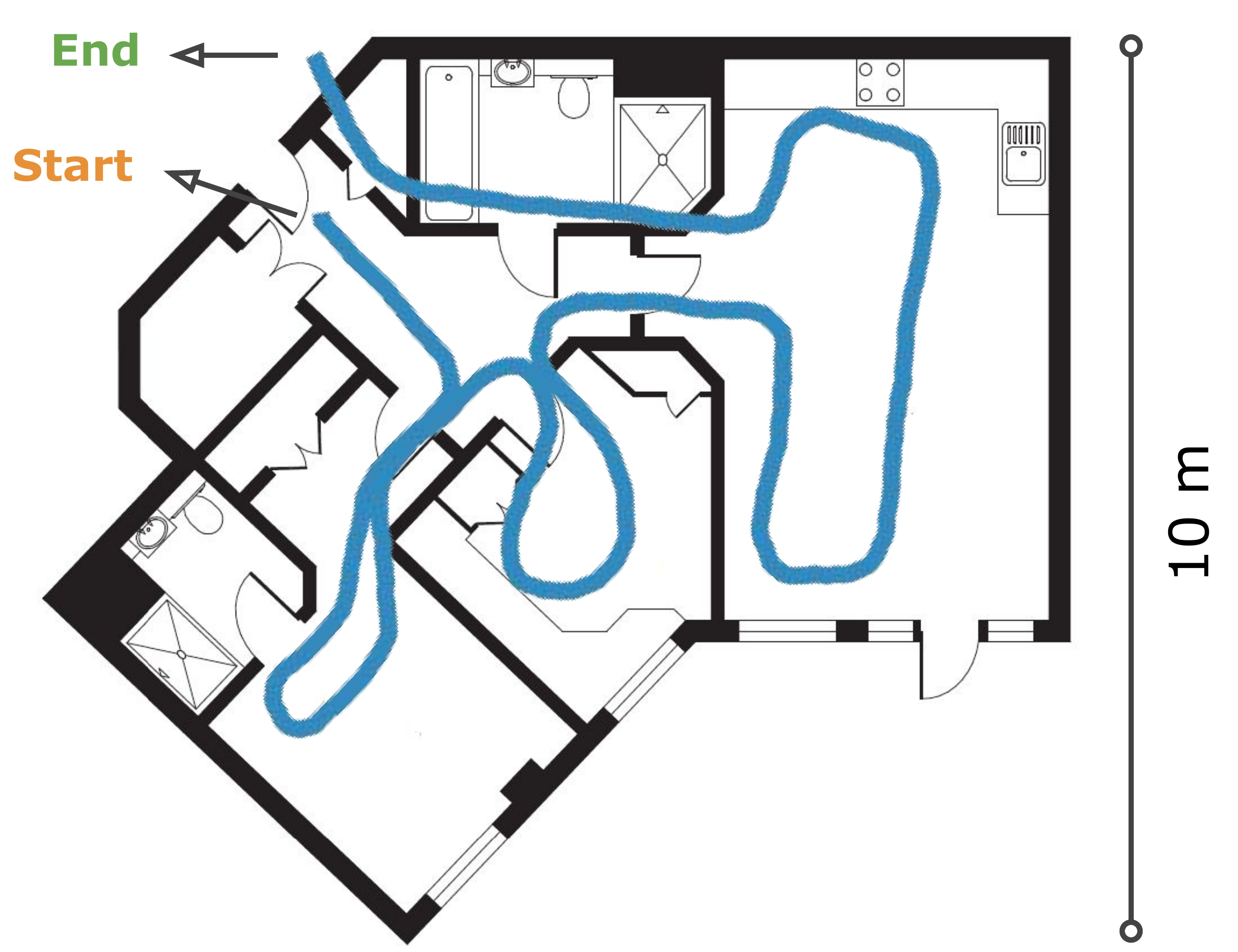} 
		\caption{\sysname estimation (blue)}
	\end{subfigure}%
\caption{\revise{\sysname examined in an \emph{unseen} household environment operating in hand-held mode. Note that the device was carried by a taller user \emph{not} involved in the original training data collection.}}
\label{fig:generalization_dai}
\end{figure}

We further evaluate the generalization ability of \sysname in unseen scenarios. Although the handheld \sysname model was trained on data collected from commercial buildings, we examine its egomotion estimation performance in an unseen household environment, where the layout and floor plan are significantly different to the training case. To fully validate the generalization, a new user (height of 185cm) carried the handheld device rather than the two shorter users in the training set (166cm and 178cm respectively). Fig.~\ref{fig:generalization_dai} shows the trajectory estimated by \sysname. Despite the previously unseen environment, \sysname demonstrates an impressive generalization performance: the mean ATE is only $0.588$m with the largest ATE being only $1.964$m. Note that in contrast deep visual odometry in particular can be poor at adapting to new environments with different visual features. By inspecting Fig.~\ref{fig:generalization_dai}, we found that the main trajectory is well revealed, implying a very good precursor for downstream tasks such as SLAM or map-matching based localization \cite{xiao2014lightweight}.

\subsection{System Efficiency} % (fold)
\label{sub:system_efficiency}

In the last experiment, we implement and compare \sysname on two platforms: an NVIDIA Jetson TX2 (TX2) and a mini netbook. For the implementation, we use TensorFlow Lite \cite{alsing2018mobile} to compress our models as per the convention of efficient on-device inference of DNNs. We focus on the comparison with VINET, which is a pervasive DNN solution for estimating egomotion on resource-constrained platforms.
Tab.~\ref{tab:running_time} suggests that our proposed \sysname is significantly faster (17 FPS on TX2) than VINET (3 FPS on TX2). This is because the transformed mmWave panoramic images are much smaller and hence more data-efficient than RGB images. Note also that compared with the single-stage attention strategy, \sysname is slightly faster as it divides the single, large mask generation process into several lightweight branches (see \sect\ref{sub:discussion_of_mixed_attention}). Concerning the model size (in .tflite format), \sysname is only $107$MB as opposed to the $171$MB for single-stage attention and $764$MB of VINET. This further validates \sysname's efficiency and suitability for more resource constrained platforms.

% \begin{table}[t]
% \small
% \caption{Runtime Analysis (Unit: Second).}
% \label{tab:running_time}
% \centering
% \begin{tabular}{|c|c|c|c|}
% \hline
% \textbf{} & \textbf{milliEgo} & \textbf{\begin{tabular}[c]{@{}c@{}}Single-Stage \\ Att. \cite{saputra2020deeptio,chen2018ionet} \end{tabular}} & \textbf{VINET} \\ \hline
% \textbf{TX2} & 0.0588 $\pm$ 0.0028 & 0.0595 $\pm$ 0.0027 & 0.2392 $\pm$ 0.0027 \\ \hline
% \textbf{Notebook} & 0.484 $\pm$ 0.009 & 0.688 $\pm$ 0.017 & 10.398 $\pm$ 0.035 \\ \hline
% \end{tabular}
% \end{table}

\begin{table}[t]
\small
\caption{Runtime Analysis (Unit: Second).}
\label{tab:running_time}
\centering
\begin{tabular}{|c|c|c|}
\hline
\textbf{} & \textbf{milliEgo} & \textbf{VINET} \\ \hline
\textbf{TX2} & 0.0588 $\pm$ 0.0028 & 0.2392 $\pm$ 0.0027 \\ \hline
\textbf{Notebook} & 0.484 $\pm$ 0.009 & 10.398 $\pm$ 0.035 \\ \hline
\end{tabular}
\end{table}

% \begin{figure}[!t]
% 	\centering
% 	\begin{subfigure}[b]{0.23\textwidth}\centering
% 		\includegraphics[width=\columnwidth]{figures/runtime/runtime_tx2.pdf} 
% 		\caption{TX2}
% 	\end{subfigure}%
% 	\hfill
% 	\begin{subfigure}[b]{0.23\textwidth}\centering
% 		\includegraphics[width=\columnwidth]{figures/runtime/runtime_notebook.pdf} 
% 		\caption{Notebook}
% 	\end{subfigure}%
% \caption{Runtime Analysis.}
% \label{fig:runtime}
% \end{figure}

%!TEX root = ../main.tex
\section{Related Work} % (fold)
\label{sec:related_work}

% general RF applications and those with mmWave tech.
% \noindent \textbf{RF-based Motion Tracking and Sensing}.
% One of the most vibrant research directions in mobile sensing is leveraging the RF waves for motion tracking and indoor localization. In the WiFi bands, prior art have used commodity WiFi chips \cite{depatla2015x,xie2019md,pu2013whole,liu2012push,jiang2013hallway,jiang2018towards,jiang2020towards,wu2019sigcomm_rim} to localize robots/humans, track their poses, or recognize hand gestures. By using a carefully customized FMCW radar \cite{adib2015multi,adib20143d,zhao2018rf,zhao2019through}, it is proved that RF signals in the WiFi band can be also used to accurately track/imagine human body dynamics, as well as recover pose estimation under NLOS scenarios. 
% With regards to mmWave itself, Babak et al. \cite{mamandipoor201460} uses FMCW hardware and applies SAR with sparse measurements in absence of device movement while Wei et al. \cite{wei2015mtrack} uses three separate, static 60GHz radios to track pen movement in the near field. In related areas, mmWave signals \cite{zhu201560ghz,zhu2017object,zhu2015reusing,wei2017facilitating,zhou2019autonomous,lu2020millimap} are also found effective for other location-based services such as mapping and imaging objects. Compared with the above works, our work is the first framework that can estimate the $6$-DoF odometry by leveraging one commodity single-chip mmWave radar and demonstrates even more robustness when combined  with other sensors.

\revise{
\noindent \textbf{Radar based Localization and Tracking}.
Prior work in this thread can be broadly divided into three categories based on the class of radar used: (1) surveillance radar, (2) mechanical scanning radar and (3) single-chip mmWave radar. Multistatic or bistatic surveillance radar is one of the most well-known technologies in the radar community and has been widely used in military for aircraft tracking \cite{caballero2018multiple,petsios2008solving,mamandipoor201460}. These are long range, high-power with custom designed signal processing chains. In the second class, to meet the mobility needs of automobiles, mechanical scanning radars (e.g., Navtech CTS350-X) are used for odometry estimation \cite{cen2018precise,park2020pharao}. However, as discussed in \cite{lu2020millimap}, mechanical radars are overly heavy and cumbersome for service robots or wearable devices. Single-chip mmWave radars have recently emerge as a low-cost and small-factor solution to the above radars and have been used in indoor mapping \cite{lu2020millimap} and ego-velocity estimation \cite{kramer2020radar}. Compared with prior art, our work uses the same single-chip radar but focuses on different tasks. As a result, \sysname has its own technical challenge in a new context and leverages a multi-sensor solution to robustly estimate the trajectory of a mobile platform.
}

\noindent \textbf{Deep Odometry Model}.
Owning to recent advances in data-driven methods, deep odometry models have gained attention in both academia and industry. A multitude of such models have been proposed and are widely used with different sensors, including RGB camera \cite{wang2017deepvo,li2018undeepvo}, depth camera \cite{yang2018deep,ummenhofer2017demon}, IMU \cite{chen2018ionet,chen2019deep} and lidar \cite{nicolai2016deep,wang2019deeppco}. A prominent feature of deep odometry lies in its end-to-end learning, which enables the modeling of many complex physical processes and disturbances \cite{wang2018end}. Beside this benefit, using deep odometry models can enable the system to skip the tedious calibration phase, increasing the adoption range even for non-domain-experts. Our work also leverage the superior modeling power of deep odometry model. Compared with the above works, we develop the first deep mmWave odometry \sysname with the low-cost single chip radar, and systematically examined its limitation and potential to assist other odometry sensors.

\noindent \textbf{Multi-modal Sensor Fusion}
Our work also falls within the scope of multi-modal sensor fusion, a technique that combines the information from multiple sensor modalities in order to provide better system performance. Different to traditional sensor fusion methods (e.g., probabilistic fusion, fuzzy reasoning, hybrid fusion, etc \cite{chandrasekaran2017survey,xu2015idyll,almalioglu2020milli}), neural network based sensor fusion methods are able to fuse heterogeneous sensor data at all levels ranging from raw data, to intermediate features, to individual decisions. This uniqueness makes deep sensor fusion an emerging topic in recent years. Early works \cite{yao2017deepsense,eitel2015multimodal} in this vein treat various sensor modalities equally in deep neural networks yet ignore their heterogeneous sensing qualities. By incorporating self-attention mechanisms \cite{vaswani2017attention} or similar concept \cite{wang2018non} in the network, recent efforts \cite{yao2019sadeepsense,liu2020giobalfusion,xue2019deepfusion,xue2020deepmv
} made progress in dynamically understanding the importance of inputs or features and robustly recognize human activity. In the domain of egomotion estimation, such a single-stage attention mechanism has also been used in multi-modal odometry to adaptively combine inertial sensors with RGB or thermal imaging cameras (e.g., SelectiveFusion \cite{chen2019selective} and DeepTIO \cite{saputra2020deeptio}). \sysname differs from the above work in several aspects.
First, \sysname is a two-stage attention fusion. It not only uses a self-attention layer to prioritize intra-sensor features, but further proposes a cross-attention layer based on the Spence's spatial perception model \cite{spence2004crossmodal} that enforces complementary inter-sensor behaviors towards achieving greater robustness. As shown by the experiment results, \sysname significantly outperforms \cite{chen2019selective,saputra2020deeptio} for egomotion estimation and yield more explainable behaviors between mmWave radars and other sensors. Secondly, \sysname is the first work that end-to-end fuses a low-cost mmWave radar with other sensors in a neural network.

%!TEX root = ../main.tex
\section{Limitations and Future Work} % (fold)
\label{sec:limitation_and_future_work}

This work focuses on a proof-of-principle egomotion estimation assisted by single-chip mmWave radar, towards our vision of enabling a high degree of spatial awareness with low-cost mobile sensing systems. There are limitations and a number of avenues for future exploration. 
\revise{Firstly, as our design and experiments have mainly focused on indoor commercial buildings, it remains unknown how \sysname will generalize to other indoor spaces (e.g, hospitals, museums and shopping malls) or even outdoor scenarios. This will likely require more diverse data collection for account for the distinct noise in different environments.}
The optimal range settings will also need to be investigated as objects are likely to be more widely dispersed. Secondly, trials on UAVs or drone platforms are needed to investigate if their rapid flying speed, long parallax and less constrained motion will cause new estimation challenges. Thirdly, as with any kind of odometry, \sysname still drifts. Therefore an obvious next step is to detect and use mmWave loop closures to refine \sysname, so that a robust system for simultaneous localization and mapping can be achieved. \revise{In addition, although we have investigated an end-to-end system e.g. using a Jetson TX2 embedded platform for real-time odometry, there is significant room for further optimization and testing the energy efficiency on mobile devices.} Lastly, it would be interesting to explore how current and emerging 60GHz MIMO WiFi standards e.g. 802.11ay/802.11az \cite{zhao2020m} can be adapted to give similar spatial information. \revise{More powerful MIMO mmWave chipsets with higher spatial and velocity resolution will also be investigated for tracking and egomotion estimation.}

% section limitation_and_future_work (end)

%!TEX root = ../main.tex
\section{Conclusion} % (fold)
\label{sec:conclusion}

This paper presents \sysname, a novel low-cost mmWave radar assisted odometry system that can robustly estimate egomotion together with an IMU. 
To address the limitation brought by the low-cost radar, \sysname proposed to use a mmWave odometry subnet to directly learn the motion transformation from noisy data. \sysname also features a cross-attention based multi-modal fusion  that effectively combines mmWave radar with inertial and other potentially available sensors. Extensive experiments were conducted with real-time prototype implementation on both robots and handheld devices. 
\sysname opens up mmWave-assisted sensing to new applications where egomotion is crucial, such as robot tracking, VR/AR/MR gaming and smartphone indoor navigation. 
% section conclusion (end)
%%%%%%%%%%%%%%%%%%%%%%%%%%%%%%%%%%%%%%%%%%%%%%%%%%%%%%%%%%%%%%%%%%%%%%%%%%%%%%%%

% \section{Acknowledgements}
% This work has been partly supported by the EPSRC Program Grant ``Mobile Robotics: Enabling a Pervasive Technology of the Future (GoW EP/M019918/1)".  

% \clearpage 
\bibliographystyle{ACM-Reference-Format}
\bibliography{ref}


\begin{thebibliography}{77}

%%% the Web address.

\ifx \showCODEN    \undefined \def \showCODEN     #1{\unskip}     \fi
\ifx \showDOI      \undefined \def \showDOI       #1{#1}\fi
\ifx \showISBNx    \undefined \def \showISBNx     #1{\unskip}     \fi
\ifx \showISBNxiii \undefined \def \showISBNxiii  #1{\unskip}     \fi
\ifx \showISSN     \undefined \def \showISSN      #1{\unskip}     \fi
\ifx \showLCCN     \undefined \def \showLCCN      #1{\unskip}     \fi
\ifx \shownote     \undefined \def \shownote      #1{#1}          \fi
\ifx \showarticletitle \undefined \def \showarticletitle #1{#1}   \fi
\ifx \showURL      \undefined \def \showURL       {\relax}        \fi
% invisible to TeX
\providecommand\bibfield[2]{#2}
\providecommand\bibinfo[2]{#2}
\providecommand\natexlab[1]{#1}
\providecommand\showeprint[2][]{arXiv:#2}

\bibitem[\protect\citeauthoryear{Aghili and Su}{Aghili and Su}{2016}]%
        {aghili2016robust}
\bibfield{author}{\bibinfo{person}{Farhad Aghili} {and}
  \bibinfo{person}{Chun-Yi Su}.} \bibinfo{year}{2016}\natexlab{}.
\newblock \showarticletitle{Robust relative navigation by integration of ICP
  and adaptive Kalman filter using laser scanner and IMU}.
\newblock \bibinfo{journal}{\emph{IEEE/ASME Transactions on Mechatronics}}
  \bibinfo{volume}{21}, \bibinfo{number}{4} (\bibinfo{year}{2016}),
  \bibinfo{pages}{2015--2026}.
\newblock


\bibitem[\protect\citeauthoryear{Agrawal, Nair, Abbeel, Malik, and
  Levine}{Agrawal et~al\mbox{.}}{2016}]%
        {agrawal2016learning}
\bibfield{author}{\bibinfo{person}{Pulkit Agrawal}, \bibinfo{person}{Ashvin~V
  Nair}, \bibinfo{person}{Pieter Abbeel}, \bibinfo{person}{Jitendra Malik},
  {and} \bibinfo{person}{Sergey Levine}.} \bibinfo{year}{2016}\natexlab{}.
\newblock \showarticletitle{Learning to poke by poking: Experiential learning
  of intuitive physics}. In \bibinfo{booktitle}{\emph{Advances in neural
  information processing systems}}.
\newblock


\bibitem[\protect\citeauthoryear{Almalioglu, Turan, Lu, Trigoni, and
  Markham}{Almalioglu et~al\mbox{.}}{2020}]%
        {almalioglu2020milli}
\bibfield{author}{\bibinfo{person}{Yasin Almalioglu}, \bibinfo{person}{Mehmet
  Turan}, \bibinfo{person}{Chris~Xiaoxuan Lu}, \bibinfo{person}{Niki Trigoni},
  {and} \bibinfo{person}{Andrew Markham}.} \bibinfo{year}{2020}\natexlab{}.
\newblock \showarticletitle{Milli-RIO: Ego-Motion Estimation with Low-Cost
  Millimetre-Wave Radar}.
\newblock \bibinfo{journal}{\emph{IEEE Sensors Journal}}
  (\bibinfo{year}{2020}).
\newblock


\bibitem[\protect\citeauthoryear{Alsing}{Alsing}{2018}]%
        {alsing2018mobile}
\bibfield{author}{\bibinfo{person}{Oscar Alsing}.}
  \bibinfo{year}{2018}\natexlab{}.
\newblock \bibinfo{title}{Mobile Object Detection using TensorFlow Lite and
  Transfer Learning}.
\newblock
\newblock


\bibitem[\protect\citeauthoryear{Alt, Rives, and Steinbach}{Alt
  et~al\mbox{.}}{2013}]%
        {alt2013reconstruction}
\bibfield{author}{\bibinfo{person}{Nicolas Alt}, \bibinfo{person}{Patrick
  Rives}, {and} \bibinfo{person}{Eckehard Steinbach}.}
  \bibinfo{year}{2013}\natexlab{}.
\newblock \showarticletitle{Reconstruction of transparent objects in
  unstructured scenes with a depth camera}. In \bibinfo{booktitle}{\emph{2013
  IEEE International Conference on Image Processing}}.
  \bibinfo{pages}{4131--4135}.
\newblock


\bibitem[\protect\citeauthoryear{Bouaziz, Tagliasacchi, and Pauly}{Bouaziz
  et~al\mbox{.}}{2013}]%
        {bouaziz2013sparse}
\bibfield{author}{\bibinfo{person}{Sofien Bouaziz}, \bibinfo{person}{Andrea
  Tagliasacchi}, {and} \bibinfo{person}{Mark Pauly}.}
  \bibinfo{year}{2013}\natexlab{}.
\newblock \showarticletitle{Sparse iterative closest point}. In
  \bibinfo{booktitle}{\emph{Computer graphics forum}},
  Vol.~\bibinfo{volume}{32}. \bibinfo{pages}{113--123}.
\newblock


\bibitem[\protect\citeauthoryear{Caballero, Cantu, Rodriguez, Gonzales,
  Castellanos, Cantu, Strait, Son, and Kim}{Caballero et~al\mbox{.}}{[n. d.]}]%
        {caballero2018multiple}
\bibfield{author}{\bibinfo{person}{Carlos~Pena Caballero},
  \bibinfo{person}{Elifaleth Cantu}, \bibinfo{person}{Jesus Rodriguez},
  \bibinfo{person}{Adolfo Gonzales}, \bibinfo{person}{Osvaldo Castellanos},
  \bibinfo{person}{Angel Cantu}, \bibinfo{person}{Megan Strait},
  \bibinfo{person}{Jae Son}, {and} \bibinfo{person}{Dongchul Kim}.}
  \bibinfo{year}{[n. d.]}\natexlab{}.
\newblock \showarticletitle{A Multiple Radar Approach for Automatic Target
  Recognition of Aircraft using Inverse Synthetic Aperture Radar}. In
  \bibinfo{booktitle}{\emph{2018 1st International Conference on Data
  Intelligence and Security (ICDIS)}}.
\newblock


\bibitem[\protect\citeauthoryear{Cadena, Carlone, Carrillo, Latif, Scaramuzza,
  Neira, Reid, and Leonard}{Cadena et~al\mbox{.}}{2016}]%
        {cadena2016past}
\bibfield{author}{\bibinfo{person}{Cesar Cadena}, \bibinfo{person}{Luca
  Carlone}, \bibinfo{person}{Henry Carrillo}, \bibinfo{person}{Yasir Latif},
  \bibinfo{person}{Davide Scaramuzza}, \bibinfo{person}{Jos{\'e} Neira},
  \bibinfo{person}{Ian Reid}, {and} \bibinfo{person}{John~J Leonard}.}
  \bibinfo{year}{2016}\natexlab{}.
\newblock \showarticletitle{Past, present, and future of simultaneous
  localization and mapping: Toward the robust-perception age}.
\newblock \bibinfo{journal}{\emph{IEEE Transactions on robotics}}
  (\bibinfo{year}{2016}).
\newblock


\bibitem[\protect\citeauthoryear{Cen and Newman}{Cen and Newman}{2018}]%
        {cen2018precise}
\bibfield{author}{\bibinfo{person}{Sarah~H Cen} {and} \bibinfo{person}{Paul
  Newman}.} \bibinfo{year}{2018}\natexlab{}.
\newblock \showarticletitle{Precise ego-motion estimation with millimeter-wave
  radar under diverse and challenging conditions}. In
  \bibinfo{booktitle}{\emph{2018 IEEE International Conference on Robotics and
  Automation (ICRA)}}.
\newblock


\bibitem[\protect\citeauthoryear{Chandrasekaran, Gangadhar, and
  Conrad}{Chandrasekaran et~al\mbox{.}}{2017}]%
        {chandrasekaran2017survey}
\bibfield{author}{\bibinfo{person}{Balasubramaniyan Chandrasekaran},
  \bibinfo{person}{Shruti Gangadhar}, {and} \bibinfo{person}{James~M Conrad}.}
  \bibinfo{year}{2017}\natexlab{}.
\newblock \showarticletitle{A survey of multisensor fusion techniques,
  architectures and methodologies}. In \bibinfo{booktitle}{\emph{SoutheastCon
  2017}}. IEEE, \bibinfo{pages}{1--8}.
\newblock


\bibitem[\protect\citeauthoryear{Chen, Lu, Markham, and Trigoni}{Chen
  et~al\mbox{.}}{2018}]%
        {chen2018ionet}
\bibfield{author}{\bibinfo{person}{Changhao Chen}, \bibinfo{person}{Xiaoxuan
  Lu}, \bibinfo{person}{Andrew Markham}, {and} \bibinfo{person}{Niki Trigoni}.}
  \bibinfo{year}{2018}\natexlab{}.
\newblock \showarticletitle{Ionet: Learning to cure the curse of drift in
  inertial odometry}. In \bibinfo{booktitle}{\emph{Thirty-Second AAAI
  Conference on Artificial Intelligence}}.
\newblock


\bibitem[\protect\citeauthoryear{Chen, Lu, Wahlstrom, Markham, and
  Trigoni}{Chen et~al\mbox{.}}{2019a}]%
        {chen2019deep}
\bibfield{author}{\bibinfo{person}{Changhao Chen}, \bibinfo{person}{Xiaoxuan
  Lu}, \bibinfo{person}{Johan Wahlstrom}, \bibinfo{person}{Andrew Markham},
  {and} \bibinfo{person}{Niki Trigoni}.} \bibinfo{year}{2019}\natexlab{a}.
\newblock \showarticletitle{Deep Neural Network Based Inertial Odometry Using
  Low-cost Inertial Measurement Units}.
\newblock \bibinfo{journal}{\emph{IEEE Transactions on Mobile Computing}}
  (\bibinfo{year}{2019}).
\newblock


\bibitem[\protect\citeauthoryear{Chen, Rosa, Miao, Lu, Wu, Markham, and
  Trigoni}{Chen et~al\mbox{.}}{2019b}]%
        {chen2019selective}
\bibfield{author}{\bibinfo{person}{Changhao Chen}, \bibinfo{person}{Stefano
  Rosa}, \bibinfo{person}{Yishu Miao}, \bibinfo{person}{Chris~Xiaoxuan Lu},
  \bibinfo{person}{Wei Wu}, \bibinfo{person}{Andrew Markham}, {and}
  \bibinfo{person}{Niki Trigoni}.} \bibinfo{year}{2019}\natexlab{b}.
\newblock \showarticletitle{Selective sensor fusion for neural visual-inertial
  odometry}. In \bibinfo{booktitle}{\emph{Proceedings of the IEEE Conference on
  Computer Vision and Pattern Recognition}}.
\newblock


\bibitem[\protect\citeauthoryear{Civera, Grasa, Davison, and Montiel}{Civera
  et~al\mbox{.}}{2010}]%
        {civera20101}
\bibfield{author}{\bibinfo{person}{Javier Civera}, \bibinfo{person}{Oscar~G
  Grasa}, \bibinfo{person}{Andrew~J Davison}, {and} \bibinfo{person}{JMM
  Montiel}.} \bibinfo{year}{2010}\natexlab{}.
\newblock \showarticletitle{1-Point RANSAC for extended Kalman filtering:
  Application to real-time structure from motion and visual odometry}.
\newblock \bibinfo{journal}{\emph{Journal of field robotics}}
  \bibinfo{volume}{27}, \bibinfo{number}{5} (\bibinfo{year}{2010}),
  \bibinfo{pages}{609--631}.
\newblock


\bibitem[\protect\citeauthoryear{Clark, Wang, Wen, Markham, and Trigoni}{Clark
  et~al\mbox{.}}{2017}]%
        {clark2017vinet}
\bibfield{author}{\bibinfo{person}{Ronald Clark}, \bibinfo{person}{Sen Wang},
  \bibinfo{person}{Hongkai Wen}, \bibinfo{person}{Andrew Markham}, {and}
  \bibinfo{person}{Niki Trigoni}.} \bibinfo{year}{2017}\natexlab{}.
\newblock \showarticletitle{Vinet: Visual-inertial odometry as a
  sequence-to-sequence learning problem}. In
  \bibinfo{booktitle}{\emph{Thirty-First AAAI Conference on Artificial
  Intelligence}}.
\newblock


\bibitem[\protect\citeauthoryear{Dahl, Sainath, and Hinton}{Dahl
  et~al\mbox{.}}{2013}]%
        {dahl2013improving}
\bibfield{author}{\bibinfo{person}{George~E Dahl}, \bibinfo{person}{Tara~N
  Sainath}, {and} \bibinfo{person}{Geoffrey~E Hinton}.}
  \bibinfo{year}{2013}\natexlab{}.
\newblock \showarticletitle{Improving deep neural networks for LVCSR using
  rectified linear units and dropout}. In \bibinfo{booktitle}{\emph{2013 IEEE
  international conference on acoustics, speech and signal processing}}.
  \bibinfo{pages}{8609--8613}.
\newblock


\bibitem[\protect\citeauthoryear{DeTone, Malisiewicz, and Rabinovich}{DeTone
  et~al\mbox{.}}{2018}]%
        {detone2018superpoint}
\bibfield{author}{\bibinfo{person}{Daniel DeTone}, \bibinfo{person}{Tomasz
  Malisiewicz}, {and} \bibinfo{person}{Andrew Rabinovich}.}
  \bibinfo{year}{2018}\natexlab{}.
\newblock \showarticletitle{Superpoint: Self-supervised interest point
  detection and description}. In \bibinfo{booktitle}{\emph{Proceedings of the
  IEEE Conference on Computer Vision and Pattern Recognition Workshops}}.
  \bibinfo{pages}{224--236}.
\newblock


\bibitem[\protect\citeauthoryear{Dosovitskiy, Fischer, Ilg, Hausser, Hazirbas,
  Golkov, Van Der~Smagt, Cremers, and Brox}{Dosovitskiy et~al\mbox{.}}{2015}]%
        {dosovitskiy2015flownet}
\bibfield{author}{\bibinfo{person}{Alexey Dosovitskiy},
  \bibinfo{person}{Philipp Fischer}, \bibinfo{person}{Eddy Ilg},
  \bibinfo{person}{Philip Hausser}, \bibinfo{person}{Caner Hazirbas},
  \bibinfo{person}{Vladimir Golkov}, \bibinfo{person}{Patrick Van Der~Smagt},
  \bibinfo{person}{Daniel Cremers}, {and} \bibinfo{person}{Thomas Brox}.}
  \bibinfo{year}{2015}\natexlab{}.
\newblock \showarticletitle{Flownet: Learning optical flow with convolutional
  networks}. In \bibinfo{booktitle}{\emph{Proceedings of the IEEE international
  conference on computer vision}}. \bibinfo{pages}{2758--2766}.
\newblock


\bibitem[\protect\citeauthoryear{Dou, Tu, Wang, Wang, Shi, and Zhang}{Dou
  et~al\mbox{.}}{2019}]%
        {dou2019dynamic}
\bibfield{author}{\bibinfo{person}{Zi-Yi Dou}, \bibinfo{person}{Zhaopeng Tu},
  \bibinfo{person}{Xing Wang}, \bibinfo{person}{Longyue Wang},
  \bibinfo{person}{Shuming Shi}, {and} \bibinfo{person}{Tong Zhang}.}
  \bibinfo{year}{2019}\natexlab{}.
\newblock \showarticletitle{Dynamic layer aggregation for neural machine
  translation with routing-by-agreement}. In \bibinfo{booktitle}{\emph{AAAI}}.
\newblock


\bibitem[\protect\citeauthoryear{Eitel, Springenberg, Spinello, Riedmiller, and
  Burgard}{Eitel et~al\mbox{.}}{2015}]%
        {eitel2015multimodal}
\bibfield{author}{\bibinfo{person}{Andreas Eitel}, \bibinfo{person}{Jost~Tobias
  Springenberg}, \bibinfo{person}{Luciano Spinello}, \bibinfo{person}{Martin
  Riedmiller}, {and} \bibinfo{person}{Wolfram Burgard}.}
  \bibinfo{year}{2015}\natexlab{}.
\newblock \showarticletitle{Multimodal deep learning for robust RGB-D object
  recognition}. In \bibinfo{booktitle}{\emph{2015 IEEE/RSJ International
  Conference on Intelligent Robots and Systems (IROS)}}.
  \bibinfo{pages}{681--687}.
\newblock


\bibitem[\protect\citeauthoryear{Gold, Rangarajan, Lu, Pappu, and
  Mjolsness}{Gold et~al\mbox{.}}{1998}]%
        {gold1998new}
\bibfield{author}{\bibinfo{person}{Steven Gold}, \bibinfo{person}{Anand
  Rangarajan}, \bibinfo{person}{Chien-Ping Lu}, \bibinfo{person}{Suguna Pappu},
  {and} \bibinfo{person}{Eric Mjolsness}.} \bibinfo{year}{1998}\natexlab{}.
\newblock \showarticletitle{New algorithms for 2D and 3D point matching: Pose
  estimation and correspondence}.
\newblock \bibinfo{journal}{\emph{Pattern recognition}} \bibinfo{volume}{31},
  \bibinfo{number}{8} (\bibinfo{year}{1998}), \bibinfo{pages}{1019--1031}.
\newblock


\bibitem[\protect\citeauthoryear{Guermandi, Shi, Dewilde, Derudder, Ahmad,
  Spagnolo, Ocket, Bourdoux, Wambacq, Craninckx, et~al\mbox{.}}{Guermandi
  et~al\mbox{.}}{2017}]%
        {guermandi201779}
\bibfield{author}{\bibinfo{person}{Davide Guermandi}, \bibinfo{person}{Qixian
  Shi}, \bibinfo{person}{Andy Dewilde}, \bibinfo{person}{Veerle Derudder},
  \bibinfo{person}{Ubaid Ahmad}, \bibinfo{person}{Annachiara Spagnolo},
  \bibinfo{person}{Ilja Ocket}, \bibinfo{person}{Andr{\'e} Bourdoux},
  \bibinfo{person}{Piet Wambacq}, \bibinfo{person}{Jan Craninckx},
  {et~al\mbox{.}}} \bibinfo{year}{2017}\natexlab{}.
\newblock \showarticletitle{A 79-GHz MIMO PMCW radar SoC in 28-nm CMOS}.
\newblock \bibinfo{journal}{\emph{IEEE Journal of Solid-State Circuits}}
  \bibinfo{volume}{52}, \bibinfo{number}{10} (\bibinfo{year}{2017}),
  \bibinfo{pages}{2613--2626}.
\newblock


\bibitem[\protect\citeauthoryear{Handa, Whelan, McDonald, and Davison}{Handa
  et~al\mbox{.}}{2014}]%
        {handa2014benchmark}
\bibfield{author}{\bibinfo{person}{Ankur Handa}, \bibinfo{person}{Thomas
  Whelan}, \bibinfo{person}{John McDonald}, {and} \bibinfo{person}{Andrew~J
  Davison}.} \bibinfo{year}{2014}\natexlab{}.
\newblock \showarticletitle{A benchmark for RGB-D visual odometry, 3D
  reconstruction and SLAM}. In \bibinfo{booktitle}{\emph{IEEE international
  conference on Robotics and automation (ICRA)}}.
\newblock


\bibitem[\protect\citeauthoryear{Hu, Huang, Zhao, Alempijevic, and
  Dissanayake}{Hu et~al\mbox{.}}{2012}]%
        {hu2012robust}
\bibfield{author}{\bibinfo{person}{Gibson Hu}, \bibinfo{person}{Shoudong
  Huang}, \bibinfo{person}{Liang Zhao}, \bibinfo{person}{Alen Alempijevic},
  {and} \bibinfo{person}{Gamini Dissanayake}.} \bibinfo{year}{2012}\natexlab{}.
\newblock \showarticletitle{A robust rgb-d slam algorithm}. In
  \bibinfo{booktitle}{\emph{2012 IEEE/RSJ International Conference on
  Intelligent Robots and Systems}}. \bibinfo{pages}{1714--1719}.
\newblock


\bibitem[\protect\citeauthoryear{Huttenlocher and Ullman}{Huttenlocher and
  Ullman}{1990}]%
        {huttenlocher1990recognizing}
\bibfield{author}{\bibinfo{person}{Daniel~P Huttenlocher} {and}
  \bibinfo{person}{Shimon Ullman}.} \bibinfo{year}{1990}\natexlab{}.
\newblock \showarticletitle{Recognizing solid objects by alignment with an
  image}.
\newblock \bibinfo{journal}{\emph{International journal of computer vision}}
  \bibinfo{volume}{5}, \bibinfo{number}{2} (\bibinfo{year}{1990}),
  \bibinfo{pages}{195--212}.
\newblock


\bibitem[\protect\citeauthoryear{Instruments}{Instruments}{[n. d.]a}]%
        {timmwave}
\bibfield{author}{\bibinfo{person}{Texas Instruments}.} \bibinfo{year}{[n.
  d.]}\natexlab{a}.
\newblock \bibinfo{title}{Automotive mmWave sensors}.
\newblock
\newblock
\urldef\tempurl%
\url{http://www.ti.com/sensors/mmwave/overview.html}
\showURL{%
\tempurl}


\bibitem[\protect\citeauthoryear{Instruments}{Instruments}{[n. d.]b}]%
        {ti_plastic}
\bibfield{author}{\bibinfo{person}{Texas Instruments}.} \bibinfo{year}{[n.
  d.]}\natexlab{b}.
\newblock \bibinfo{title}{mmWave sensors Overview}.
\newblock
\newblock
\urldef\tempurl%
\url{https://www.ti.com/sensors/mmwave-radar/overview.html}
\showURL{%
\tempurl}


\bibitem[\protect\citeauthoryear{Instruments}{Instruments}{[n. d.]c}]%
        {ti_training}
\bibfield{author}{\bibinfo{person}{Texas Instruments}.} \bibinfo{year}{[n.
  d.]}\natexlab{c}.
\newblock \bibinfo{title}{mmWave Training Series}.
\newblock
\newblock
\urldef\tempurl%
\url{https://training.ti.com/mmwave-training-series}
\showURL{%
\tempurl}


\bibitem[\protect\citeauthoryear{Joosting}{Joosting}{[n. d.]}]%
        {radar_drone}
\bibfield{author}{\bibinfo{person}{Jean-Pierre Joosting}.} \bibinfo{year}{[n.
  d.]}\natexlab{}.
\newblock \bibinfo{title}{Radar enables heavy lifting drones to navigate in
  complex environments}.
\newblock
\newblock
\urldef\tempurl%
\url{https://www.eenewseurope.com/news/radar-enables-heavy-lifting-drones-navigate-complex-environments?from=singlemessage&isappinstalled=0#}
\showURL{%
\tempurl}


\bibitem[\protect\citeauthoryear{Kim and Kim}{Kim and Kim}{2013}]%
        {kim2013image}
\bibfield{author}{\bibinfo{person}{Deok-Hwa Kim} {and}
  \bibinfo{person}{Jong-Hwan Kim}.} \bibinfo{year}{2013}\natexlab{}.
\newblock \showarticletitle{Image-Based ICP algorithm for visual odometry using
  a RGB-D sensor in a dynamic environment}.
\newblock In \bibinfo{booktitle}{\emph{Robot Intelligence Technology and
  Applications 2012}}. \bibinfo{pages}{423--430}.
\newblock


\bibitem[\protect\citeauthoryear{Kramer, Stahoviak, Santamaria-Navarro,
  Agha-mohammadi, and Heckman}{Kramer et~al\mbox{.}}{2020}]%
        {kramer2020radar}
\bibfield{author}{\bibinfo{person}{Andrew Kramer}, \bibinfo{person}{Carl
  Stahoviak}, \bibinfo{person}{Angel Santamaria-Navarro},
  \bibinfo{person}{Ali-akbar Agha-mohammadi}, {and}
  \bibinfo{person}{Christoffer Heckman}.} \bibinfo{year}{2020}\natexlab{}.
\newblock \showarticletitle{Radar-Inertial Ego-Velocity Estimation for Visually
  Degraded Environments}. In \bibinfo{booktitle}{\emph{IEEE International
  Conference on Robotics and Automation (ICRA)}}.
\newblock


\bibitem[\protect\citeauthoryear{LAPEDUS}{LAPEDUS}{2017}]%
        {radar_mutiple}
\bibfield{author}{\bibinfo{person}{MARK LAPEDUS}.}
  \bibinfo{year}{2017}\natexlab{}.
\newblock \bibinfo{title}{Here Comes High-Res Car Radar}.
\newblock
\newblock
\urldef\tempurl%
\url{https://semiengineering.com/here-comes-high-res-car-radar/}
\showURL{%
\tempurl}


\bibitem[\protect\citeauthoryear{Leutenegger, Lynen, Bosse, Siegwart, and
  Furgale}{Leutenegger et~al\mbox{.}}{2015}]%
        {leutenegger2015keyframe}
\bibfield{author}{\bibinfo{person}{Stefan Leutenegger}, \bibinfo{person}{Simon
  Lynen}, \bibinfo{person}{Michael Bosse}, \bibinfo{person}{Roland Siegwart},
  {and} \bibinfo{person}{Paul Furgale}.} \bibinfo{year}{2015}\natexlab{}.
\newblock \showarticletitle{Keyframe-based visual--inertial odometry using
  nonlinear optimization}.
\newblock \bibinfo{journal}{\emph{The International Journal of Robotics
  Research}} \bibinfo{volume}{34}, \bibinfo{number}{3} (\bibinfo{year}{2015}),
  \bibinfo{pages}{314--334}.
\newblock


\bibitem[\protect\citeauthoryear{Li, Zhang, and Xia}{Li et~al\mbox{.}}{2016}]%
        {li2016vehicle}
\bibfield{author}{\bibinfo{person}{Bo Li}, \bibinfo{person}{Tianlei Zhang},
  {and} \bibinfo{person}{Tian Xia}.} \bibinfo{year}{2016}\natexlab{}.
\newblock \showarticletitle{Vehicle detection from 3d lidar using fully
  convolutional network}.
\newblock \bibinfo{journal}{\emph{arXiv preprint arXiv:1608.07916}}
  (\bibinfo{year}{2016}).
\newblock


\bibitem[\protect\citeauthoryear{Li, Wang, Long, and Gu}{Li
  et~al\mbox{.}}{2018}]%
        {li2018undeepvo}
\bibfield{author}{\bibinfo{person}{Ruihao Li}, \bibinfo{person}{Sen Wang},
  \bibinfo{person}{Zhiqiang Long}, {and} \bibinfo{person}{Dongbing Gu}.}
  \bibinfo{year}{2018}\natexlab{}.
\newblock \showarticletitle{Undeepvo: Monocular visual odometry through
  unsupervised deep learning}. In \bibinfo{booktitle}{\emph{2018 IEEE
  international conference on robotics and automation (ICRA)}}.
  \bibinfo{pages}{7286--7291}.
\newblock


\bibitem[\protect\citeauthoryear{Liu, Yao, Li, Liu, Wang, Shao, and
  Abdelzaher}{Liu et~al\mbox{.}}{2020}]%
        {liu2020giobalfusion}
\bibfield{author}{\bibinfo{person}{Shengzhong Liu}, \bibinfo{person}{Shuochao
  Yao}, \bibinfo{person}{Jinyang Li}, \bibinfo{person}{Dongxin Liu},
  \bibinfo{person}{Tianshi Wang}, \bibinfo{person}{Huajie Shao}, {and}
  \bibinfo{person}{Tarek Abdelzaher}.} \bibinfo{year}{2020}\natexlab{}.
\newblock \showarticletitle{GIobalFusion: A Global Attentional Deep Learning
  Framework for Multisensor Information Fusion}.
\newblock \bibinfo{journal}{\emph{Proceedings of the ACM on Interactive,
  Mobile, Wearable and Ubiquitous Technologies}} \bibinfo{volume}{4},
  \bibinfo{number}{1} (\bibinfo{year}{2020}), \bibinfo{pages}{1--27}.
\newblock


\bibitem[\protect\citeauthoryear{Lu, Li, Zhao, Chen, Xie, Wen, Tan, and
  Trigoni}{Lu et~al\mbox{.}}{2018}]%
        {lu2018simultaneous}
\bibfield{author}{\bibinfo{person}{Chris~Xiaoxuan Lu}, \bibinfo{person}{Yang
  Li}, \bibinfo{person}{Peijun Zhao}, \bibinfo{person}{Changhao Chen},
  \bibinfo{person}{Linhai Xie}, \bibinfo{person}{Hongkai Wen},
  \bibinfo{person}{Rui Tan}, {and} \bibinfo{person}{Niki Trigoni}.}
  \bibinfo{year}{2018}\natexlab{}.
\newblock \showarticletitle{Simultaneous localization and mapping with power
  network electromagnetic field}. In \bibinfo{booktitle}{\emph{Proceedings of
  the 24th annual international conference on mobile computing and networking
  (MobiCom)}}.
\newblock


\bibitem[\protect\citeauthoryear{Lu, Rosa, Zhao, Wang, Chen, Stankovic,
  Trigoni, and Markham}{Lu et~al\mbox{.}}{2020}]%
        {lu2020millimap}
\bibfield{author}{\bibinfo{person}{Chris~Xiaoxuan Lu}, \bibinfo{person}{Stefano
  Rosa}, \bibinfo{person}{Peijun Zhao}, \bibinfo{person}{Bing Wang},
  \bibinfo{person}{Changhao Chen}, \bibinfo{person}{John~A. Stankovic},
  \bibinfo{person}{Niki Trigoni}, {and} \bibinfo{person}{Andrew Markham}.}
  \bibinfo{year}{2020}\natexlab{}.
\newblock \showarticletitle{See Through Smoke: Robust Indoor Mapping with
  Low-cost mmWave Radar}. In \bibinfo{booktitle}{\emph{ACM International
  Conference on Mobile Systems, Applications, and Services (MobiSys)}}.
\newblock


\bibitem[\protect\citeauthoryear{Lu, Cabrol, Steinbach, and Pragada}{Lu
  et~al\mbox{.}}{2013}]%
        {lu2013measurement}
\bibfield{author}{\bibinfo{person}{Jonathan~S Lu}, \bibinfo{person}{Patrick
  Cabrol}, \bibinfo{person}{Daniel Steinbach}, {and}
  \bibinfo{person}{Ravikumar~V Pragada}.} \bibinfo{year}{2013}\natexlab{}.
\newblock \showarticletitle{Measurement and characterization of various outdoor
  60 GHz diffracted and scattered paths}. In
  \bibinfo{booktitle}{\emph{MILCOM}}.
\newblock


\bibitem[\protect\citeauthoryear{Lymberopoulos, Liu, Yang, Choudhury,
  Handziski, and Sen}{Lymberopoulos et~al\mbox{.}}{2015}]%
        {lymberopoulos2015realistic}
\bibfield{author}{\bibinfo{person}{Dimitrios Lymberopoulos},
  \bibinfo{person}{Jie Liu}, \bibinfo{person}{Xue Yang},
  \bibinfo{person}{Romit~Roy Choudhury}, \bibinfo{person}{Vlado Handziski},
  {and} \bibinfo{person}{Souvik Sen}.} \bibinfo{year}{2015}\natexlab{}.
\newblock \showarticletitle{A realistic evaluation and comparison of indoor
  location technologies: Experiences and lessons learned}. In
  \bibinfo{booktitle}{\emph{Proceedings of the 14th international conference on
  information processing in sensor networks}}.
\newblock


\bibitem[\protect\citeauthoryear{MacCartney, Deng, Sun, and
  Rappaport}{MacCartney et~al\mbox{.}}{2016}]%
        {maccartney2016millimeter}
\bibfield{author}{\bibinfo{person}{George~R MacCartney}, \bibinfo{person}{Sijia
  Deng}, \bibinfo{person}{Shu Sun}, {and} \bibinfo{person}{Theodore~S
  Rappaport}.} \bibinfo{year}{2016}\natexlab{}.
\newblock \showarticletitle{Millimeter-wave human blockage at 73 GHz with a
  simple double knife-edge diffraction model and extension for directional
  antennas}. In \bibinfo{booktitle}{\emph{IEEE Vehicular Technology Conference
  (VTC-Fall)}}.
\newblock


\bibitem[\protect\citeauthoryear{Mamandipoor, Malysa, Arbabian, Madhow, and
  Noujeim}{Mamandipoor et~al\mbox{.}}{2014}]%
        {mamandipoor201460}
\bibfield{author}{\bibinfo{person}{Babak Mamandipoor}, \bibinfo{person}{Greg
  Malysa}, \bibinfo{person}{Amin Arbabian}, \bibinfo{person}{Upamanyu Madhow},
  {and} \bibinfo{person}{Karam Noujeim}.} \bibinfo{year}{2014}\natexlab{}.
\newblock \showarticletitle{60 ghz synthetic aperture radar for short-range
  imaging: Theory and experiments}. In \bibinfo{booktitle}{\emph{ACSSC}}.
\newblock


\bibitem[\protect\citeauthoryear{Myronenko and Song}{Myronenko and
  Song}{2010}]%
        {myronenko2010point}
\bibfield{author}{\bibinfo{person}{Andriy Myronenko} {and}
  \bibinfo{person}{Xubo Song}.} \bibinfo{year}{2010}\natexlab{}.
\newblock \showarticletitle{Point set registration: Coherent point drift}.
\newblock \bibinfo{journal}{\emph{IEEE transactions on pattern analysis and
  machine intelligence}} \bibinfo{volume}{32}, \bibinfo{number}{12}
  (\bibinfo{year}{2010}), \bibinfo{pages}{2262--2275}.
\newblock


\bibitem[\protect\citeauthoryear{Nicolai, Skeele, Eriksen, and
  Hollinger}{Nicolai et~al\mbox{.}}{2016}]%
        {nicolai2016deep}
\bibfield{author}{\bibinfo{person}{Austin Nicolai}, \bibinfo{person}{Ryan
  Skeele}, \bibinfo{person}{Christopher Eriksen}, {and}
  \bibinfo{person}{Geoffrey~A Hollinger}.} \bibinfo{year}{2016}\natexlab{}.
\newblock \showarticletitle{Deep learning for laser based odometry estimation}.
  In \bibinfo{booktitle}{\emph{RSS workshop Limits and Potentials of Deep
  Learning in Robotics}}.
\newblock


\bibitem[\protect\citeauthoryear{Nuttall and Bradshaw}{Nuttall and
  Bradshaw}{[n. d.]}]%
        {pixel4}
\bibfield{author}{\bibinfo{person}{Chris Nuttall} {and} \bibinfo{person}{Tim
  Bradshaw}.} \bibinfo{year}{[n. d.]}\natexlab{}.
\newblock \bibinfo{title}{Google draws on old radar technology for its motion
  sensor Pixel 4 smartphone}.
\newblock
\newblock
\urldef\tempurl%
\url{https://www.ft.com/content/02c051ec-f005-11e9-ad1e-4367d8281195}
\showURL{%
\tempurl}


\bibitem[\protect\citeauthoryear{Park, Shin, and Kim}{Park
  et~al\mbox{.}}{2020}]%
        {park2020pharao}
\bibfield{author}{\bibinfo{person}{Yeong~Sang Park}, \bibinfo{person}{Young-Sik
  Shin}, {and} \bibinfo{person}{Ayoung Kim}.} \bibinfo{year}{2020}\natexlab{}.
\newblock \showarticletitle{PhaRaO: Direct Radar Odometry using Phase
  Correlation}. In \bibinfo{booktitle}{\emph{IEEE ICRA}}.
\newblock


\bibitem[\protect\citeauthoryear{Parmar, Vaswani, Uszkoreit, Kaiser, Shazeer,
  Ku, and Tran}{Parmar et~al\mbox{.}}{2018}]%
        {parmar2018image}
\bibfield{author}{\bibinfo{person}{Niki Parmar}, \bibinfo{person}{Ashish
  Vaswani}, \bibinfo{person}{Jakob Uszkoreit}, \bibinfo{person}{{\L}ukasz
  Kaiser}, \bibinfo{person}{Noam Shazeer}, \bibinfo{person}{Alexander Ku},
  {and} \bibinfo{person}{Dustin Tran}.} \bibinfo{year}{2018}\natexlab{}.
\newblock \showarticletitle{Image transformer}. In
  \bibinfo{booktitle}{\emph{ICML}}.
\newblock


\bibitem[\protect\citeauthoryear{Petsios, Alivizatos, and Uzunoglu}{Petsios
  et~al\mbox{.}}{2008}]%
        {petsios2008solving}
\bibfield{author}{\bibinfo{person}{Michail~N Petsios},
  \bibinfo{person}{Emmanouil~G Alivizatos}, {and} \bibinfo{person}{Nikolaos~K
  Uzunoglu}.} \bibinfo{year}{2008}\natexlab{}.
\newblock \showarticletitle{Solving the association problem for a multistatic
  range-only radar target tracker}.
\newblock \bibinfo{journal}{\emph{Signal Processing}} (\bibinfo{year}{2008}).
\newblock


\bibitem[\protect\citeauthoryear{Quigley, Conley, Gerkey, Faust, Foote, Leibs,
  Wheeler, and Ng}{Quigley et~al\mbox{.}}{2009}]%
        {quigley2009ros}
\bibfield{author}{\bibinfo{person}{Morgan Quigley}, \bibinfo{person}{Ken
  Conley}, \bibinfo{person}{Brian Gerkey}, \bibinfo{person}{Josh Faust},
  \bibinfo{person}{Tully Foote}, \bibinfo{person}{Jeremy Leibs},
  \bibinfo{person}{Rob Wheeler}, {and} \bibinfo{person}{Andrew~Y Ng}.}
  \bibinfo{year}{2009}\natexlab{}.
\newblock \showarticletitle{ROS: an open-source Robot Operating System}. In
  \bibinfo{booktitle}{\emph{ICRA workshop on open source software}},
  Vol.~\bibinfo{volume}{3}. \bibinfo{pages}{5}.
\newblock


\bibitem[\protect\citeauthoryear{Radu and Marina}{Radu and Marina}{2013}]%
        {radu2013himloc}
\bibfield{author}{\bibinfo{person}{Valentin Radu} {and}
  \bibinfo{person}{Mahesh~K Marina}.} \bibinfo{year}{2013}\natexlab{}.
\newblock \showarticletitle{HiMLoc: Indoor smartphone localization via activity
  aware pedestrian dead reckoning with selective crowdsourced WiFi
  fingerprinting}. In \bibinfo{booktitle}{\emph{International conference on
  indoor positioning and indoor navigation}}.
\newblock


\bibitem[\protect\citeauthoryear{Richardson and Moskal}{Richardson and
  Moskal}{2011}]%
        {richardson2011strengths}
\bibfield{author}{\bibinfo{person}{Jeffrey~J Richardson} {and}
  \bibinfo{person}{L~Monika Moskal}.} \bibinfo{year}{2011}\natexlab{}.
\newblock \showarticletitle{Strengths and limitations of assessing forest
  density and spatial configuration with aerial LiDAR}.
\newblock \bibinfo{journal}{\emph{Remote Sensing of Environment}}
  \bibinfo{volume}{115}, \bibinfo{number}{10} (\bibinfo{year}{2011}),
  \bibinfo{pages}{2640--2651}.
\newblock


\bibitem[\protect\citeauthoryear{Rong and Sichitiu}{Rong and Sichitiu}{2006}]%
        {rong2006angle}
\bibfield{author}{\bibinfo{person}{Peng Rong} {and} \bibinfo{person}{Mihail~L
  Sichitiu}.} \bibinfo{year}{2006}\natexlab{}.
\newblock \showarticletitle{Angle of arrival localization for wireless sensor
  networks}. In \bibinfo{booktitle}{\emph{SECON}}.
\newblock


\bibitem[\protect\citeauthoryear{ROS.org}{ROS.org}{[n. d.]}]%
        {gmapping}
\bibfield{author}{\bibinfo{person}{ROS.org}.} \bibinfo{year}{[n.
  d.]}\natexlab{}.
\newblock \bibinfo{title}{gmapping}.
\newblock
\newblock
\urldef\tempurl%
\url{http://wiki.ros.org/gmapping}
\showURL{%
\tempurl}


\bibitem[\protect\citeauthoryear{Saputra, de~Gusmao, Lu, Almalioglu, Rosa,
  Chen, Wahlstrom, Wang, Markham, and Trigoni}{Saputra et~al\mbox{.}}{2020}]%
        {saputra2020deeptio}
\bibfield{author}{\bibinfo{person}{Muhamad Risqi~U Saputra},
  \bibinfo{person}{Pedro Porto~Buarque de Gusmao},
  \bibinfo{person}{Chris~Xiaoxuan Lu}, \bibinfo{person}{Yasin Almalioglu},
  \bibinfo{person}{Stefano Rosa}, \bibinfo{person}{Changhao Chen},
  \bibinfo{person}{Johan Wahlstrom}, \bibinfo{person}{Wei Wang},
  \bibinfo{person}{Andrew Markham}, {and} \bibinfo{person}{Niki Trigoni}.}
  \bibinfo{year}{2020}\natexlab{}.
\newblock \showarticletitle{Deeptio: A deep thermal-inertial odometry with
  visual hallucination}.
\newblock \bibinfo{journal}{\emph{IEEE Robotics and Automation Letters}}
  (\bibinfo{year}{2020}).
\newblock


\bibitem[\protect\citeauthoryear{Saputra, Markham, and Trigoni}{Saputra
  et~al\mbox{.}}{2018}]%
        {saputra2018visual}
\bibfield{author}{\bibinfo{person}{Muhamad Risqi~U Saputra},
  \bibinfo{person}{Andrew Markham}, {and} \bibinfo{person}{Niki Trigoni}.}
  \bibinfo{year}{2018}\natexlab{}.
\newblock \showarticletitle{Visual SLAM and structure from motion in dynamic
  environments: A survey}.
\newblock \bibinfo{journal}{\emph{ACM Computing Surveys (CSUR)}}
  \bibinfo{volume}{51}, \bibinfo{number}{2} (\bibinfo{year}{2018}).
\newblock


\bibitem[\protect\citeauthoryear{Shen, Gowda, and Roy~Choudhury}{Shen
  et~al\mbox{.}}{2018}]%
        {shen2018closing}
\bibfield{author}{\bibinfo{person}{Sheng Shen}, \bibinfo{person}{Mahanth
  Gowda}, {and} \bibinfo{person}{Romit Roy~Choudhury}.}
  \bibinfo{year}{2018}\natexlab{}.
\newblock \showarticletitle{Closing the gaps in inertial motion tracking}. In
  \bibinfo{booktitle}{\emph{Proceedings of the 24th Annual International
  Conference on Mobile Computing and Networking}}. \bibinfo{pages}{429--444}.
\newblock


\bibitem[\protect\citeauthoryear{Spence and Driver}{Spence and Driver}{2004}]%
        {spence2004crossmodal}
\bibfield{author}{\bibinfo{person}{Charles Spence} {and} \bibinfo{person}{Jon
  Driver}.} \bibinfo{year}{2004}\natexlab{}.
\newblock \bibinfo{booktitle}{\emph{Crossmodal space and crossmodal
  attention}}.
\newblock \bibinfo{publisher}{Oxford University Press}.
\newblock


\bibitem[\protect\citeauthoryear{Srivastava, Greff, and Schmidhuber}{Srivastava
  et~al\mbox{.}}{2015}]%
        {srivastava2015training}
\bibfield{author}{\bibinfo{person}{Rupesh~K Srivastava}, \bibinfo{person}{Klaus
  Greff}, {and} \bibinfo{person}{J{\"u}rgen Schmidhuber}.}
  \bibinfo{year}{2015}\natexlab{}.
\newblock \showarticletitle{Training very deep networks}. In
  \bibinfo{booktitle}{\emph{Advances in neural information processing
  systems}}. \bibinfo{pages}{2377--2385}.
\newblock


\bibitem[\protect\citeauthoryear{Ummenhofer, Zhou, Uhrig, Mayer, Ilg,
  Dosovitskiy, and Brox}{Ummenhofer et~al\mbox{.}}{2017}]%
        {ummenhofer2017demon}
\bibfield{author}{\bibinfo{person}{Benjamin Ummenhofer},
  \bibinfo{person}{Huizhong Zhou}, \bibinfo{person}{Jonas Uhrig},
  \bibinfo{person}{Nikolaus Mayer}, \bibinfo{person}{Eddy Ilg},
  \bibinfo{person}{Alexey Dosovitskiy}, {and} \bibinfo{person}{Thomas Brox}.}
  \bibinfo{year}{2017}\natexlab{}.
\newblock \showarticletitle{Demon: Depth and motion network for learning
  monocular stereo}. In \bibinfo{booktitle}{\emph{Proceedings of the IEEE
  Conference on Computer Vision and Pattern Recognition}}.
\newblock


\bibitem[\protect\citeauthoryear{Uttam and Culshaw}{Uttam and Culshaw}{1985}]%
        {uttam1985precision}
\bibfield{author}{\bibinfo{person}{Deepak Uttam} {and} \bibinfo{person}{B
  Culshaw}.} \bibinfo{year}{1985}\natexlab{}.
\newblock \showarticletitle{Precision time domain reflectometry in optical
  fiber systems using a frequency modulated continuous wave ranging technique}.
\newblock \bibinfo{journal}{\emph{Journal of Lightwave Technology}}
  (\bibinfo{year}{1985}).
\newblock


\bibitem[\protect\citeauthoryear{Vaswani, Shazeer, Parmar, Uszkoreit, Jones,
  Gomez, Kaiser, and Polosukhin}{Vaswani et~al\mbox{.}}{2017}]%
        {vaswani2017attention}
\bibfield{author}{\bibinfo{person}{Ashish Vaswani}, \bibinfo{person}{Noam
  Shazeer}, \bibinfo{person}{Niki Parmar}, \bibinfo{person}{Jakob Uszkoreit},
  \bibinfo{person}{Llion Jones}, \bibinfo{person}{Aidan~N Gomez},
  \bibinfo{person}{{\L}ukasz Kaiser}, {and} \bibinfo{person}{Illia
  Polosukhin}.} \bibinfo{year}{2017}\natexlab{}.
\newblock \showarticletitle{Attention is all you need}. In
  \bibinfo{booktitle}{\emph{Advances in neural information processing
  systems}}.
\newblock


\bibitem[\protect\citeauthoryear{Wang, Clark, Wen, and Trigoni}{Wang
  et~al\mbox{.}}{2017}]%
        {wang2017deepvo}
\bibfield{author}{\bibinfo{person}{Sen Wang}, \bibinfo{person}{Ronald Clark},
  \bibinfo{person}{Hongkai Wen}, {and} \bibinfo{person}{Niki Trigoni}.}
  \bibinfo{year}{2017}\natexlab{}.
\newblock \showarticletitle{Deepvo: Towards end-to-end visual odometry with
  deep recurrent convolutional neural networks}. In
  \bibinfo{booktitle}{\emph{2017 IEEE International Conference on Robotics and
  Automation (ICRA)}}. \bibinfo{pages}{2043--2050}.
\newblock


\bibitem[\protect\citeauthoryear{Wang, Clark, Wen, and Trigoni}{Wang
  et~al\mbox{.}}{2018a}]%
        {wang2018end}
\bibfield{author}{\bibinfo{person}{Sen Wang}, \bibinfo{person}{Ronald Clark},
  \bibinfo{person}{Hongkai Wen}, {and} \bibinfo{person}{Niki Trigoni}.}
  \bibinfo{year}{2018}\natexlab{a}.
\newblock \showarticletitle{End-to-end, sequence-to-sequence probabilistic
  visual odometry through deep neural networks}.
\newblock \bibinfo{journal}{\emph{The International Journal of Robotics
  Research}} (\bibinfo{year}{2018}).
\newblock


\bibitem[\protect\citeauthoryear{Wang, Saputra, Zhao, Gusmao, Yang, Chen,
  Markham, and Trigoni}{Wang et~al\mbox{.}}{2019}]%
        {wang2019deeppco}
\bibfield{author}{\bibinfo{person}{Wei Wang}, \bibinfo{person}{Muhamad Risqi~U
  Saputra}, \bibinfo{person}{Peijun Zhao}, \bibinfo{person}{Pedro Gusmao},
  \bibinfo{person}{Bo Yang}, \bibinfo{person}{Changhao Chen},
  \bibinfo{person}{Andrew Markham}, {and} \bibinfo{person}{Niki Trigoni}.}
  \bibinfo{year}{2019}\natexlab{}.
\newblock \showarticletitle{DeepPCO: End-to-End Point Cloud Odometry through
  Deep Parallel Neural Network}.
\newblock \bibinfo{journal}{\emph{International Conference on Intelligent
  Robots and Systems}} (\bibinfo{year}{2019}).
\newblock


\bibitem[\protect\citeauthoryear{Wang, Girshick, Gupta, and He}{Wang
  et~al\mbox{.}}{2018b}]%
        {wang2018non}
\bibfield{author}{\bibinfo{person}{Xiaolong Wang}, \bibinfo{person}{Ross
  Girshick}, \bibinfo{person}{Abhinav Gupta}, {and} \bibinfo{person}{Kaiming
  He}.} \bibinfo{year}{2018}\natexlab{b}.
\newblock \showarticletitle{Non-local neural networks}. In
  \bibinfo{booktitle}{\emph{CVPR}}.
\newblock


\bibitem[\protect\citeauthoryear{Ward}{Ward}{1969}]%
        {ward1969handbook}
\bibfield{author}{\bibinfo{person}{DK~Barton~HR Ward}.}
  \bibinfo{year}{1969}\natexlab{}.
\newblock \bibinfo{booktitle}{\emph{Handbook of radar measurement}}.
\newblock


\bibitem[\protect\citeauthoryear{Weston, Cen, Newman, and Posner}{Weston
  et~al\mbox{.}}{2018}]%
        {weston2018probably}
\bibfield{author}{\bibinfo{person}{Rob Weston}, \bibinfo{person}{Sarah Cen},
  \bibinfo{person}{Paul Newman}, {and} \bibinfo{person}{Ingmar Posner}.}
  \bibinfo{year}{2018}\natexlab{}.
\newblock \showarticletitle{Probably unknown: Deep inverse sensor modelling in
  radar}. In \bibinfo{booktitle}{\emph{ICRA}}.
\newblock


\bibitem[\protect\citeauthoryear{Xiao, Wen, Markham, and Trigoni}{Xiao
  et~al\mbox{.}}{2014}]%
        {xiao2014lightweight}
\bibfield{author}{\bibinfo{person}{Zhuoling Xiao}, \bibinfo{person}{Hongkai
  Wen}, \bibinfo{person}{Andrew Markham}, {and} \bibinfo{person}{Niki
  Trigoni}.} \bibinfo{year}{2014}\natexlab{}.
\newblock \showarticletitle{Lightweight map matching for indoor localisation
  using conditional random fields}. In \bibinfo{booktitle}{\emph{Proceedings of
  the 13th International Symposium on Information Processing in Sensor
  Networks}}.
\newblock


\bibitem[\protect\citeauthoryear{Xu, Zheng, and Hranilovic}{Xu
  et~al\mbox{.}}{2015}]%
        {xu2015idyll}
\bibfield{author}{\bibinfo{person}{Qiang Xu}, \bibinfo{person}{Rong Zheng},
  {and} \bibinfo{person}{Steve Hranilovic}.} \bibinfo{year}{2015}\natexlab{}.
\newblock \showarticletitle{IDyLL: Indoor localization using inertial and light
  sensors on smartphones}. In \bibinfo{booktitle}{\emph{ACM International Joint
  Conference on Pervasive and Ubiquitous Computing}}.
\newblock


\bibitem[\protect\citeauthoryear{Xue, Jiang, Miao, Yuan, Ma, Ma, Wang, Yao, Xu,
  Zhang, et~al\mbox{.}}{Xue et~al\mbox{.}}{2019}]%
        {xue2019deepfusion}
\bibfield{author}{\bibinfo{person}{Hongfei Xue}, \bibinfo{person}{Wenjun
  Jiang}, \bibinfo{person}{Chenglin Miao}, \bibinfo{person}{Ye Yuan},
  \bibinfo{person}{Fenglong Ma}, \bibinfo{person}{Xin Ma},
  \bibinfo{person}{Yijiang Wang}, \bibinfo{person}{Shuochao Yao},
  \bibinfo{person}{Wenyao Xu}, \bibinfo{person}{Aidong Zhang}, {et~al\mbox{.}}}
  \bibinfo{year}{2019}\natexlab{}.
\newblock \showarticletitle{DeepFusion: A Deep Learning Framework for the
  Fusion of Heterogeneous Sensory Data}. In
  \bibinfo{booktitle}{\emph{Proceedings of the Twentieth ACM International
  Symposium on Mobile Ad Hoc Networking and Computing}}.
  \bibinfo{pages}{151--160}.
\newblock


\bibitem[\protect\citeauthoryear{Yan, Li, Xie, Bao, Liao, Huang, Ren, Ahmed,
  Wang, et~al\mbox{.}}{Yan et~al\mbox{.}}{2016}]%
        {yan2016multipath}
\bibfield{author}{\bibinfo{person}{Yan Yan}, \bibinfo{person}{Long Li},
  \bibinfo{person}{Guodong Xie}, \bibinfo{person}{Changjing Bao},
  \bibinfo{person}{Peicheng Liao}, \bibinfo{person}{Hao Huang},
  \bibinfo{person}{Yongxiong Ren}, \bibinfo{person}{Nisar Ahmed},
  \bibinfo{person}{Zhe Wang}, {et~al\mbox{.}}} \bibinfo{year}{2016}\natexlab{}.
\newblock \showarticletitle{Multipath effects in millimetre-wave wireless
  communication using orbital angular momentum multiplexing}.
\newblock \bibinfo{journal}{\emph{Scientific reports}}  \bibinfo{volume}{6}
  (\bibinfo{year}{2016}), \bibinfo{pages}{33482}.
\newblock


\bibitem[\protect\citeauthoryear{Yang, Shi, and Carlone}{Yang
  et~al\mbox{.}}{2020}]%
        {yang2020teaser}
\bibfield{author}{\bibinfo{person}{Heng Yang}, \bibinfo{person}{Jingnan Shi},
  {and} \bibinfo{person}{Luca Carlone}.} \bibinfo{year}{2020}\natexlab{}.
\newblock \showarticletitle{TEASER: Fast and Certifiable Point Cloud
  Registration}.
\newblock \bibinfo{journal}{\emph{arXiv preprint arXiv:2001.07715}}
  (\bibinfo{year}{2020}).
\newblock


\bibitem[\protect\citeauthoryear{Yang, Wang, Stuckler, and Cremers}{Yang
  et~al\mbox{.}}{2018}]%
        {yang2018deep}
\bibfield{author}{\bibinfo{person}{Nan Yang}, \bibinfo{person}{Rui Wang},
  \bibinfo{person}{Jorg Stuckler}, {and} \bibinfo{person}{Daniel Cremers}.}
  \bibinfo{year}{2018}\natexlab{}.
\newblock \showarticletitle{Deep virtual stereo odometry: Leveraging deep depth
  prediction for monocular direct sparse odometry}. In
  \bibinfo{booktitle}{\emph{Proceedings of the European Conference on Computer
  Vision (ECCV)}}. \bibinfo{pages}{817--833}.
\newblock


\bibitem[\protect\citeauthoryear{Yao, Hu, Zhao, Zhang, and Abdelzaher}{Yao
  et~al\mbox{.}}{2017}]%
        {yao2017deepsense}
\bibfield{author}{\bibinfo{person}{Shuochao Yao}, \bibinfo{person}{Shaohan Hu},
  \bibinfo{person}{Yiran Zhao}, \bibinfo{person}{Aston Zhang}, {and}
  \bibinfo{person}{Tarek Abdelzaher}.} \bibinfo{year}{2017}\natexlab{}.
\newblock \showarticletitle{Deepsense: A unified deep learning framework for
  time-series mobile sensing data processing}. In
  \bibinfo{booktitle}{\emph{Proceedings of the 26th International Conference on
  World Wide Web}}.
\newblock


\bibitem[\protect\citeauthoryear{Yao, Zhao, Shao, Liu, Liu, Hao, Piao, Hu, Su,
  and Abdelzaher}{Yao et~al\mbox{.}}{2019}]%
        {yao2019sadeepsense}
\bibfield{author}{\bibinfo{person}{Shuochao Yao}, \bibinfo{person}{Yiran Zhao},
  \bibinfo{person}{Huajie Shao}, \bibinfo{person}{Dongxin Liu},
  \bibinfo{person}{Shengzhong Liu}, \bibinfo{person}{Yifan Hao},
  \bibinfo{person}{Ailing Piao}, \bibinfo{person}{Shaohan Hu},
  \bibinfo{person}{Lu Su}, {and} \bibinfo{person}{Tarek~F Abdelzaher}.}
  \bibinfo{year}{2019}\natexlab{}.
\newblock \showarticletitle{SADeepSense: Self-Attention Deep Learning Framework
  for Heterogeneous On-Device Sensors in Internet of Things Applications}. In
  \bibinfo{booktitle}{\emph{IEEE INFOCOM 2019-IEEE Conference on Computer
  Communications}}.
\newblock


\bibitem[\protect\citeauthoryear{Ye, Li, Huang, Du, and Zhang}{Ye
  et~al\mbox{.}}{2018}]%
        {ye20183d}
\bibfield{author}{\bibinfo{person}{Xiaoqing Ye}, \bibinfo{person}{Jiamao Li},
  \bibinfo{person}{Hexiao Huang}, \bibinfo{person}{Liang Du}, {and}
  \bibinfo{person}{Xiaolin Zhang}.} \bibinfo{year}{2018}\natexlab{}.
\newblock \showarticletitle{3d recurrent neural networks with context fusion
  for point cloud semantic segmentation}. In
  \bibinfo{booktitle}{\emph{Proceedings of the European Conference on Computer
  Vision (ECCV)}}.
\newblock


\bibitem[\protect\citeauthoryear{Zhao, Woodford, Wei, Kun, and Zhang}{Zhao
  et~al\mbox{.}}{2020}]%
        {zhao2020m}
\bibfield{author}{\bibinfo{person}{Renjie Zhao}, \bibinfo{person}{Timothy
  Woodford}, \bibinfo{person}{Teng Wei}, \bibinfo{person}{Qian Kun}, {and}
  \bibinfo{person}{Xinyu Zhang}.} \bibinfo{year}{2020}\natexlab{}.
\newblock \showarticletitle{M-Cube: A Millimeter-Wave Massive MIMO Software
  Radio}. In \bibinfo{booktitle}{\emph{ACM MobiCom}}.
\newblock


\end{thebibliography}

\end{document}